%% file: main.tex
\documentclass{article}

  \usepackage[main, final]{neurips_2026}

\usepackage[utf8]{inputenc} 
\usepackage[T1]{fontenc}    
\usepackage{hyperref}       
\usepackage{url}            
\usepackage{booktabs}       
\usepackage{tabularx}
\usepackage{array}
\usepackage{placeins}
\usepackage{mathrsfs}

\newcolumntype{Y}{>{\raggedright\arraybackslash}X}
\usepackage{amsfonts}       
\usepackage{nicefrac}       
\usepackage{microtype}      
\usepackage{xcolor}         

\usepackage{amsmath}
\usepackage{amssymb}
\usepackage{mathtools}
\usepackage{amsthm}

\theoremstyle{plain}
\newtheorem{theorem}{Theorem}[section]
\newtheorem{proposition}[theorem]{Proposition}
\newtheorem{lemma}[theorem]{Lemma}
\newtheorem{corollary}[theorem]{Corollary}
\theoremstyle{definition}

\newtheorem{assumption}[theorem]{Assumption}
\theoremstyle{remark}
\newtheorem{remark}[theorem]{Remark}

\title{From Likelihood Convergence to Parameter Convergence in POMDPs}

\author{
  \textbf{Jack Zhang}$^{1}$ \quad
  \textbf{Saurabh Amin}$^{1}$ \quad
  \textbf{Jiawei Zhang}$^{2}$ \quad
  \textbf{Patrick Jaillet}$^{1}$ \\[3pt]
  $^{1}$Operations Research Center, Massachusetts Institute of Technology \\
  Cambridge, MA 02139 \\
  $^{2}$Computer Sciences Department, University of Wisconsin--Madison \\
  Madison, WI 53706 \\[3pt]
  \texttt{\{jackz111,amins,jaillet\}@mit.edu} \\
  \texttt{jzhang2924@wisc.edu}
}

\begin{document}

\maketitle

\begin{abstract}
Likelihood-based methods are widely used to learn the parameters of partially observable Markov decision processes (POMDPs). While finite-sample guarantees for policy suboptimality exist for certain POMDP subclasses, analogous guarantees for parameter recovery remain less well understood. For any candidate POMDP parameter, we bound its parameter recovery error in terms of its likelihood gap to the ground truth, the structural conditioning of the observation and transition kernels, and the behavior policy used to collect data. The bound applies in both the undercomplete and overcomplete regimes, with the latter following naturally from the undercomplete analysis. Because the result applies to any sufficiently good candidate, it yields finite-sample guarantees for the maximum-likelihood estimator and any procedure producing a parameter with controlled likelihood gap. This is particularly useful for near-optimal learning in infinite-horizon POMDPs, where existing approaches largely rely on model-based planning with accurate transition and observation kernel estimates. The bound separates the contributions of the behavior policy and model, showing how data-collection choices, including action coverage, and their effects on mixing can accelerate parameter learning.
\end{abstract}

\section{Introduction}
\label{sec: introduction}
Partially observable Markov decision processes (POMDPs) provide a general framework for
sequential decision making under partial observability, with applications in robotics \citep{koenig1996unsupervised}, healthcare \citep{li2018actor}, recommendation systems \citep{boutilier2002pomdp}, and operations research \citep{bravo2019use}. POMDPs introduce uncertainty about the underlying state: the agent must make decisions based on its observation-action history or belief state. This uncertainty substantially increases the difficulty of planning and learning \citep{kaelbling1998planning}. Indeed, even finite-horizon planning with a known POMDP model is PSPACE-complete \citep{papadimitriou1987complexity}.

A prominent line of work studies \emph{model-based learning} for POMDPs, in which model parameters are estimated from data and used for planning. The learned model enables recursive belief-state updates from observation and action histories. The resulting belief states serve as sufficient statistics for decision-making within the learned POMDP \citep{kaelbling1998planning}. This belief-state representation allows the agent to account for both reward maximization and information gathering \citep{spaan2012partially}, while leveraging the extensive body of POMDP planning methods developed over the past two decades \citep{shani2013survey}. In particular, existing methods for learning in infinite-horizon POMDPs have largely followed this model-based paradigm, relying on estimates of the transition and observation kernels to compute belief states and obtain near-optimal policies \citep{ross2009sensitivity}.

Two major approaches to model-based POMDP learning are spectral methods and likelihood-based methods. Spectral methods estimate cross-moment statistics of observation sequences and employ matrix or tensor decompositions to recover the model kernels, providing finite-sample guarantees under suitable assumptions while avoiding direct optimization of the nonconvex likelihood. Likelihood-based methods, by contrast, seek to maximize the likelihood of observed trajectories, often using the Baum--Welch algorithm \citep{baum1970maximization}, and are widely used in practice.

\subsection{Motivation}
Spectral and likelihood-based methods face complementary limitations, leaving an important gap in finite-sample kernel recovery. Standard spectral methods for POMDPs require a full-column-rank observation kernel for identifiability, thereby restricting their guarantees to \emph{undercomplete} POMDPs, where the number of latent states does not exceed the number of observations. Moreover, stable recovery requires the observation kernel to admit a well-conditioned left inverse, ensuring that estimation errors in empirical moments are not excessively amplified. In \emph{overcomplete} POMDPs, where the number of latent states exceeds the number of observations, the observation kernel is necessarily rank-deficient. Consequently, distinct latent-state distributions can induce identical one-step observation distributions, and standard spectral recovery guarantees no longer apply.

Likelihood-based methods do not require the observation kernel to have full column rank and can be formulated in both undercomplete and overcomplete regimes. However, their finite-sample guarantees for parameter recovery remain limited. Classical results establish consistency and asymptotic normality \citep{leroux1992maximum,bickel1998asymptotic}, while finite-sample analyses have been developed for specific algorithms such as Baum--Welch \citep{yang2017statistical}. A central remaining challenge is to establish a general relationship that translates the likelihood gap of an arbitrary candidate model relative to the true model into a bound on parameter recovery error. Such a relationship would provide an estimator-agnostic recovery guarantee, allowing any candidate with a certified likelihood-gap bound to inherit a corresponding parameter-error bound, regardless of how it was computed.

Our focus on parameter recovery, rather than policy suboptimality, is motivated by the role of accurate transition and observation kernels in belief-state estimation and near-optimal control for infinite-horizon POMDPs. For finite-horizon POMDPs, near-optimal learning can sometimes be achieved by estimating predictive distributions of future observations without recovering the full model parameters. In contrast, existing approaches to near-optimal learning in infinite-horizon POMDPs have largely been model-based, relying on estimates of the transition and observation kernels (up to latent-state relabeling) for belief-state updates \citep{ross2009sensitivity,russo2026spectral}. However, whether accurate kernel recovery is fundamentally necessary for near-optimal learning remains unclear. Finite-sample guarantees directly targeting policy suboptimality primarily address the finite-horizon setting, whereas spectral recovery guarantees generally require undercomplete models. Consequently, these two lines of results do not fully cover infinite-horizon POMDPs, particularly in the overcomplete regime.

Beyond its role in policy learning, parameter recovery is also a well-established objective in latent-variable modeling. Hidden Markov models (HMMs), which can be viewed as POMDPs with a singleton action set, have long been studied independently of control objectives: recovering their transition and observation kernels enables latent-state inference, forecasting, simulation, and interpretation of the underlying dynamics. Similarly, in POMDPs, the transition and observation kernels characterize the underlying dynamics and observation mechanism independently of the reward specification. Recovering these kernels therefore yields a model reusable across multiple downstream tasks and analyses with different reward kernels. Appendix~\ref{apx:reusable model for downstream reward} provides a concrete example.

The theoretical and practical motivations raise the central question: Can we establish finite-sample guarantees for POMDP parameter recovery that translate the likelihood gap of an arbitrary candidate model into a quantitative parameter-error bound in both undercomplete and overcomplete regimes?

\subsection{Related work}\label{subsec: related work}
\paragraph{Spectral methods.}
Spectral learning for discrete POMDPs was introduced by \citet{azizzadenesheli2016reinforcement}, with subsequent extensions to rich-observation MDPs \citep{azizzadenesheli2016reinforcement2} and episodic PAC learning \citep{guo2016pac}. These approaches build on the broader spectral literature for latent-variable models \citep{anandkumar2012method,anandkumar2014tensor,hsu2012spectral} and require undercompleteness. More recently, \citet{xiong2022sublinear} used spectral methods to establish sublinear regret for infinite-horizon POMDPs with state-dependent rewards. \citet{russo2025achieving} developed an action-wise spectral estimator achieving $\widetilde{O}(\sqrt{T})$ regret with known observation kernels. \citet{russo2026spectral} later extended this approach to unknown observation kernels through mixed spectral estimation, recovering both kernels from data collected under multiple policies while retaining the $\widetilde{O}(\sqrt{T})$ regret guarantee. Depiste these advances, the methods of \citet{xiong2022sublinear} and \citet{russo2025achieving,russo2026spectral} continue to require undercompleteness. \citet{shaw2026toward} extended spectral learning beyond this restriction by employing predictive state representations and tensor decomposition to recover model parameters up to a similarity transformation. Their results accommodate certain overcomplete POMDPs but provide asymptotic rather than finite-sample guarantees.

\paragraph{Likelihood-based methods for HMMs.} Classical results establish consistency and asymptotic normality of maximum likelihood estimators (MLEs) for HMMs \citep{leroux1992maximum,bickel1998asymptotic}, with extensions to general state spaces \citep{douc2001asymptotics,douc2011consistency}. Finite-sample guarantees for parameter recovery, however, remain limited. \citet{lehericy2021nonasymptotic} bounds the KL divergence between the observation distributions induced by the MLE and the true model, rather than directly controlling parameter error. \citet{yang2017statistical} establish finite-sample contraction of Baum--Welch's iterates to a neighborhood of the true parameters under local regularity, mixing, and a uniform-positivity assumption on the transition kernel. These results characterize how closely the estimated model reproduces the true distribution of observation sequences and the convergence of Baum–Welch, respectively, rather than directly relating likelihood gaps to parameter recovery errors.

\paragraph{Likelihood-based methods for POMDPs.}
The model-dependent component of POMDP trajectory likelihood has an input-output HMM structure, even under adaptive behavior policies \citep{krishnamurthy2016partially}. This connection allows likelihood-based techniques for HMMs to be adapted to POMDPs. In POMDP learning, likelihood-based estimation is often incorporated into broader frameworks, including Optimistic MLE approaches of \citet{chen2023partially} and \citet{liu2022partially,liu2023optimistic}, as well as related procedures for hindsight-observable variants \citep{lee2023learning,guo2024sample,guo2023provably}. These works primarily establish sample-complexity guarantees for policy suboptimality in episodic POMDPs. Separately, \citet{jahromi2022online} employ Bayesian posterior sampling to establish sublinear regret for infinite-horizon POMDPs with known observation kernels, although their general-case guarantee relies on additional assumptions concerning parameter and belief estimation.

\subsection{Contributions}
\textbf{Finite-sample, candidate-agnostic bounds linking likelihood gaps to parameter recovery errors in undercomplete and overcomplete POMDPs.} For any candidate model, we bound its parameter recovery error, up to a permutation of the latent states, in terms of its likelihood gap relative to the true model, the structural conditioning of the kernels, and the sample size. This guarantee applies to MLE and, more generally, to any procedure that produces a parameter estimate with a controlled likelihood gap. Both the undercomplete and overcomplete regimes are covered by a unified framework, with the analysis of the latter building on that of the former. To the best of our knowledge, this is the first finite-sample guarantee for full parameter recovery in overcomplete POMDPs.

\textbf{Full kernel recovery provides a reusable model beyond fixed episodic tasks.} As discussed in Section~\ref{subsec: related work}, near-optimal policy-learning guarantees without full parameter recovery typically concern episodic POMDPs with fixed horizons, fixed initial-state distributions, and reset-based data collection. By contrast, our result provides an estimator-agnostic parameter-recovery guarantee that supports model-based planning in infinite-horizon POMDPs \citep{ross2009sensitivity}. The resulting model supports planning from arbitrary initial-state distributions, replanning without system resets, and other downstream tasks requiring accurate kernel estimates.

\textbf{Explicit dependence on the policy.} Our bound distinguishes the structural properties of the model from the effects of the behavior policy, explicitly quantifying the role of the behavior policy through its action coverage and the stationary distribution and mixing rate of the induced Markov chain.

\subsection{Paper organization}
Section~\ref{sec: problem setup} formalizes the POMDP model and the likelihood formulation. Section~\ref{sec: main result} establishes our main results, beginning with the undercomplete case and extending to the overcomplete case. Section~\ref{sec: discussion} summarizes our main contributions. Section~\ref{sec: future work} discusses limitations and future work.

\section{Problem setup}\label{sec: problem setup}
We consider a POMDP with finite state, action, and observation spaces, $\mathcal{M} := (\mathcal{S},\mathcal{A},\{T^{a}_{\beta^*}\}_{a\in\mathcal{A}}, R,\Omega,O_{\varphi^*},\gamma,\mu_{\beta^*})$, where $\mathcal{S}$, $\mathcal{A}$, and $\Omega$ denote the state, action, and observation spaces, respectively; \(\{T_{\beta^*}^{a}\}_{a\in\mathcal{A}}\) is the transition kernel; \(R:\mathcal S\times\mathcal A\to\mathbb R\) is the reward function; \(O_{\varphi^*}:\mathcal{S}\rightarrow\Delta(\Omega)\) is the observation kernel; \(\mu_{\beta^*}\) is the initial-state distribution. We use \(\beta\) to parameterize the transition kernels and initial-state distribution, and \(\varphi\) to parametrize the observation kernel. The ground-truth parameter is denoted by \(\theta^*=(\beta^*,\varphi^*)\).

We focus on the single-trajectory setting, where temporal dependence requires mixing-based concentration. With independent trajectories, the same analysis applies to the pooled data. Hence the single-trajectory case captures the main dependence-related difficulty.

We consider a trajectory \(\tau=(o_{1:n},a_{1:n})\) collected under a behavior policy \(\pi\) and the ground-truth parameter \(\theta^*\), and let \(s_{1:n}\) denote the corresponding latent state sequence. The assumptions on $\pi$ are specified in Sections~\ref{subsec: main result-undercomplete} and \ref{subsec: main result-overcomplete} for under- and overcomplete regimes. For any \(\theta=(\beta,\varphi)\), the model-dependent component of the complete-data trajectory likelihood is 
$
f_\theta(o_1^n,s_1^n;a_1^n)
    =\mu_{\beta}(s_{1})\prod_{i=2}^{n}T^{a_{i-1}}_{\beta}(s_{i}|s_{i-1})\prod_{i=1}^{n}O_{\varphi}(o_{i}|s_{i})$.

Since \(\pi\) is fixed and independent of \(\theta\), its contribution to the trajectory likelihood does not affect the relative likelihood between any pair of model parameters $(\theta,\theta')$. Accordingly, we define the model-dependent component of the observed-data log-likelihood, as
$
\ell_n(\theta)
:= \log \sum_{s_{1:n}\in\mathcal S^n}
f_\theta(o_{1:n},s_{1:n}; a_{1:n})$.

A small log-likelihood gap in $\ell_{n}$ does not, in general, guarantee accurate recovery of the transition and observation kernels. We therefore seek to identify conditions under which such a gap yields a quantitative bound on parameter recovery error, up to latent-state relabeling.

\section{Main result}\label{sec: main result}
\subsection{Undercomplete POMDPs \texorpdfstring{(\(|\mathcal S|\le |\Omega|\))}{(|S| <= |Omega|)}}\label{subsec: main result-undercomplete}
For \(\theta=(\beta,\varphi)\), let
\[
b_{t,\theta}(s)\coloneqq \mathbb P_\theta(s_t=s\mid o_{1:t},a_{1:t-1}),
\qquad
\bar b_{1,\theta}\coloneqq \mu_\beta,\qquad
\bar b_{t,\theta}(s)\coloneqq \mathbb P_\theta(s_t=s\mid o_{1:t-1},a_{1:t-1}).
\]
The standard Bayes filter gives
\[
\bar b_{t,\theta}(s)=\sum_{s'}T^{a_{t-1}}_\beta(s\mid s')b_{t-1,\theta}(s'),
\qquad
b_{t,\theta}(s)=
\frac{O_\varphi(o_t\mid s)\bar b_{t,\theta}(s)}
{\sum_{s'}O_\varphi(o_t\mid s')\bar b_{t,\theta}(s')}.
\]
Since \(\bar b_{t,\theta}\) is the pre-observation belief at time \(t\), it induces the predictive distribution of \(o_t\). For this reason, we call $\bar{b}_{t,\theta}$ the predictive belief at time $t$ induced by $\theta$, while $b_{t,\theta}$ is the posterior belief.

Having defined the predictive and posterior beliefs, we next specify the behavior policy used to collect trajectory data, as the subsequent assumptions and analysis depend on this choice.

\begin{assumption}\label{ass: behavior policy undercomplete}
The behavior policy satisfies \(\pi\in\Pi_{\rm obs}:=\{\pi':\Omega\to\Delta(\mathcal A)\}\).
\end{assumption}

\begin{assumption}\label{ass: minimum policy mass}
    $\pi(a|o)\geq \pi_{\min}>0$ for all $a\in\mathcal{A}$ and $o\in\Omega$.
\end{assumption}

We then use predictive beliefs to express the trajectory likelihood in two equivalent forms: one as a product over individual observations and the other as a product over consecutive observation pairs.

\begin{proposition}
\label{prop:factorization}
For $\theta=(\beta,\varphi)$,
$
\ell_{n}(\theta)
=
\sum_{i=1}^{n}
\log\left(\sum_{s \in \mathcal S} \bar b_{i,\theta}(s)\, O_{\varphi}(o_{i} \mid s)\right).
$
\end{proposition}

\begin{proof}
    See Appendix~\ref{apx:factorization proof}.
\end{proof}

\begin{proposition}\label{prop:length-2 factorization}
Assume $n$ is even for notational simplicity. Then, for any $\theta=(\beta,\varphi)$, we have
$
\ell_n(\theta)
=
\sum_{k=1}^{n/2}\log\Big(
\sum_{s\in\mathcal S}
\bar b_{2k-1,\theta}(s)\;
O_{\varphi}(o_{2k-1}\mid s)
\sum_{s'\in\mathcal S}T^{a_{2k-1}}_{\beta}(s'\mid s)\,O_{\varphi}(o_{2k}\mid s')
\Big)$.
\end{proposition}

\begin{proof}
    See Appendix~\ref{apx: 2-step factorization proof}.
\end{proof}

The one-step and two-step likelihood representations allow us to relate likelihood gaps to discrepancies between the corresponding predictive distributions. Specifically, we establish two high-probability inequalities showing that the cumulative squared discrepancy between the candidate and ground-truth predictive distributions is bounded by the log-likelihood gap plus a statistical error term.

\begin{lemma}\label{lem:quadratic bound for O}
    For any candidate parameter $\hat{\theta}=(\hat{\beta},\hat{\varphi})$, there exists a universal constant $c>0$ such that for any $\delta \in (0,1)$, with probability at least $1-\delta$, it holds that
    \begin{equation}\label{equ: general one-step bound}
    \begin{aligned}
    \sum_{i=1}^{n}\left(\sum_{o}\left|\sum_{s}\bar{b}_{i,\hat{\theta}}(s)O_{\hat{\varphi}}(o|s)-\sum_{s}\bar{b}_{i,\theta^*}(s)O_{\varphi^*}(o|s)\right|\right)^{2}
    \leq  4\left(\ell_{n}(\theta^*)-\ell_{n}(\hat{\theta})+2\log(1/\delta)\right).
    \end{aligned}
    \end{equation}
\end{lemma}
\begin{proof}
    See Appendix~\ref{apx: proof for quadratic bound for O}.
\end{proof}

\begin{lemma}\label{lem:length-2 quadratic bound for O}
For any candidate parameter $\hat{\theta}=(\hat{\beta},\hat{\varphi})$, for any
$\delta\in(0,1)$, with probability at least $1-\delta$, it holds that
\begin{equation}\label{equ: general two-step bound}
\begin{aligned}
\sum_{k=1}^{n}
&\Bigg(
\sum_{o,o'}
\Bigg|
\sum_{s\in\mathcal S}\bar b_{k,\hat\theta}(s)\,
\mathbb{P}_{\hat\theta}(o,o'\mid s,\; a_{k})
-
\sum_{s\in\mathcal S}\bar b_{k,\theta^*}(s)\,
\mathbb{P}_{\theta^*}(o,o'\mid s,\; a_{k})
\Bigg|
\Bigg)^{2}\\
\le
&\frac{8}{\pi_{\min}}
\left(
\ell_n(\theta^*)-\ell_n(\hat\theta)
+2\log(2/\delta)
\right)+4,
\end{aligned}
\end{equation}
where
$
\mathbb{P}_{\theta}(o,o'\mid s,\; a_{k}=a)
=
O_{\varphi}(o\mid s)\sum_{s'\in\mathcal S}T^{a}_{\beta}(s'\mid s)\,O_{\varphi}(o'\mid s')$.

\end{lemma}
\begin{proof}
    See Appendix~\ref{apx: proof for 2-step mismatch thm}.
\end{proof}

\begin{remark}\label{remark: extending pointwise results}
Lemmas~\ref{lem:quadratic bound for O} and~\ref{lem:length-2 quadratic bound for O} provide pointwise guarantees. Their uniform counterparts are given in Appendices~\ref{apx: proof for universal quadratic bound for O} and~\ref{apx: global generalization of length-2 quadratic bound}, respectively.
\end{remark}

Lemma~\ref{lem:quadratic bound for O} and \ref{lem:length-2 quadratic bound for O} relate likelihood gaps to the cumulative discrepancies between one- and two-step predictive observation distributions. We next translate these bounds into recovery guarantees for the underlying POMDP kernels. This step requires identifiability, as distinct kernels can otherwise induce identical distributions over observed trajectories.

For $A\in\mathbb{R}^{m\times n}$ let $\sigma_{\min}(A)$ be the $\min\{m,n\}$-th largest singular value of $A$.

\begin{assumption}
\label{ass:alpha_undercomplete}
There exists $\sigma_{O} > 0$ such that
$
\sigma_{\min}(O_{\varphi^*}) \ge \sigma_{O}.
$
\end{assumption}
This is the \(\alpha\)-weakly revealing condition of \citet{liu2022partially}, with $\alpha = \sigma_{O}$. It ensures that $O_{\varphi^*}$ has full column rank and admits a well-conditioned left inverse, which is possible only when $|\mathcal S|\leq|\Omega|$.
Consequently, distinct latent-state distributions induce distinct observation distributions. Our analysis additionally relies on a nondegeneracy condition on the transition kernel, introduced next.

\begin{assumption}\label{ass: full-rank transition kernels}
There exists $\sigma_{T} > 0$ such that, for every $a\in\mathcal{A}$, 
$
\sigma_{\min}(T^{a}_{\beta^*}) \ge \sigma_{T}.
$
\end{assumption}
\begin{remark}\label{remark: extending the full-rank assumption of T}
Assumption~\ref{ass: full-rank transition kernels} can be relaxed to an aggregate conditioning requirement on the transition kernel. For clarify, we retain Assumption~\ref{ass: full-rank transition kernels} in the main text and present the relaxed condition and corresponding proof modifications in Appendix~\ref{apx: replacing the full-rank transition kernels assumption}.
\end{remark}

We next investigate how closeness of predictive distributions translates into recovery guarantees for the underlying POMDP kernels. To this end, we collect the predictive beliefs along the trajectory into empirical belief matrices. Define
$
\bar{B}:=\big[\bar{b}_{1,\theta^*}\ \cdots\ \bar{b}_{n,\theta^*}\big]^\top$, $\widehat B:=\big[\bar{b}_{1,\hat{\theta}}\ \cdots\ \bar{b}_{n,\hat{\theta}}\big]^\top$. Furthermore, for each action \(a\in\mathcal A\), let $\mathcal{I}_{a}\coloneqq\{t|a_{t}=a\}$, $\mathcal{N}_{a}\coloneqq|\mathcal{I}_{a}|$, and define the corresponding action-restricted belief matrices
$\bar{B}_{a}:=\big[\bar{b}_{\mathcal{I}_{a}(1),\theta^*},\ \cdots\ ,\bar{b}_{\mathcal{I}_{a}(\mathcal{N}_{a}),\theta^*}\big]^\top$, $\widehat{B}_{a}:=\big[\bar{b}_{\mathcal{I}_{a}(1),\hat{\theta}},\ \cdots\ ,\bar{b}_{\mathcal{I}_{a}(\mathcal{N}_{a}),\hat{\theta}}\big]^\top$.

Well-conditioned left inverses for $\bar{B}$ and $\bar{B}_{a}$ ensures that the predictive beliefs are sufficiently informative to establish kernel-recovery bounds from discrepancies in the one-step and two-step predictive distributions. Rather than imposing these conditioning requirements as separate assumptions, we derive them from two properties of the data-generating process: uniform action coverage and ergodicity of the policy-induced latent-state Markov chain. Together with the preceding structural assumptions, these properties ensure sufficient state-action coverage and, after burn-in, well-conditioned belief matrices with high probability. We define ergodicity in the following assumption.

\begin{assumption}\label{ass: ergodicity}
    The Markov chain induced by $\pi$ with the transition kernel
$
\mathbb{P}_{\theta^*}^{\pi}(s' \mid s)
=
\sum_{o \in \Omega} O_{\varphi^*}(o \mid s)\sum_{a \in \mathcal{A}} \pi(a \mid o)\,T^{a}_{\beta^*}(s' \mid s)$ is ergodic.
Let $d^{\pi}$ denote the unique stationary distribution of this chain, and define
$
d_{\min}^{\pi} := \min_{s \in \mathcal{S}} d^{\pi}(s) > 0$.
\end{assumption}

Assumption~\ref{ass: ergodicity} ensures geometric convergence of the policy-induced latent-state Markov chain to stationarity. We state the corresponding mixing bound below (see \citet{kontorovich2014uniform}).

\begin{corollary}\label{cor: geometric mixing}
    Let
$
\|\mu-\nu\|_{\mathrm{TV}}
=
\sup_{A \subseteq \mathcal{S}} |\mu(A)-\nu(A)|$.
    Then, there exist constants $1 \le G(\pi) < \infty$ and $0 < \rho(\pi) < 1$ such that for every initial state $s \in \mathcal{S}$ and every integer $t \ge 1$,
\begin{equation}\label{equ:mixing definition}
\left\|
\mathbb{P}^{\pi}_{\theta^*}(s_t \in \cdot \mid s_1=s) - d^{\pi}
\right\|_{\mathrm{TV}}
\le 
G(\pi)\,\rho(\pi)^{\,t-1},
\end{equation}
where $G(\pi)$ and $\rho(\pi)$ characterize geometric mixing of the policy-induced Markov chain.
\end{corollary}

We discuss the above assumptions and their practical implications in Appendix~\ref{apx: discussion of the assumptions}, including which can be relaxed and which are technical artifacts of the analysis rather than intrinsic to the setting.

We now combine the structural, coverage, and mixing conditions to establish quantitative conditioning of the action-restricted predictive-belief matrices. Assumptions~\ref{ass:alpha_undercomplete} and \ref{ass: full-rank transition kernels} provide separation and invertibility, while Assumption~\ref{ass: minimum policy mass} ensures action coverage and Assumption~\ref{ass: ergodicity} together with Corollary~\ref{cor: geometric mixing} guarantees state coverage and geometric mixing. These conditions yield the following finite-sample lower bound on the smallest singular value of these matrices.

\begin{lemma}[Informal]\label{lem: sigma B_a proof}
Suppose that Assumptions~\ref{ass: minimum policy mass}, \ref{ass:alpha_undercomplete}, \ref{ass: full-rank transition kernels}, and \ref{ass: ergodicity} hold. For any
\(\delta\in(0,1)\), if $n$ is bounded below by a universal polynomial with nonnegative exponents in
\[
|\mathcal S|,\ \ |\Omega|,\ 
\frac{1}{\pi_{\min}},\ 
\frac{1}{d_{\min}^{\pi}},\ 
\frac{1}{\sigma_T},\ 
\frac{1}{\sigma_O},\
\log(1/\delta),\
\log(G(\pi)),\ 
\frac{1}{-\log(\rho(\pi))},\]
then with probability at least $1-\delta$ there exists a universal constant $c$ such that for all $a\in\mathcal{A}$,
\begin{align*}
\sigma_{\min}(\bar{B}_{a})^{2}\geq c (n-t_{\mathrm{burn}})|\mathcal A|\pi_{\min}^{2}
\sigma_T^{2}\sigma_O^{2}(d_{\min}^{\pi})^{2}.
\end{align*}

With sufficient post-burn-in samples, the action-restricted predictive-belief matrices are uniformly well conditioned with high probability, with smallest singular values scaling optimally as $\sqrt{n}$.
\end{lemma}

\begin{proof}
See Appendix~\ref{apx: proof for lem sigma B_a} for the formal lemma statement and proof.
\end{proof}

Lemmas~\ref{lem:quadratic bound for O} and~\ref{lem:length-2 quadratic bound for O} bound discrepancies between candidate and ground-truth predictive distributions via the likelihood gap, while
Lemma~\ref{lem: sigma B_a proof} establishes quantitative conditioning of the action-restricted predictive-belief matrices. Together, they yield a finite-sample POMDP kernel-recovery guarantee with explicit dependence on the likelihood gap, kernel conditioning, sample size, and behavior policy.

\begin{theorem}\label{thm: approximate permutation for undercomplete POMDPs}
Given any parameter $\hat{\theta}$, suppose Assumptions~\ref{ass: minimum policy mass}, \ref{ass:alpha_undercomplete}, \ref{ass: full-rank transition kernels}, and \ref{ass: ergodicity} hold.

Fix a confidence level $\delta\in(0,1)$
and define $
d=(|\mathcal A||\mathcal S|+1)(|\mathcal S|-1)
+|\mathcal S|(|\Omega|-1)$,
$H_n(\delta)
=
G_n+2d\log(2|\Omega|n)+2\log(4/\delta)+2$. If $n$ is bounded below by a universal polynomial,
with nonnegative exponents, in
\begin{equation}\label{equ:relevent parameters}
|\mathcal{A}|,\ 
\frac{1}{\pi_{\min}},\ 
\frac{1}{d_{\min}^{\pi}},\ 
\frac{1}{\sigma_T},\ 
\frac{1}{\sigma_O},\ |\mathcal S|,\  |\Omega|,\
H_n(\delta),\
\log(G(\pi)),\ 
\frac{1}{-\log(\rho(\pi))},\ 
\end{equation}
then with probability at least $1-\delta$,
the following holds simultaneously for all
$\hat\theta=(\hat\beta,\hat\varphi)\in\Theta$
satisfying
$\ell_n(\theta^*)-\ell_n(\hat\theta)\le G_n$:
there exists a permutation matrix $P$ such that
\[
\left\|O_{\varphi^*}P-O_{\hat{\varphi}}\right\|_{1,\mathrm{ent}}
\leq
C_O \sqrt{\frac{H_n(\delta)}{ n}},\qquad
\left\|P^\top T_{\beta^*}^{a}P-T_{\hat{\beta}}^{a}\right\|_{1,\mathrm{ent}}
\leq
C_T \sqrt{\frac{H_n(\delta)}{ n}},
\]
for every \(a\in\mathcal A\), where for a matrix $A$, we define $\|A\|_{1,\mathrm{ent}}\coloneqq \sum_{i}\sum_{j}|A_{ij}|$.
Here \(C_O,C_T\) are problem-dependent constants that depend polynomially on the first seven quantities in~\eqref{equ:relevent parameters}.

In particular, a sufficiently small likelihood gap guarantees accurate recovery of the POMDP kernels, up to a common permutation of the latent states and uniformly over the parameter space, with high probability.
\end{theorem}

\begin{proof}
    See Appendix~\ref{apx: proof of approximate permutation} for the formal theorem statement and proof.
\end{proof}

With observed rewards, the same latent-state permutation aligns the reward kernel, as follows.

\begin{corollary}
Under the assumptions of Theorem~\ref{thm: approximate permutation for undercomplete POMDPs}, suppose additionally
that rewards are observed, take values in a finite alphabet, and are conditionally independent of observations given the state and action. Assume further that $\hat{\theta}$ satisfies the corresponding
likelihood-gap condition for the joint
observation--action--reward data. Then, with probability at least $1-\delta$, there exists a permutation matrix $P$ satisfying the conclusions of Theorem~\ref{thm: approximate permutation for undercomplete POMDPs} and the additional bound
\[
\max_{a\in\mathcal A}
\|R_{\theta^*}^aP-R_{\hat{\theta}}^a\|_{1,\mathrm{ent}}
    \le C_R\sqrt{\frac{H_n(\delta)}{n}},
\]
where $R^{a}_{\theta^*}$ and $R^{a}_{\hat{\theta}}$ denote the true and estimated conditional reward kernels for action $a$, respectively. The constant $C_R$ depends on the problem parameters analogously to $C_O$ and $C_T$ as well as the to size of the reward alphabet.
\end{corollary}

\begin{proof}
    See Appendix~\ref{apx: short reward-alignment corollary}.
\end{proof}

Thus, the observation and transition kernels, together with the relevant reward quantities, are recovered up to a common permutation of the latent states, providing a consistent model for planning.

Furthermore, our recovery guarantee does not require optimization over the unrestricted POMDP class. In particular, prior structural knowledge can be incorporated by restricting the candidate model class, provided that the true model remains feasible. This yields the following immediate corollary.

\begin{corollary}
Let $\Theta_{\mathrm{str}}\subseteq \Theta$ be any structured POMDP class containing the true parameter $\theta^*$, where $\Theta$ is the unrestricted model class. Then any $\hat{\theta} \in\Theta_{\mathrm{str}}$ satisfying the likelihood-gap condition of Theorem~\ref{thm: approximate permutation for undercomplete POMDPs} enjoys the same likelihood-to-parameter recovery guarantee.
\end{corollary}

Thus, known structural constraints may be imposed on the candidate class without modifying our recovery analysis (see Appendix~\ref{apx: example for constrained model class} for an example).

\subsection{Overcomplete POMDPs \texorpdfstring{(\(|\mathcal S|> |\Omega|\))}{(|S| > |Omega|)}}\label{subsec: main result-overcomplete}
For overcomplete POMDPs, we extend the framework in Section~\ref{subsec: main result-undercomplete} by replacing one-step observability with its multi-step analogue. The main proof strategy, including likelihood decomposition, the relationship between likelihood gaps and predictive-distribution discrepancies, and predictive-belief conditioning, remains applicable. We therefore focus on modified assumptions and present streamlined bounds, suppressing polynomial dependencies analogous to those in Section~\ref{subsec: main result-undercomplete}.

Assumption~\ref{ass:alpha_undercomplete} cannot hold in the overcomplete regime, since \(O_{\varphi^*}\in\mathbb R^{|\Omega|\times|\mathcal S|}\) cannot have full column
rank when \(|\mathcal S|>|\Omega|\).

For $m \in \mathbb{N}$ and a prescribed action sequence $w\in\mathcal{A}^{m-1}$, define the corresponding $m$-step observation kernel
$[\mathcal{M}^{(m)}_{\theta}[w]]$ under $\theta=(\beta,\varphi)$:  $[\mathcal{M}^{(m)}_{\theta}[w]](z,s)=\mathbb{P}_\theta
\big(
o_{1:m}=z
\mid
s_1 = s,\,
a_{1:m-1}=w
\big)$, for $z \in \Omega^{m}$ 
and $s \in \mathcal S$. This leads to the following multi-step identifiability assumption.

\begin{assumption}
\label{ass:multi_step_revelation}
There exist $m\in\mathbb N$ and $\sigma_M>0$ such that the set
$
\mathcal W
:=
\{
w\in\mathcal A^{m-1}:
\sigma_{\min}(\mathcal M_{\theta^*}^{(m)}[w])\ge \sigma_M
\}
$
is nonempty.
\end{assumption}

In the rest of Section \ref{subsec: main result-overcomplete}, we assume that Assumption~\ref{ass:multi_step_revelation} holds for a fixed $m$. It requires a prescribed action sequence \(w\in\mathcal{A}^{m-1}\) for which \(\mathcal M_{\theta^*}^{(m)}[w]\) admits a well-conditioned left inverse. Since \(\mathcal M_{\theta^*}^{(m)}[w]\in\mathbb R^{|\Omega|^m\times|\mathcal S|}\), this requires \(|\Omega|^m\ge|\mathcal S|\). Under such an action sequence, distinct latent-state distributions induce discint distributions over length-$m$ observation blocks, even if indistinguishable from one-step observations. When \(m=1\), this reduces to Assumption~\ref{ass:alpha_undercomplete}.

Let \(\oplus\) denote concatenation of action strings, and define \(\mathcal A\oplus\mathcal A\oplus\mathcal W\oplus\mathcal A\) as the set of all strings \(a_1\oplus a_2\oplus w\oplus a'\), where \(a_1,a_2,a'\in\mathcal A\) and \(w\in\mathcal W\). We formalize the behavior policy below.

\begin{assumption}\label{ass: L-step policy coverage}
$\pi$ is blockwise: at the start of each block, it samples a string
$(a_1,a_2,w,a')\in\mathcal A\oplus\mathcal A\oplus\mathcal W\oplus\mathcal A$
from $\pi$ and executes it throughout the block.
Moreover, there exists $\pi_{\min}^{\mathrm{blk}}>0$ such that for every $a_1\in\mathcal A$, there exists
$a_2\in\mathcal A$ such that the set
$
\mathcal W_{\bar a}
:=\left\{
w\in\mathcal W:
\sum_{a'\in\mathcal A}
\pi(a_1\oplus a_2\oplus w\oplus a')
\geq\pi_{\min}^{\mathrm{blk}}
\right\}$
is nonempty, where we denote $(a_1,a_2)$ as $\bar a$.
\end{assumption}

Assumption~\ref{ass: L-step policy coverage} ensures that, for every \(a_1\in\mathcal A\), there exist \(a_2\in\mathcal A\) and \(w\in\mathcal W\) such that the behavior policy samples a block beginning with \(a_1\oplus a_2\oplus w\) with probability at least \(\pi_{\min}^{\mathrm{blk}}>0\).
 
\begin{remark}
Assumption~\ref{ass: L-step policy coverage} can be relaxed to allow action strings outside $\mathcal W$, provided the required informative strings are sampled with probability bounded below by a positive constant. We retain the stronger assumption for simplicity of exposition and proof.
\end{remark}

The following proposition decomposes the trajectory likelihood into consecutive blocks of possibly different lengths, expressing each factor using the multi-step observation kernels introduced above.

\begin{proposition}\label{prop:factorization_general}
For any $\theta=(\beta,\varphi)$, let
$
1=\tau_1<\tau_2<\cdots<\tau_{N+1}=n+1
$
be a sequence of cut points, and define $m_i=\tau_{i+1}-\tau_i$ for each
$i\in[N]$. Then
\[
\ell_n(\theta)
=
\sum_{i=1}^{N}
\log\left(
\sum_{s\in\mathcal S}
\bar b_{\tau_i,\theta}(s)
\left[
\mathcal M_{\theta}^{(m_i)}
\!\left[a_{\tau_i:\tau_{i+1}-2}\right]
\right](o_{\tau_i:\tau_{i+1}-1},\,s)
\right).
\]
Here, when $m_i=1$, the action sequence $a_{\tau_i:\tau_{i+1}-2}$ is understood
as the empty sequence.
\end{proposition}

\begin{proof}
    The result follows by the same argument used to derive the two-step factorization in Proposition~\ref{prop:length-2 factorization} from the one-step factorization in Proposition~\ref{prop:factorization}. Specifically, grouping consecutive one-step likelihood factors into blocks of lengths $m_i$ yields the claimed expression, with each block represented by its corresponding $m_i$-step observation kernel.
\end{proof}

We now extend Lemmas~\ref{lem:quadratic bound for O} and~\ref{lem:length-2 quadratic bound for O} to multi-step observation blocks.

\begin{lemma}\label{lem:quadratic bound_block predictive distributions}
For each $i\in[N]$, let $t_i=1+(i-1)(m+2)$ and define the
block predictive distribution
$
p_{t_i,\theta}(x)
:=\sum_{s\in\mathcal S}\bar{b}_{t_i,\theta}(s)
\left[\mathcal M_\theta^{(m+2)}[a_{t_i:t_i+m}]\right](x,s)$, $x\in\Omega^{m+2},$
where $N=\lfloor n/(m+2)\rfloor$.
Then, given a candidate parameter
$\hat\theta=(\hat\beta,\hat\varphi)$,
for any $\delta\in(0,1)$, with probability at least
$1-\delta$,
\begin{equation*}
\sum_{i=1}^{N}
\|p_{t_i,\hat\theta}-p_{t_i,\theta^*}\|_1^2
\leq
\frac{8}{\pi_{\min}^{\mathrm{blk}}}
\left(
\ell_n(\theta^*)-\ell_n(\hat\theta)
+2\log(2/\delta)
\right)+4.
\end{equation*}

\end{lemma}
\begin{proof}
    The claim follows by applying the argument of Lemma~\ref{lem:length-2 quadratic bound for O} to the blockwise likelihood factorization in Proposition~\ref{prop:factorization_general}.
\end{proof}

Under Assumption~\ref{ass: L-step policy coverage}, the blockwise behavior policy induces a Markov chain on the latent states at block boundaries, $(s_{1+k(m+2)})_{k\ge 0}$. We impose the following ergodicity condition on this chain.

\begin{assumption}\label{ass: L-step ergodicity}
    The block-level Markov chain induced by $\pi$ is ergodic with stationary distribution $d_{\mathrm{blk}}^{\pi}$. Define
$(d_{\mathrm{blk}}^{\pi})_{\min} := \min_{s \in \mathcal{S}} d_{\mathrm{blk}}^{\pi}(s) > 0$, and let $G_{\mathrm{blk}}(\pi)$ and $\rho_{\mathrm{blk}}(\pi)$ denote its geometric mixing constants, defined analogously to those in Corollary~\ref{cor: geometric mixing}.
\end{assumption}

\begin{remark}\label{remark: block-level policy assumptions}
    Assumptions~\ref{ass: L-step policy coverage} and~\ref{ass: L-step ergodicity} are sufficient, rather than minimal, conditions on $\pi$. We impose them to simplify the analysis.
\end{remark}

To identify POMDP kernels up to a permutation of the latent states, we impose an additional condition requiring distinct latent states to induce distinct observation distributions.
\begin{assumption}
\label{ass:distinct-observation-columns}
We observation distributions of distinct latent states satisfy
    $\Delta_{O}\coloneqq\min_{s\neq s'}\|O_{\varphi^*}(\cdot\mid s)- O_{\varphi^*}(\cdot\mid s')\|> 0.$
\end{assumption}
Appendix~\ref{apx: justification for distinct observation columns} shows that, without this assumption, kernel recovery can be information-theoretically impossible in general overcomplete POMDPs.

A more detailed discussion of Assumptions~\ref{ass:multi_step_revelation}, \ref{ass: L-step policy coverage}, \ref{ass: L-step ergodicity}, and \ref{ass:distinct-observation-columns} is provided in Appendix~\ref{apx: discussion of the assumptions}.

We next investigate the conditioning of predictive-belief matrices. To this end, we introduce their blockwise counterparts. Let $N=\lfloor n/(m+2)\rfloor$, $t_i=1+(i-1)(m+2)$, and
$
\bar b^{\mathrm{blk}}_{i,\theta}:=\bar b_{t_i,\theta}$, $i\in[N]$.
Define
$
\bar B^{\mathrm{blk}}
:=
[
\bar b^{\mathrm{blk}}_{1,\theta^*}\ \cdots\
\bar b^{\mathrm{blk}}_{N,\theta^*}
]^\top$ and
$\widehat B^{\mathrm{blk}}
:=
[
\bar b^{\mathrm{blk}}_{1,\hat\theta}\ \cdots\
\bar b^{\mathrm{blk}}_{N,\hat\theta}
]^\top$. For each $\tilde{a}=(a_{1},a_{2})\in\mathcal{A} \oplus \mathcal{A}$ and $w\in\mathcal{W}$, let
$\mathcal I^{\mathrm{blk}}_{\tilde{a},w}:=\{i\in[k]:(a_{t_i},a_{t_i+1},\ldots,a_{t_i+m})=(a_{1},a_{2},w)\}$,
$N^{\mathrm{blk}}_{\tilde{a},w}:=|\mathcal I^{\mathrm{blk}}_{\tilde{a},w}|$, and 
$\bar B^{\mathrm{blk}}_{\tilde{a},w}:=
[\bar b^{\mathrm{blk}}_{\mathcal I^{\mathrm{blk}}_{\tilde{a},w}(1),\theta^*}
\ \cdots\
\bar b^{\mathrm{blk}}_{\mathcal I^{\mathrm{blk}}_{\tilde{a},w}(N^{\mathrm{blk}}_{\tilde{a},w}),\theta^*}]^\top$. The matrix $\widehat{B}^{\mathrm{blk}}_{\tilde{a},w}$ is defined analogously.

The conditioning of the block-level matrices is central to our subsequent analysis. However, the assumptions introduced thus far are insufficient to guarantee polynomial sample complexity for kernel recovery, as demonstrated by the following proposition.

\begin{proposition}
\label{prop:exponential-lower-bound}
There exist overcomplete POMDPs satisfying Assumptions~\ref{ass: full-rank transition kernels}, \ref{ass:multi_step_revelation}, \ref{ass: L-step policy coverage}, \ref{ass: L-step ergodicity}, \ref{ass:distinct-observation-columns}, such that learning
their transition and observation kernels to accuracy $O(1/|\mathcal{S}|)$, up to a
common state permutation, requires $\exp(\Omega(|\mathcal{S}|))$ observations.
\end{proposition}
\begin{proof}
    See Appendix~\ref{apx: example for exponential lower bound}.
\end{proof}

The pathological instances underlying Proposition~\ref{prop:exponential-lower-bound} exhibit exponentially ill-conditioned backward observability: certain latent-state directions are nearly indistinguishable from past observation histories. To rule out such instances, let $\rho$ be a distribution over $\mathcal{A}^L$ and define
\[
\Phi_{L}^{\rho}(x)
:=
\sum_{\boldsymbol a}\rho(\boldsymbol a)
\sum_{y\in\Omega^{L}}
\left|
x^\top K_{\boldsymbol a,y}d_{\mathrm{blk}}^\pi
\right|,
\qquad
K_{\boldsymbol a,y}
:=
T_{\beta^*}^{a_{L}}D_{o_{L}}
\cdots T_{\beta^*}^{a_1}D_{o_1},
\]
where $\boldsymbol a\in\mathcal{A}^{L}\sim\rho$ is a block
of $L$ actions, $y=(o_1,\ldots,o_{L})$, and
$D_o=\operatorname{diag}(O_{\varphi^*}(o\mid\cdot))$.
The quantity $\Phi_L^\rho(x)$ measures how strongly length-$L$ observation histories excite a latent-state direction $x$ when the corresponding action blocks are sampled according to $\rho$.

For $L=qH$, where $H=m+2$, let
$
\rho_L^\pi(\boldsymbol a)
:=
\prod_{j=0}^{q-1}
\pi(a_{jH+1},\ldots,a_{(j+1)H})$,
denote the length-$L$ action-string distribution induced
by $\pi$. We write
$\Phi_L^\pi:=\Phi_L^{\rho_L^\pi}$.
We may restrict to $L\in H\mathbb N$, since
$\Phi_L^\pi(x)$ is nondecreasing in $L$
(see Appendix~\ref{apx: quantitative backward sufficient and necessary}). Then, we impose the following assumption.

\begin{assumption}[Quantitative backward excitation]
\label{ass:backward-excitation}
For fixed universal constants $c,C>0$, there exists a
horizon $L\in(m+2)\mathbb N$ with $L\le|\mathcal S|^C$
such that
$
\Phi_L^\pi(x)\ge c|\mathcal S|^{-C}\|x\|_2$, $\forall x\in\mathbb R^{|\mathcal S|}.
$
\end{assumption}

We discuss the assumption further in Appendix~\ref{apx: discussion of the assumptions}. The following proposition establishes its sufficiency for polynomial sample complexity in kernel recovery and shows that, without it, the remaining assumptions permit instances with exponential sample complexity.

\begin{proposition}
\label{prop:backward-excitation-complexity}
Suppose Assumptions~\ref{ass: full-rank transition kernels}, \ref{ass:multi_step_revelation}, \ref{ass: L-step policy coverage}, \ref{ass: L-step ergodicity}, \ref{ass:distinct-observation-columns}, and \ref{ass:backward-excitation} hold, with all relevant model parameters bounded by \(\operatorname{poly}(|\mathcal S|)\). Then the kernels can be estimated,
up to a common permutation of the latent states,
to accuracy $\varepsilon$ with probability at least
$1-\delta$ using a trajectory of length
$\operatorname{poly}(|\mathcal{S}|,\varepsilon^{-1},\log(1/\delta))$. Conversely, without
Assumption~\ref{ass:backward-excitation}, the remaining assumptions admit
instances requiring $\exp(\Omega(|\mathcal{S}|))$ observations
for any estimator, even at accuracy
$\varepsilon=\Theta(|\mathcal{S}|^{-1})$ and constant confidence.
\end{proposition}
\begin{proof}
    See Appendix~\ref{apx: quantitative backward sufficient and necessary}.
\end{proof}

We can now establish the quantitative conditioning of the block-level predictive-belief matrices.

\begin{lemma}\label{lem: sigma B_a proof_overcomplete}
Suppose Assumption~\ref{ass: L-step ergodicity} holds for $\pi$, Assumption~\ref{ass:backward-excitation} holds for $\pi$
with parameters $L$ and $\eta$, and Assumption~\ref{ass: L-step policy coverage} holds. Fix a confidence level
$\delta\in(0,1)$. If $n$ is bounded below by a universal polynomial
with nonnegative powers in
$
|\mathcal S|, \ |\mathcal{A}|,\ \frac{1}{\pi^{\mathrm{blk}}_{\min}}, \
\log(1/\delta),\
\log(G_{\mathrm{blk}}(\pi)),\
\frac{1}{-\log(\rho_{\mathrm{blk}}(\pi))}, \ L,\
1/\eta,$
then, with probability at least $1-\delta$, there exists a universal constant $c^{\mathrm{blk}}$
such that for every $a_{1}\in\mathcal{A}$, there exists $a_{2}\in\mathcal{A}$ and $w\in\mathcal{W}^{\pi}_{\tilde{a}}$, where $\tilde{a}=(a_{1},a_{2})$, satisfying 
$\sigma_{\min}^{2}(\bar B^{\mathrm{blk}}_{\tilde{a},w})\ge c^{\mathrm{blk}}\pi^{\mathrm{blk}}_{\min}\frac{\eta^2}{L}n$.
\end{lemma}

\begin{proof}
See Appendix~\ref{apx: justification for overcomplete predictive belief full-rankedness}.
\end{proof}

We now have the necessary ingredients to extend the kernel-recovery guarantee of Theorem~\ref{thm: approximate permutation for undercomplete POMDPs} to overcomplete POMDPs.

\begin{theorem}\label{thm: approximate permutation for overcomplete POMDPs}
Suppose Assumptions~\ref{ass: full-rank transition kernels}, \ref{ass:multi_step_revelation}, \ref{ass: L-step policy coverage}, \ref{ass: L-step ergodicity}, \ref{ass:distinct-observation-columns}, and \ref{ass:backward-excitation} hold. Fix a confidence level
\(\delta\in(0,1)\), and define $
d=(|\mathcal A||\mathcal S|+1)(|\mathcal S|-1)
+|\mathcal S|(|\Omega|-1)$,
$H_n(\delta)
=
G_n+2d\log(2|\Omega|n)+2\log(4/\delta)+2$.
If \(n\) is bounded below by a universal polynomial, with nonnegative powers, in
\begin{equation}\label{equ:relevent parameters_overcomplete}
|\mathcal{A}|,\ 
\frac{1}{\pi_{\min}^{\mathrm{blk}}},\ 
\frac{1}{(d_{\min}^{\pi})^{\mathrm{blk}}},\ 
\frac{1}{\sigma_T},\ 
\frac{1}{\sigma_M},\ \frac{1}{\Delta_{O}}, \ |\mathcal S|,\  |\Omega|,\ \frac{L}{\eta^{2}},\
H_n(\delta),\ 
\log(G_{\mathrm{blk}}(\pi)),\ 
\frac{1}{-\log(\rho_{\mathrm{blk}}(\pi))},
\end{equation}
then with probability at least \(1-\delta\), the following holds simultaneously for all
$\hat\theta=(\hat\beta,\hat\varphi)\in\Theta$
satisfying
$\ell_n(\theta^*)-\ell_n(\hat\theta)\le G_n$: there exists a permutation matrix \(P\) such that for every $a\in\mathcal{A}$,
\[
\left\|O_{\varphi^*}P-O_{\hat{\varphi}}\right\|_{1,\mathrm{ent}}
=O\left(\sqrt{\frac{H_n(\delta)}{n}}\right),\qquad
\left\|P^\top T_{\beta^*}^{a}P-T_{\hat{\beta}}^{a}\right\|_{1,\mathrm{ent}}
=O\left(\sqrt{\frac{H_n(\delta)}{n}}\right),
\]
where the constants in $O(\cdot)$ are bounded by a universal polynomial in the first nine quantities in~\eqref{equ:relevent parameters_overcomplete}.
\end{theorem}

\begin{proof}
    See Appendix~\ref{apx: proof of approximate permutation_overcomplete}.
\end{proof}

Thus, despite the failure of one-step observability in overcomplete POMDPs, likelihood-based kernel recovery remains polynomially sample-efficient under multi-step observability, blockwise coverage and ergodicity, pairwise distinct observation columns, and inverse-polynomial distinguishability of latent-state directions from past observations.

\section{Discussion}\label{sec: discussion}
In this work, we establish finite-sample guarantees that translate likelihood gaps into parameter recovery error bounds. By combining trajectory likelihood decompositions with quantitative conditioning bounds for predictive-belief matrices, our analysis covers both undercomplete and overcomplete POMDPs. To the best of our knowledge, such finite-sample guarantees have not previously been established in the likelihood-based literature, while existing spectral methods have not established finite-sample guarantees for full kernel recovery in the overcomplete regime. Furthermore, our guarantees apply to any parameter with a controlled likelihood gap, regardless of how it is obtained. Thus, they can be combined with POMDP learning algorithms that produce such estimates.

\section{Limitations and future work}\label{sec: future work}
Our analysis adapts the observation-kernel assumptions from the undercomplete to the overcomplete regime, while retaining the same full-rank condition on the transition kernels in both. Investigating whether this transition-kernel condition can be weakened is an important direction for future work.

As noted in Remark~\ref{remark: block-level policy assumptions}, Assumptions~\ref{ass: L-step policy coverage} and~\ref{ass: L-step ergodicity} are sufficient, rather than minimal, conditions for extending the undercomplete analysis to the overcomplete regime. Characterizing weaker behavior-policy assumptions under which likelihood gaps still yield parameter recovery guarantees remains an important direction for future work. This is particularly relevant in the overcomplete setting, where the behavior policy plays a more active role: by selecting informative action sequences, it can help distinguish latent-state mixtures that cannot be separated through one-step observations alone.

While our theoretical results identify key properties of effective behavior policies, they do not provide a tractable procedure for selecting them. In practice, transition and observation kernels are unknown, motivating robust policy design over uncertainty sets of plausible models. Developing efficient algorithms that balance action coverage, stationary state coverage, and mixing, while accounting for informative action blocks in the overcomplete regime, remains an important direction for future work.

A separate practical consideration arises when latent states have semantic labels and a known, externally supplied reward kernel is tied to those labels. Since our results recover the model only up to a permutation of the latent states, the recovered kernels must be aligned with the external labeling. Such alignment is possible when states have known or approximately known transition or observation signatures, such as distinctive sensor patterns or transition behavior, that allow the recovered states to be matched to their semantic labels. Exact signatures are not required; sufficiently separated signatures or observations strongly associated with particular states may suffice.

Although Section~\ref{sec: introduction} highlights the benefits of POMDP kernel recovery for latent-state inference, forecasting, simulation, and downstream tasks, it remains unclear whether full recovery is necessary for near-optimal policy learning in infinite-horizon POMDPs. Existing guarantees in the infinite-horizon setting typically rely on accurate estimation of the transition and observation kernels. If full recovery is necessary in this setting, it would be valuable to characterize subclasses where weaker forms of model identification suffice; otherwise, an important direction is to develop near-optimal algorithms that circumvent full kernel recovery for infinite-horizon POMDPs.

\begin{ack}
This research was partially supported by the AFOSR grant FA9550-23-1-0190.

The authors declare no competing interests.
\end{ack}

\bibliographystyle{plainnat}
\bibliography{references}


\appendix

\section{}\label{apx:reusable model for downstream reward}
\noindent\textbf{Example: AI-operated warehouse-robot fleet.}
Consider an autonomous warehouse in which mobile robots repeatedly transport inventory while their drivetrain condition is only partially observed. At each maintenance epoch, the latent condition (state) of a robot is
\[
\mathcal{S} = \{H,D,C\},
\]
where \(H\) denotes healthy, \(D\) denotes degraded but operational, and \(C\) denotes maintenance-critical. The controller chooses among
\[
\mathcal{A} = \{a_{\mathrm{red}}, a_{\mathrm{high}}, a_{\mathrm{svc}}\},
\]
corresponding to reduced-speed operation, high-throughput operation, and service/repair. The available telemetry is discretized into
\[
\Omega
=
\{
\text{nominal},
\text{vibration anomaly},
\text{thermal/current anomaly},
\text{ambiguous}
\}.
\]

The transition kernel describes action-dependent deterioration and recovery, while the observation kernel describes the vibration, temperature, motor-current, and diagnostic signatures generated by each health regime. These kernels characterize the robot and its sensing system rather than a particular control objective. By contrast, the reward can be changed to emphasize throughput, energy consumption, safety, delivery deadlines, maintenance cost, or equipment longevity.

To be more specific, because the kernels describe the underlying environment independently of the chosen reward, recovering them once supports a range of downstream objectives and analyses. First, the kernels provide a system-identification model of how operating intensity affects deterioration and how each latent health condition appears in the observable telemetry. Second, the same recovered model can be reused when the downstream reward changes, for example from peak-throughput maximization to energy-efficient operation or safety-critical maintenance. Third, one may perform sensitivity analysis by perturbing wear, repair-success, or sensor-error probabilities; policy-level counterfactual simulation by evaluating alternative service schedules or operating intensities; and model-based off-policy evaluation of a candidate policy. Thus the transition and observation kernels form a reusable description of the environment, whereas the reward specifies a particular downstream task.

\section{}\label{apx:factorization proof}

Recall that $a_1^n$ is the realized action sequence under
the fixed, parameter-independent behavior policy $\pi$.
By the chain rule, the joint trajectory likelihood satisfies
\[
    \mathbb{P}_\theta(o_{1:n},a_{1:n})
    = \prod_{i=1}^n
    \mathbb{P}_\theta(o_i \mid o_{1:i-1},a_{1:i-1})
    \pi(a_i \mid o_i).
\]

Define $f_{\theta}(o_{1:n},a_{1:n})$ as the model-dependent component of the observed-data likelihood. Then, since the behavior policy $\pi$ is independent of $\theta$, we have
\begin{equation}\label{equ: factorization}
    f_\theta(o_{1:n},a_{1:n})
    = \prod_{i=1}^n
    \mathbb{P}_\theta(o_i \mid o_{1:i-1},a_{1:i-1}).
\end{equation}

Next, condition on the latent state $s_i$ and marginalize:
\begin{equation}\label{eq:product-rule}
\begin{aligned}
\mathbb{P}_{\theta}(o_i \mid o_{1:i-1}, a_{1:i-1})
&=
\sum_{s}
\mathbb{P}_{\theta}(o_i, s_i=s \mid o_{1:i-1}, a_{1:i-1})\\
&=
\sum_{s}
\mathbb{P}_{\theta}(s_i=s \mid o_{1:i-1}, a_{1:i-1})\,
\mathbb{P}_{\theta}(o_i \mid s_i=s, o_{1:i-1}, a_{1:i-1}).
\end{aligned}
\end{equation}
By the conditional independence of emissions given the current state (i.e., $o_i \perp (o_{1:i-1},a_{1:i-1}) \mid s_i$),
\begin{equation}
\label{eq:emission-ci}
\mathbb{P}_{\theta}(o_i \mid s_i=s, o_{1:i-1}, a_{1:i-1})
~=~
\mathbb{P}_{\theta}(o_i \mid s_i=s)
~=~
O_{\varphi}(o_i\mid s).
\end{equation}
Substituting \eqref{eq:emission-ci} into \eqref{eq:product-rule} and using the definition
$\bar b_{i,\theta}(s) \coloneqq \mathbb{P}_{\theta}(s_i=s\mid o_{1:i-1}, a_{1:i-1})$ gives
\begin{equation}
\label{eq:one-step-predictive}
\mathbb{P}_{\theta}(o_i \mid o_{1:i-1}, a_{1:i-1})
~=~
\sum_{s} \bar b_{i,\theta}(s)\, O_{\varphi}(o_i\mid s).
\end{equation}
Combining \eqref{equ: factorization} and \eqref{eq:one-step-predictive} yields
\[
f_{\theta}(o_{1:n}, a_{1:n})
~=~
\prod_{i=1}^n \Big( \sum_{s} \bar b_{i,\theta}(s)\, O_{\varphi}(o_i\mid s) \Big).
\]
Taking logs proves the claimed expression for $\ell_n(\theta)$.

\section{}\label{apx: 2-step factorization proof}
Recall that the model-dependent component of the trajectory likelihood satisfies
\begin{equation*}
    f_\theta(o_{1:n},a_{1:n})
    = \prod_{i=1}^n
    \mathbb{P}_\theta(o_i \mid o_{1:i-1},a_{1:i-1}).
\end{equation*}
For notational convenience, let
\[
    q_i := \mathbb{P}_\theta(o_i \mid o_{1:i-1},a_{1:i-1}).
\]
Recall that from \eqref{eq:one-step-predictive}, we have
\begin{equation*}
    q_i = \sum_{s\in\mathcal{S}}
    \bar{b}_{i,\theta}(s) O_\varphi(o_i\mid s).
\end{equation*}

Next, we express $\bar{b}_{2k,\theta}$ in terms of $\bar{b}_{2k-1,\theta}$.
By Bayes' rule,
\begin{align*}
    &\mathbb{P}_\theta(s_{2k-1}=s \mid
    o_{1:2k-1},a_{1:2k-1}) \notag\\
    &\qquad =
    \frac{\bar{b}_{2k-1,\theta}(s)
    O_\varphi(o_{2k-1}\mid s)}{q_{2k-1}},
\end{align*}
where conditioning on $a_{2k-1}$ provides no additional
information about the latent state given $o_{2k-1}$,
since the behavior policy is observation-based.
Applying the transition kernel and marginalizing
over $s_{2k-1}$ then gives
\begin{align*}
    \bar{b}_{2k,\theta}(s')
    &= \sum_{s\in\mathcal{S}}
    \mathbb{P}_\theta(s_{2k-1}=s \mid
    o_{1:2k-1},a_{1:2k-1})
    T_\beta^{a_{2k-1}}(s'\mid s)
    \notag\\
    &= \frac{1}{q_{2k-1}}
    \sum_{s\in\mathcal{S}}
    \bar{b}_{2k-1,\theta}(s)
    O_\varphi(o_{2k-1}\mid s)
    T_\beta^{a_{2k-1}}(s'\mid s).
\end{align*}

Consequently,
\begin{align*}
    q_{2k-1}q_{2k}
    &= q_{2k-1}
    \sum_{s'\in\mathcal{S}}
    \bar{b}_{2k,\theta}(s')
    O_\varphi(o_{2k}\mid s')
    \notag\\
    &= \sum_{s\in\mathcal{S}}
    \bar{b}_{2k-1,\theta}(s)
    O_\varphi(o_{2k-1}\mid s)
    \sum_{s'\in\mathcal{S}}
    T_\beta^{a_{2k-1}}(s'\mid s)
    O_\varphi(o_{2k}\mid s').
\end{align*}

Finally, since $n$ is even, grouping consecutive
terms in the likelihood decomposition yields
\begin{align*}
    \ell_n(\theta)
    &= \sum_{k=1}^{n/2}
    \log(q_{2k-1}q_{2k}) \\
    &= \sum_{k=1}^{n/2}
    \log\Bigg(
    \sum_{s\in\mathcal{S}}
    \bar{b}_{2k-1,\theta}(s)
    O_\varphi(o_{2k-1}\mid s)
    \sum_{s'\in\mathcal{S}}
    T_\beta^{a_{2k-1}}(s'\mid s)
    O_\varphi(o_{2k}\mid s')
    \Bigg),
\end{align*}
as claimed.

\section{}\label{apx: proof for quadratic bound for O}
First, we acknowledge that Lemma~\ref{lem:quadratic bound for O} is inspired by the high-level argument underlying Propositions~13 and~14 of \citet{liu2022partially}, but the statement and proof are tailored to our predictive-distribution formulation.

Assume we have a series of predictive belief states $\{\bar{b}_{i,\theta^*}\}_{i=1}^{n}$. Then, for each $i\in [n]$, we sample $o_{i}$ according to the distribution $\mathbb{P}^{*}_{i}(o)\coloneqq \sum_{s}\bar{b}_{i,\theta^*}(s)O_{\varphi^*}(o|s)$. Assume we have another series of predictive belief states $\{\bar{b}_{i,\hat{\theta}}\}_{i=1}^{n}$. Then, define $D\coloneqq\{(o_{i},\bar{b}_{i,\theta^*},\bar{b}_{i,\hat{\theta}})\}_{i=1}^{n}$. Similarly, define $\widehat{\mathbb{P}}_{i}(o)\coloneqq \sum_{s}\bar{b}_{i,\hat{\theta}}(s)O_{\hat{\varphi}}(o|s)$. For a certain $o_{i}$, let $\tilde{o}_{i}$ be a ``tangent" observation independently sampled from the same probability distribution $\mathbb{P}^{*}_{i}$. Then, we define $\widetilde{D}\coloneqq \{(\tilde{o}_{i},\bar{b}_{i,\theta^*},\bar{b}_{i,\hat{\theta}})\}_{i=1}^{n}$. Then, we introduce an auxiliary lemma adapted from \citet{liu2022partially}. See Appendix~\ref{apx: proof of adapted lemma 15} for the proof.

\begin{lemma}\label{lemma: adapted lemma 15}
    Let $\ell\coloneqq \ell(o_{i}, \bar{b}_{i,\theta^*},\bar{b}_{i,\hat{\theta}})$ be a real-valued function that maps an observation $o_{i}$ and two predictive belief states $\bar{b}_{i,\theta^*}$ and $\bar{b}_{i,\hat{\theta}}$ to $\mathbb{R}$. Let $L(D)=\sum_{i=1}^{n}\ell(o_{i}, \bar{b}_{i,\theta^*},\bar{b}_{i,\hat{\theta}})$, and $L(\widetilde{D})=\sum_{i=1}^{n}\ell(\tilde{o}_{i},\bar{b}_{i,\theta^*},\bar{b}_{i,\hat{\theta}})$. Then,
    \[\mathbb{E}\Big[\exp{\Big(L(D)-\log\mathbb{E}[\exp(L(\tilde{D}))|D]\Big)}\Big]=1.\]
\end{lemma}

Given Lemma~\ref{lemma: adapted lemma 15}, we define
    \[
\ell_{\hat{\theta}}(o,\bar{b}_{i,\theta^*},\bar{b}_{i,\hat{\theta}}) =
\begin{cases} 
    \frac{1}{2}\log\left(\frac{\sum_{s}\bar{b}_{i,\hat{\theta}}(s)O_{\hat{\varphi}}(o|s)}{\sum_{s}\bar{b}_{i,\theta^*}(s)O_{\varphi^*}(o|s)}\right)=\frac{1}{2}\log\left(\frac{\widehat{\mathbb{P}}_{i}(o)}{\mathbb{P}^{*}_{i}(o)}\right) & \text{if } \mathbb{P}^{*}_{i}(o) \neq 0, \\
    0, & \text{otherwise}.
\end{cases}
\]

By Lemma \ref{lemma: adapted lemma 15} and Markov's inequality, with probability at least \(1 - \delta\), we have

\[
- \log \mathbb{E}_{\widetilde{D}}[\exp(L(\widetilde{D})) \mid D] < - L(D) + \log(1/\delta)
\]

Then, by the definition of \( L \) and by the inequality \( -\log x \ge 1 - x \),

\begin{align*}
    - \log \mathbb{E}_{\widetilde{D}}[\exp(L(\widetilde{D})) \mid D] = &- \sum_{i=1}^{n} \log \mathbb{E}_{o \sim \mathbb{P}^{*}_{i}} \left[ \sqrt{\frac{\widehat{\mathbb{P}}_{i}(o) }{ \mathbb{P}^{*}_{i}(o) } }\right] \\
\ge& \sum_{i=1}^{n} \left( 1 - \mathbb{E}_{o \sim \mathbb{P}^{*}_{i}} \left[ \sqrt{\frac{\widehat{\mathbb{P}}_{i}(o) }{ \mathbb{P}^{*}_{i}(o) } }\right] \right).
\end{align*}

Furthermore, by algebra,

\begin{equation}\label{equ: length one lemma 15 algebra}
\begin{aligned}
&\sum_{i=1}^{n} \left( 1 - \mathbb{E}_{o \sim \mathbb{P}^{*}_{i}} \left[ \sqrt{\frac{\widehat{\mathbb{P}}_{i}(o) }{ \mathbb{P}^{*}_{i}(o) } }\right] \right)\\
=& \sum_{i=1}^{n} \left( 1 - \sum_{o} \sqrt{ \widehat{\mathbb{P}}_{i}(o)\mathbb{P}^{*}_{i}(o) } \right)\\
=& \frac{1}{2} \sum_{i=1}^{n} \left( \sum_{o} \left( \sqrt{\widehat{\mathbb{P}}_{i}(o)} - \sqrt{\mathbb{P}^{*}_{i}(o)} \right)^2 \right)\\
\ge &\frac{1}{8} \sum_{i=1}^{n} \left[ \sum_{o} \left( \sqrt{\widehat{\mathbb{P}}_{i}(o)} - \sqrt{\mathbb{P}^{*}_{i}(o)} \right)^2\right]\left[\sum_{o} \left( \sqrt{\widehat{\mathbb{P}}_{i}(o)} + \sqrt{\mathbb{P}^{*}_{i}(o)} \right)^2 \right]\\
\ge& \frac{1}{8} \sum_{i=1}^{n}\left( \sum_{o} \left| \widehat{\mathbb{P}}_{i}(o) -  \mathbb{P}^{*}_{i}(o)\right|\right)^{2}.
\end{aligned}
\end{equation}

where the last inequality follows from the Cauchy--Schwarz inequality. Putting all relations together,  
we have that with probability at least \(1 - \delta\),

\[
\frac{1}{8} \sum_{i=1}^{n}\left( \sum_{o} \left| \widehat{\mathbb{P}}_{i}(o) -  \mathbb{P}^{*}_{i}(o)\right|\right)^{2}\le - L(D)+ \log(1/\delta) .
\]

We plug in the definition of $\widehat{\mathbb{P}}_{i}(o)$ and $\mathbb{P}^{*}_{i}(o)$ to obtain the left-hand side of \eqref{equ: general one-step bound}, and expand $-L(D)$ and use the result from Proposition~\ref{prop:factorization} to obtain the right-hand side of \eqref{equ: general one-step bound}.

\subsection{}\label{apx: proof of adapted lemma 15}
It is important to mention that all expectations that appear in the lemma and in this proof exist because each $o_{i}$ and $\tilde{o}_{i}$ can only take finitely many different values whereas $\{\bar{b}_{i,\theta^*}\}_{i=1}^{n}$ and $\{\bar{b}_{i,\hat{\theta}}\}_{i=1}^{n}$ are fixed and do not affect the expectations.

Define \( E_i = \mathbb{E}[\exp\left(\ell(\tilde{o}_{i},\bar{b}_{i,\theta^*},\bar{b}_{i,\hat{\theta}})\right) \mid \bar{b}_{i,\theta^*},\bar{b}_{i,\hat{\theta}}] \) and

\[
u_n = \mathbb{E} \left[ \exp\left( L(\mathcal{D}) - \log \mathbb{E}[\exp(L(\widetilde{\mathcal{D}})) \mid \mathcal{D}] \right) \right] 
= \mathbb{E} \left[ \frac{\exp\left(\sum_{i=1}^n \ell(o_{i},\bar{b}_{i,\theta^*},\bar{b}_{i,\hat{\theta}})\right)}{\mathbb{E}[\exp\left(\sum_{i=1}^n \ell(\tilde{o}_{i},\bar{b}_{i,\theta^*},\bar{b}_{i,\hat{\theta}})\right) \mid \mathcal{D}]} \right].
\]

We want to show that \( u_n = 1 \). By the definition of \( \{ \tilde{o}_i \}_{i=1}^{n} \), some calculation gives

\[
\mathbb{E} \left[ \exp \left( \sum_{i=1}^{n}\ell(\tilde{o}_{i},\bar{b}_{i,\theta^*},\bar{b}_{i,\hat{\theta}}) \right) \middle| \mathcal{D} \right] = \prod_{i=1}^n E_i.
\]

Plugging this into the last expression obtained for \( u_n \) and using the tower rule with 

\[
\mathcal{D}_{n-1} = (\bar{b}_{1,\theta^*},\bar{b}_{1,\hat{\theta}}, o_{1}, \ldots, \bar{b}_{n-1,\theta^*},\bar{b}_{n-1,\hat{\theta}}, o_{n-1}, \bar{b}_{n,\theta^*},\bar{b}_{n,\hat{\theta}})
\]

we get

\[
u_n = \mathbb{E} \left[ \mathbb{E} \left[ \frac{\exp(\sum_{i=1}^n \ell(o_{i},\bar{b}_{i,\theta^*},\bar{b}_{i,\hat{\theta}})}{\prod_{i=1}^n E_i} \middle| D_{n-1} \right] \right] 
= \mathbb{E} \left[ \frac{\mathbb{E}[\exp(\sum_{i=1}^n \ell(o_{i},\bar{b}_{i,\theta^*},\bar{b}_{i,\hat{\theta}})) \mid D_{n-1}]}{\prod_{i=1}^n E_i} \right],
\]

where we used that \( E_1, \ldots, E_n \) are \( \sigma(D_{t-1}) \)-measurable. Now,

\[
\mathbb{E} \left[ \exp \left( \sum_{i=1}^n \ell(o_{i},\bar{b}_{i,\theta^*},\bar{b}_{i,\hat{\theta}}) \right) \middle| D_{n-1} \right] 
= \mathbb{E} \left[ \exp \left( \sum_{i=1}^{n-1} \ell(o_{i},\bar{b}_{i,\theta^*},\bar{b}_{i,\hat{\theta}}) \right) \middle| D_{n-1} \right] E_n.
\]

Plugging this back into our previous expression,

\begin{align*}
u_n &= \mathbb{E} \left[ \frac{ \mathbb{E} \left[ \exp\left( \sum_{i=1}^{n-1} \ell(o_{i},\bar{b}_{i,\theta^*},\bar{b}_{i,\hat{\theta}}) \right) \mid D_{n-1} \right] E_n }{ \left( \prod_{i=1}^{n-1} E_i \right) E_n } \right] \\
&= \mathbb{E} \left[ \mathbb{E} \left[ \frac{ \exp\left( \sum_{i=1}^{n-1} \ell(o_{i},\bar{b}_{i,\theta^*},\bar{b}_{i,\hat{\theta}}) \right) }{ \prod_{i=1}^{n-1} E_i } \bigg| D_{n-1} \right] \right] \\
&= \mathbb{E} \left[ \frac{ \exp\left( \sum_{i=1}^{n-1} \ell(o_{i},\bar{b}_{i,\theta^*},\bar{b}_{i,\hat{\theta}}) \right) }{ \prod_{i=1}^{n-1} E_i } \right] \\
&= u_{n-1} = \cdots = u_0 = 1,
\end{align*}

where the second equality used that \( E_1, \ldots, E_{t-1} \) is \( \sigma(D_{t-1}) \)-measurable.

\section{}\label{apx: proof for universal quadratic bound for O}
In this appendix, we extend the pointwise guarantee of Lemma~\ref{lem:quadratic bound for O} to a uniform statement over the parameter space. We begin by formally stating the resulting global guarantee.

\begin{corollary}\label{cor:universal quadratic bound for O}
Let
\[
\Theta
=
\Delta(\mathcal S)
\times
\prod_{a\in\mathcal A,\,s\in\mathcal S}\Delta(\mathcal S)
\times
\prod_{s\in\mathcal S}\Delta(\Omega)
\]
be the POMDP parameter space, and define
\[
d
=
(|\mathcal A||\mathcal S|+1)(|\mathcal S|-1)
+
|\mathcal S|(|\Omega|-1),
\qquad
K=\max\{|\mathcal S|,|\Omega|\}=|\Omega|.
\]
For $\theta=(\beta,\varphi)\in\Theta$, let
\[
p_{i,\theta}(o)
=
\sum_{s\in\mathcal S}
\bar b_{i,\theta}(s)O_\varphi(o\mid s)
\]
denote the predictive observation probability at time $i$.

There exists a universal constant $c>0$ such that, for any
$\varepsilon\in(0,1]$ and $\delta\in(0,1)$, with probability
at least $1-\delta$, the following holds simultaneously for
all $\hat\theta\in\Theta$: there exists
$\tilde\theta\in\Theta$ satisfying
$\|\tilde\theta-\hat\theta\|_\infty\leq\varepsilon$
and
\begin{align}
&\sum_{i=1}^n
\left(
\sum_{o\in\Omega}
\left|
p_{i,\tilde\theta}(o)
-
p_{i,\theta^*}(o)
\right|
\right)^2
\nonumber\\
&\quad\leq
c\left(
\sum_{i=1}^n
\log\frac{p_{i,\theta^*}(o_i)}
{p_{i,\hat\theta}(o_i)}
+
d\log\left(
2K\max\left\{n,\frac{1}{\varepsilon}\right\}
\right)
+
\log(1/\delta)
+2
\right).
\label{eq:uniform_prediction}
\end{align}
Here, a zero denominator in the likelihood ratio is
interpreted as giving an infinite right-hand side.
\end{corollary}

\begin{remark}
Corollary~\ref{cor:universal quadratic bound for O} shows that a small likelihood gap for any candidate parameter $\hat{\theta}$ implies a small prediction-error bound for some nearby discretized parameter $\tilde{\theta}$. Since $\|\hat{\theta}-\tilde{\theta}\|_{\infty}\leq \varepsilon$, any later bound on $\|\tilde{\theta}-\theta^*\|$ then be transferred back to $\hat{\theta}$ by paying only an additional $\varepsilon$-level discretization error. Furthermore, the discretization cost appears only through a logarithmic term.
\end{remark}

\begin{proof}
We discretize the parameter space using a deterministic grid.
Let
\[
h=\frac{\min\{\varepsilon,1/n\}}{K}.
\]
For each simplex in $\Theta$, we partition its first $k-1$
coordinates into intervals of length at most $h$, where $k$
is the dimension of that simplex. The last coordinate is
determined by the simplex constraint. Taking the Cartesian
product of these partitions gives a collection of nonempty
grid cells $C_1,\ldots,C_{N_h}$ covering $\Theta$.

Since $\Theta$ has $d$ free coordinates, the number of cells
satisfies
\[
N_h
\leq
\left\lceil\frac{1}{h}\right\rceil^d
\leq
\left(\frac{2}{h}\right)^d,
\]
where the last inequality follows from $h\leq1$.
Consequently,
\begin{equation}
\log N_h
\leq
d\log\left(
2K\max\left\{n,\frac{1}{\varepsilon}\right\}
\right).
\label{eq:grid_cardinality}
\end{equation}

We next construct a deterministic representative $\theta_j$
for each cell $C_j$. In particular, $\theta_j$ is deterministic, and will depend only on the cell and
not on the observed trajectory.

Consider one simplex row of a cell, denoted by
$C\subseteq\Delta(k)$. For each coordinate $r$, define
\[
u_r=\sup_{p\in C}p_r,
\qquad
Z=\sum_{r=1}^k u_r,
\]
and let
\[
q_r=\frac{u_r}{Z},
\qquad r=1,\ldots,k.
\]
Clearly, $q\in\Delta(k)$.

We first bound the normalization factor $Z$. Since
$u_r\geq p_r$ for every $p\in C$, we have $Z\geq1$.
Let $p^\circ$ be a point in the closure of $C$ maximizing
the last coordinate $p_k$. Then $u_k=p_k^\circ$.
Since the first $k-1$ coordinates vary by at most $h$
within the cell,
\[
u_r\leq p_r^\circ+h,
\qquad r=1,\ldots,k-1.
\]
Therefore,
\[
Z
=
\sum_{r=1}^{k-1}u_r+u_k
\leq
\sum_{r=1}^{k-1}(p_r^\circ+h)+p_k^\circ
=
1+(k-1)h.
\]
Thus,
\begin{equation}
1\leq Z\leq1+(k-1)h\leq1+Kh.
\label{eq:normalization_bound}
\end{equation}

For every $p\in C$ and every coordinate $r$, we have
$u_r\geq p_r$. It follows that
\begin{equation}
q_r
=
\frac{u_r}{Z}
\geq
\frac{p_r}{1+Kh}.
\label{eq:coordinate_domination}
\end{equation}

Moreover, the variation of each of the first $k-1$
coordinates within the cell is at most $h$, while the
variation of the last coordinate is at most $(k-1)h$.
Hence,
\[
0\leq u_r-p_r\leq(k-1)h
\]
for every $r$. Using \eqref{eq:normalization_bound},
we obtain
\[
\begin{aligned}
|q_r-p_r|
&=
\frac{|(u_r-p_r)-p_r(Z-1)|}{Z}\\
&\leq
(k-1)h
\leq Kh,
\end{aligned}
\]
where we used the fact that both $u_r-p_r$ and
$p_r(Z-1)$ belong to $[0,(k-1)h]$.

We apply this construction independently to every
simplex row of each grid cell $C_j$. The resulting
parameter $\theta_j\in\Theta$ is deterministic and
satisfies, for every $\theta\in C_j$,
\begin{equation}
\|\theta_j-\theta\|_\infty
\leq Kh\leq\varepsilon,
\label{eq:representative_distance}
\end{equation}
and
\begin{equation}
(\theta_j)_r
\geq
\frac{\theta_r}{1+Kh}
\label{eq:representative_domination}
\end{equation}
for every probability coordinate $r$ of the initial
distribution, transition kernel, and observation kernel.
The representative $\theta_{j}$ need not belong to $C_j$ itself.

We now compare the likelihoods under $\theta_j$ and
an arbitrary $\theta\in C_j$. Recall that
\[
\ell_{n}(\theta)
=
\sum_{s_{1:n}}
\mu_\beta(s_1)
\prod_{i=1}^n O_\varphi(o_i\mid s_i)
\prod_{i=1}^{n-1}
T_\beta^{a_i}(s_{i+1}\mid s_i).
\]

From Proposition~\ref{prop:factorization},
\[
\ell_{n}(\theta)
=
\prod_{i=1}^n p_{i,\theta}(o_i).
\]

Each product inside the sum defining $\ell_{n}(\theta)$
contains one initial-state probability, $n$ observation
probabilities, and $n-1$ transition probabilities,
for a total of $2n$ model-dependent factors.
By \eqref{eq:representative_domination}, every such
factor under $\theta_j$ is at least the corresponding
factor under $\theta$ divided by $1+Kh$. Thus,
for every hidden-state trajectory,
\[
\begin{aligned}
&\mu_{\beta_j}(s_1)
\prod_{i=1}^n O_{\varphi_j}(o_i\mid s_i)
\prod_{i=1}^{n-1}
T_{\beta_j}^{a_i}(s_{i+1}\mid s_i)
\\
&\qquad\geq
(1+Kh)^{-2n}
\mu_\beta(s_1)
\prod_{i=1}^n O_\varphi(o_i\mid s_i)
\prod_{i=1}^{n-1}
T_\beta^{a_i}(s_{i+1}\mid s_i).
\end{aligned}
\]

Therefore,
\[
\ell_{n}(\theta_{j})
\geq
(1+Kh)^{-2n}\ell_{n}(\theta).
\]
Taking logarithms, we obtain
\begin{align}
\log\frac{\ell_{n}(\theta^*)}{\ell_{n}(\theta_{j})}
&\leq
\log\frac{\ell_{n}(\theta^*)}{\ell_{n}(\theta)}
+
2n\log(1+Kh)
\nonumber\\
&\leq
\log\frac{\ell_{n}(\theta^*)}{\ell_{n}(\theta)}
+
2nKh
\nonumber\\
&\leq
\log\frac{\ell_{n}(\theta^*)}{\ell_{n}(\theta)}
+2,
\label{eq:likelihood_comparison}
\end{align}
where the last inequality follows from $Kh\leq1/n$.
This comparison holds deterministically for every
observed trajectory and every $\theta\in C_j$.

Finally, we apply Lemma~\ref{lem:quadratic bound for O} to the deterministic
representatives $\theta_1,\ldots,\theta_{N_h}$,
with failure probability $\delta/N_h$ for each one.
A union bound implies that, with probability at least
$1-\delta$, simultaneously for every
$j\in\{1,\ldots,N_h\}$,
\begin{align}
&\sum_{i=1}^n
\left(
\sum_{o\in\Omega}
|p_{i,\theta_j}(o)-p_{i,\theta^*}(o)|
\right)^2
\nonumber\\
&\quad\leq
c\left(
\log\frac{L_{\theta^*}(D)}{L_{\theta_j}(D)}
+
\log(N_h/\delta)
\right).
\label{eq:union_bound_representatives}
\end{align}
This union bound is valid because the representatives
are fixed before observing the data.

On this event, consider any
$\hat\theta\in\Theta$, possibly selected
using the observed trajectory. Let $C_j$ be a grid
cell containing $\hat\theta$, and choose
$\tilde\theta=\theta_j$.
By \eqref{eq:representative_distance},
\[
\|\tilde\theta-\hat\theta\|_\infty
\leq\varepsilon.
\]
Furthermore, combining
\eqref{eq:union_bound_representatives},
\eqref{eq:likelihood_comparison}, and
\eqref{eq:grid_cardinality}, we have
\[
\begin{aligned}
&\sum_{i=1}^n
\left(
\sum_{o\in\Omega}
|p_{i,\tilde\theta}(o)-p_{i,\theta^*}(o)|
\right)^2
\\
&\quad\leq
c\left(
\log\frac{L_{\theta^*}(D)}
{L_{\hat\theta}(D)}
+
\log(N_h/\delta)
+2
\right)
\\
&\quad\leq
c\left(
\sum_{i=1}^n
\log\frac{p_{i,\theta^*}(o_i)}
{p_{i,\hat\theta}(o_i)}
+
d\log\left(
2K\max\left\{n,\frac{1}{\varepsilon}\right\}
\right)
+
\log(1/\delta)
+2
\right).
\end{aligned}
\]
Since the high-probability event holds simultaneously
for all deterministic representatives, the conclusion
holds for every $\hat\theta\in\Theta$ on the same
event. This proves the claim.
\end{proof}

\section{}\label{apx: proof for 2-step mismatch thm}
First, we acknowledge that Lemma~\ref{lem:length-2 quadratic bound for O} is inspired by the high-level argument underlying Propositions~13 and~14 of \citet{liu2022partially}, but the statement and proof are substantially modified to tailor to our predictive-distribution formulation.

For notational convenience, define
\[\mathbb{P}_k^*(o,o'\mid a)
:=
\sum_{s\in\mathcal S}
\bar b_{k,\theta^*}(s)
\mathbb{P}_{\theta^*}(o,o'\mid s,a)
\]
and
\[
\widehat{\mathbb{P}}_k(o,o'\mid a)
:=
\sum_{s\in\mathcal S}
\bar b_{k,\hat\theta}(s)
\mathbb{P}_{\hat\theta}(o,o'\mid s,a).
\]

Let $H_{k-1}$ denote the observed
history preceding $o_k$, including the
previously selected actions. Since the
behavior policy chooses $a_k$ after
observing $o_k$, the conditional joint
distributions of $(o_k,a_k,o_{k+1})$ are
\begin{align*}
R_k^*(o,a,o')
&:=\pi(a\mid o)\mathbb{P}_k^*(o,o'\mid a),\\
\widehat R_k(o,a,o')
&:=\pi(a\mid o)\widehat{\mathbb{P}}_k(o,o'\mid a).
\end{align*}
In particular, $R_k^*$ is the true
conditional distribution of
$(o_k,a_k,o_{k+1})$ given
$H_{k-1}$.

Define the Hellinger affinity
\[
C_k:=
\sum_{o,a,o'}
\sqrt{
R_k^*(o,a,o')
\widehat R_k(o,a,o')
}.
\]
Both $R_k^*$ and $\widehat R_k$ are
probability distributions, so
$0\le C_k\le1$.

For $k<n$, let
\[
\lambda_k
:=
\frac12\log
\frac{\widehat{ \mathbb{P}}_k(o_k,o_{k+1}\mid a_k)}
{\mathbb{P}_k^*(o_k,o_{k+1}\mid a_k)}.
\]
The behavior-policy factors cancel from
this likelihood ratio. Moreover,
\begin{align}
&\mathbb E_{\theta^*}
\left[
\exp{(\lambda_k)}\mid H_{k-1}
\right]
\nonumber\\
&=
\sum_{o,a,o'}R_k^*(o,a,o')
\sqrt{
\frac{\widehat{\mathbb{P}}_k(o,o'\mid a)}
{\mathbb{P}_k^*(o,o'\mid a)}
}
\nonumber\\
&=
\sum_{o,a,o'}
\pi(a\mid o)
\sqrt{
\mathbb{P}_k^*(o,o'\mid a)
\widehat{\mathbb{P}}_k(o,o'\mid a)
}
\nonumber\\
&=C_k.
\label{eq:corrected-pair-normalization}
\end{align}

We also define the one-step predictive
observation distributions
\begin{align*}
g_k^*(o)
&:=
\sum_s\bar b_{k,\theta^*}(s)
O_{\varphi^*}(o\mid s),\\
\widehat g_k(o)
&:=
\sum_s\bar b_{k,\hat\theta}(s)
O_{\hat\varphi}(o\mid s).
\end{align*}
For the boundary observations
$j\in\{1,n\}$, set
\[
\xi_j:=
\frac12\log
\frac{\widehat g_j(o_j)}{g_j^*(o_j)}
\]
and
\[
B_j:=
\sum_o\sqrt{g_j^*(o)\widehat g_j(o)}.
\]
Again, $0\le B_j\le1$, and
\begin{equation}
\mathbb E_{\theta^*}
\left[
\exp{(\xi_j)}\mid H_{j-1}
\right]=B_j.
\label{eq:corrected-boundary-normalization}
\end{equation}

We now divide the likelihood factors into
two nonoverlapping collections, as in
the original proof. Let
\begin{align*}
I_1&:=\{k<n:k\text{ is odd}\},\\
I_2&:=\{k<n:k\text{ is even}\}.
\end{align*}
To account for the boundary observations,
define
\[
J_1:=
\begin{cases}
\varnothing,&n\text{ is even},\\
\{n\},&n\text{ is odd},
\end{cases}
\]
and
\[
J_2:=
\begin{cases}
\{1,n\},&n\text{ is even},\\
\{1\},&n\text{ is odd}.
\end{cases}
\]
For $r\in\{1,2\}$, write
\[
L_r:=
\sum_{k\in I_r}\lambda_k
+
\sum_{j\in J_r}\xi_j
\]
and define
\[
M_r:=
\frac{\exp(L_r)}
{
\displaystyle
\prod_{k\in I_r}C_k
\prod_{j\in J_r}B_j
}.
\]

Within each collection, the observation
pairs do not overlap, and any remaining
observations appear as singleton boundary
terms. By
\eqref{eq:corrected-pair-normalization},
\eqref{eq:corrected-boundary-normalization},
and iterated conditional expectations,
\[
\mathbb E_{\theta^*}[M_r]=1,
\qquad r=1,2.
\]

If the observed trajectory has zero likelihood under
$\hat\theta$, the desired bound holds trivially.
Thus, we assume its likelihood is positive.

Applying Markov's inequality and a union
bound, with probability at least
$1-\delta$ we have, simultaneously for
$r=1,2$,
\[
\log M_r\le\log\frac{2}{\delta}.
\]
Since $-\log x\ge1-x$ for $x>0$, and
$B_j\le1$, it follows that
\begin{align}
\sum_{k\in I_r}(1-C_k)
&\le
-\sum_{k\in I_r}\log C_k
\nonumber\\
&\le
-L_r+\log\frac{2}{\delta}.
\label{eq:corrected-concentration}
\end{align}

Next, by the definition of $C_k$,
\begin{align}
1-C_k
&=
\frac12
\sum_{o,a,o'}
\pi(a\mid o)
\left(
\sqrt{\widehat{\mathbb{P}}_k(o,o'\mid a)}
-
\sqrt{\mathbb{P}_k^*(o,o'\mid a)}
\right)^2
\nonumber\\
&\ge
\frac{\pi_{\min}}2
\sum_{o,o'}
\left(
\sqrt{\widehat{\mathbb{P}}_k(o,o'\mid a_k)}
-
\sqrt{\mathbb{P}_k^*(o,o'\mid a_k)}
\right)^2.
\label{eq:corrected-hellinger}
\end{align}
The last inequality follows from
$\pi(a\mid o)\ge\pi_{\min}$ and by
retaining only the terms corresponding
to the realized action $a_k$.

Furthermore, by Cauchy--Schwarz,
\begin{align*}
&\left(
\sum_{o,o'}
\left|
\widehat{\mathbb{P}}_k(o,o'\mid a_k)
-\mathbb{P}_k^*(o,o'\mid a_k)
\right|
\right)^2\\
&\le
\left[
\sum_{o,o'}
\left(
\sqrt{\widehat{\mathbb{P}}_k(o,o'\mid a_k)}
-
\sqrt{\mathbb{P}_k^*(o,o'\mid a_k)}
\right)^2
\right]\\
&\quad\cdot
\left[
\sum_{o,o'}
\left(
\sqrt{\widehat{\mathbb{P}}_k(o,o'\mid a_k)}
+
\sqrt{\mathbb{P}_k^*(o,o'\mid a_k)}
\right)^2
\right]\\
&\le
4\sum_{o,o'}
\left(
\sqrt{\widehat{\mathbb{P}}_k(o,o'\mid a_k)}
-
\sqrt{\mathbb{P}_k^*(o,o'\mid a_k)}
\right)^2,
\end{align*}
where the final inequality uses the fact
that both $\widehat{\mathbb{P}}_k(\cdot,\cdot\mid a_k)$
and $\mathbb{P}_k^*(\cdot,\cdot\mid a_k)$ are
probability distributions.

Combining this with
\eqref{eq:corrected-hellinger} gives
\begin{equation}
1-C_k
\ge
\frac{\pi_{\min}}8
\left(
\sum_{o,o'}
\left|
\widehat{\mathbb{P}}_k(o,o'\mid a_k)
-\mathbb{P}_k^*(o,o'\mid a_k)
\right|
\right)^2.
\label{eq:corrected-L1-hellinger}
\end{equation}

Substituting
\eqref{eq:corrected-L1-hellinger}
into \eqref{eq:corrected-concentration}
and summing over $r=1,2$, we obtain
\begin{align}
&\frac{\pi_{\min}}8
\sum_{k=1}^{n-1}
\left(
\sum_{o,o'}
\left|
\widehat{\mathbb{P}}_k(o,o'\mid a_k)
-\mathbb{P}_k^*(o,o'\mid a_k)
\right|
\right)^2
\nonumber\\
&\qquad\le
-L_1-L_2+2\log\frac{2}{\delta}.
\label{eq:corrected-combined}
\end{align}

It remains to relate $L_1+L_2$ to the
model-dependent likelihood. By the
sequential predictive factorization,
for every realized observation pair,
\begin{align*}
\mathbb{P}_k^*(o_k,o_{k+1}\mid a_k)
&=g_k^*(o_k)g_{k+1}^*(o_{k+1}),\\
\widehat{\mathbb{P}}_k(o_k,o_{k+1}\mid a_k)
&=\widehat g_k(o_k)
\widehat g_{k+1}(o_{k+1}).
\end{align*}
Here the predictive distributions at
time $k+1$ incorporate the observed
$o_k$ and the selected action $a_k$.

Consequently,
\[
\lambda_k
=
\frac12\log
\frac{\widehat g_k(o_k)}{g_k^*(o_k)}
+
\frac12\log
\frac{\widehat g_{k+1}(o_{k+1})}
{g_{k+1}^*(o_{k+1})}.
\]
Since every pair index $k<n$ appears
exactly once in $I_1\cup I_2$, and the
two boundary terms appear exactly once
in $J_1\cup J_2$, we have
\begin{align*}
L_1+L_2
&=
\sum_{k=1}^{n-1}\lambda_k
+\xi_1+\xi_n\\
&=
\sum_{k=1}^{n}
\log\frac{\widehat g_k(o_k)}
{g_k^*(o_k)}\\
&=
\ell_n(\hat\theta)-\ell_n(\theta^*).
\end{align*}
The final equality follows because
the behavior-policy factors are
parameter-independent and therefore
cancel from the likelihood ratio.

Substituting this identity into
\eqref{eq:corrected-combined} yields
\begin{align}
&\frac{\pi_{\min}}8
\sum_{k=1}^{n-1}
\left(
\sum_{o,o'}
\left|
\widehat{\mathbb{P}}_k(o,o'\mid a_k)
-\mathbb{P}_k^*(o,o'\mid a_k)
\right|
\right)^2
\nonumber\\
&\qquad\le
\ell_n(\theta^*)-\ell_n(\hat\theta)
+2\log\frac{2}{\delta}.
\label{eq:corrected-final-intermediate}
\end{align}

Finally, since the total $\ell_1$
distance between two probability
distributions is at most $2$,
\[
\left(
\sum_{o,o'}
\left|
\widehat{\mathbb{P}}_n(o,o'\mid a_n)
-\mathbb{P}_n^*(o,o'\mid a_n)
\right|
\right)^2
\le4.
\]
Combining this with
\eqref{eq:corrected-final-intermediate}
establishes
\[
\sum_{k=1}^{n}
\left(
\sum_{o,o'}
\left|
\widehat{\mathbb{P}}_k(o,o'\mid a_k)
-\mathbb{P}_k^*(o,o'\mid a_k)
\right|
\right)^2
\le
\frac8{\pi_{\min}}
\left(
\ell_n(\theta^*)-\ell_n(\hat\theta)
+2\log\frac2\delta
\right)+4.
\]
This completes the proof.

\section{}\label{apx: global generalization of length-2 quadratic bound}
In this Appendix, we state the global counterpart of Lemma~\ref{lem:length-2 quadratic bound for O}. Moreover, we simply state the result without proving it since the same argument from Appendix~\ref{apx: proof for universal quadratic bound for O} applies.
\begin{corollary}\label{cor:universal length-2 quadratic bound for O}
Let
\[
\Theta
=
\Delta(\mathcal S)
\times
\prod_{a\in\mathcal A,\,s\in\mathcal S}\Delta(\mathcal S)
\times
\prod_{s\in\mathcal S}\Delta(\Omega)
\]
be the POMDP parameter space, and define
\[
d=(|\mathcal A||\mathcal S|+1)(|\mathcal S|-1)
+|\mathcal S|(|\Omega|-1),
\qquad
K=\max\{|\mathcal S|,|\Omega|\}=|\Omega|.
\]
For $\theta=(\beta,\varphi)\in\Theta$, let
\[
\mathbb P_\theta(o,o'\mid s,a)
=
O_\varphi(o\mid s)
\sum_{s'\in\mathcal S}
T_\beta^a(s'\mid s)O_\varphi(o'\mid s').
\]

For any $\varepsilon\in(0,1]$ and $\delta\in(0,1)$,
with probability at least $1-\delta$, the following
holds simultaneously for all $\hat\theta\in\Theta$:
there exists $\tilde\theta\in\Theta$ satisfying
$\|\tilde\theta-\hat\theta\|_\infty\leq\varepsilon$
and
\begin{align}
&\sum_{k=1}^n
\Bigg(
\sum_{o,o'\in\Omega}
\Bigg|
\sum_{s\in\mathcal S}
\bar b_{k,\tilde\theta}(s)
\mathbb P_{\tilde\theta}(o,o'\mid s,a_k)
-
\sum_{s\in\mathcal S}
\bar b_{k,\theta^*}(s)
\mathbb P_{\theta^*}(o,o'\mid s,a_k)
\Bigg|
\Bigg)^2
\nonumber\\
&\quad\leq
\frac{8}{\pi_{\min}}
\left(
\ell_n(\theta^*)-\ell_n(\hat\theta)
+
2d\log\left(
2K\max\left\{n,\frac{1}{\varepsilon}\right\}
\right)
+
2\log(2/\delta)
+2
\right)+4.
\label{eq:uniform_two_step}
\end{align}
Here, the bound is understood to be vacuous if
$\ell_n(\widehat\theta)=-\infty$.
\end{corollary}
\begin{remark}
Corollary~\ref{cor:universal length-2 quadratic bound for O} shows that a small likelihood gap for any candidate parameter $\hat{\theta}$ implies a small length-2 prediction-error bound for some nearby discretized parameter $\tilde{\theta}$. Since $\|\hat{\theta}-\tilde{\theta}\|_{\infty}\leq \varepsilon$, any later bound on $\|\tilde{\theta}-\theta^*\|$ then be transferred back to $\hat{\theta}$ by paying only an additional $\varepsilon$-level discretization error. Furthermore, the discretization cost appears only through a logarithmic term.
\end{remark}

\section{}\label{apx: discussion of the assumptions}
Table~\ref{tab:assumptions} and \ref{tab:assumptions_continued} summarize the main assumptions used in our
analysis, their roles, possible relaxations, and practical implications. The key takeaway is that several
assumptions are convenient sufficient conditions rather than minimal
requirements: some form of observability and coverage is fundamental, while
the particular one-step, conditioning, policy, and mixing conditions can often
be relaxed.

\begin{table*}[!t]
\centering
\caption{
Summary of the main assumptions, their roles in the analysis, possible
relaxations, and practical implications.
}
\label{tab:assumptions}
\small
\setlength{\tabcolsep}{6pt}
\renewcommand{\arraystretch}{1.15}

\begin{tabularx}{\textwidth}{
    @{}
    >{\raggedright\arraybackslash}p{0.09\textwidth}
    Y
    Y
    Y
    @{}
}
\toprule
\textbf{Assumption}
&
\textbf{Why needed}
&
\textbf{How relaxable}
&
\textbf{Practical implications}
\\
\midrule

\ref{ass: behavior policy undercomplete}
&
Makes the induced latent process a stationary Markov chain, enabling mixing
and concentration analysis.
&
Imposed for convenience. Finite-memory policies can be handled by augmenting
the latent state with policy memory.
&
Arbitrary full-history adaptation is not covered, but state augmentation
handles the broad class of finite-memory adaptive exploration.
\\[2pt]

\ref{ass: minimum policy mass}
&
Ensures that each action is selected with uniformly positive probability under every observation.
&
Some form of observation–action excitation is necessary for uniform recovery guarantees. Without sufficient coverage, belief matrices can become rank deficient, rendering distinct POMDPs observationally indistinguishable.
&
When certain actions are rarely selected under some observations, the collected data may be insufficient for reliable kernel recovery. Randomizing action selection or combining data from different behavior policies can improve coverage.
\\[2pt]

\ref{ass:alpha_undercomplete}
&
Makes latent states distinguishable from one-step observations and controls
the rate through $\sigma_O$.
&
Some observability is necessary for latent-state identification. If one-step
observability fails but Assumption~\ref{ass:multi_step_revelation} holds, the overcomplete analysis
applies.
&
Small $\sigma_O$ makes recovery ill-conditioned and raises sample complexity;
longer observation windows may restore separation under Assumption~\ref{ass:multi_step_revelation}.
\\[2pt]

\ref{ass: full-rank transition kernels}
&
Ensures stable inversion of the transition matrices in recovery steps that require their inverses.
&
It is mentioned in Remark~\ref{remark: extending the full-rank assumption of T} that \ref{ass: full-rank transition kernels} can be replaced by the aggregate conditioning condition in 
Appendix~\ref{apx: replacing the full-rank transition kernels assumption}; the stated assumption is only a convenient sufficient condition.
&
Near-singularity amplifies predictive-distribution error and increases sample
complexity.
\\[2pt]

\ref{ass: ergodicity}
&
Ergodicity ensures positive visitation of all states; mixing controls
dependence and enables concentration for a single trajectory.
&
Without ergodicity, unvisited states are unlearnable. Multiple independent
trajectories can replace within-trajectory mixing if their reset distribution
provides adequate coverage.
&
Longer runs help with slow mixing but not missing recurrent classes; covering
them requires resets or broadly initialized trajectories.
\\
\bottomrule
\end{tabularx}
\end{table*}
\begin{table*}[!t]
\centering
\caption{
Summary of the main assumptions, their roles in the analysis, possible
relaxations, and practical implications.
}
\label{tab:assumptions_continued}
\small
\setlength{\tabcolsep}{6pt}
\renewcommand{\arraystretch}{1.15}

\begin{tabularx}{\textwidth}{
    @{}
    >{\raggedright\arraybackslash}p{0.09\textwidth}
    Y
    Y
    Y
    @{}
}
\toprule
\textbf{Assumption}
&
\textbf{Why needed}
&
\textbf{How relaxable}
&
\textbf{Practical implications}
\\
\midrule

\ref{ass:multi_step_revelation}
&
Uses length-$m$ observation blocks and informative action strings when
one-step observations do not separate latent mixtures; $m=1$ recovers
Assumption~\ref{ass:alpha_undercomplete}.
&
Some multi-step separation is necessary when the observation kernel is
rank-deficient or when the POMDP is overcomplete. Assumption~\ref{ass:multi_step_revelation} is
sufficient, but is not claimed to be minimal.
&
Choose informative action sequences that produce distinguishable observation patterns across latent states. Use the shortest observation horizon \(m\) that provides sufficient separation, since longer sequences increase sample requirements.
\\[2pt]

\ref{ass: L-step policy coverage}, \ref{ass: L-step ergodicity}
&
Assumption~\ref{ass: L-step policy coverage} ensures positive sampling of informative strings and a Markov
block process; Assumption~\ref{ass: L-step ergodicity} supplies ergodicity for mixing and
concentration.
&
Kernel identification requires coverage of informative action strings and
latent states, but Assumptions~\ref{ass: L-step policy coverage} and \ref{ass: L-step ergodicity} are sufficient rather than claimed
minimal conditions.
&
Rare informative actions or slow mixing reduce the effective sample size.
Uniform randomization lower-bounds informative-action sampling probabilities,
while independent trajectories reduce dependence.
\\[2pt]

\ref{ass:distinct-observation-columns}
&
Without this, observationally indistinguishable latent states can admit distinct, observationally equivalent POMDP representations that are not related by state permutations, making kernel identification up to permutation impossible in general.
&
It is necessary as established in Appendix~\ref{apx: justification for distinct observation columns}.
&
If two latent states produce identical patterns of observable signals, distinguishing them requires additional information, and their corresponding transition dynamics may not be uniquely identifiable from data.
\\[2pt]
\ref{ass:backward-excitation}
&
Prevents exponentially ill-conditioned backward observability, where latent-state directions are nearly indistinguishable from past observation histories, leading to exponential sample complexity.
&
Necessary in general, as dropping it permits exponentially hard instances even under the remaining assumptions. More strongly, it is strictly necessary for our proof, since polynomial conditioning of the predictive belief matrix requires polynomial backward excitation.
&
Design data collection to generate sufficiently diverse beliefs about the current latent state. Otherwise, repeatedly observing nearly identical mixtures of latent states makes it difficult to disentangle their individual dynamics, even when future observations are informative.
\\
\bottomrule
\end{tabularx}
\end{table*}

\section{}\label{apx: proof for predictive belief full-rankedness}
In this appendix, we prove the following proposition, which constitutes the main technical component of the proof of Lemma~\ref{lem: sigma B_a proof} in Appendix~\ref{apx: proof for lem sigma B_a}.

\begin{proposition}\label{prop: predictive belief full-rankedness_formal}
Under Assumption~\ref{ass:alpha_undercomplete}, \ref{ass: full-rank transition kernels}, \ref{ass: minimum policy mass}, and \ref{ass: ergodicity}, let
\[t_{\mathrm{burn}}=1+
\left\lceil
\frac{\log\!\bigl(8G(\pi)/d_{\min}^{\pi}\bigr)}
{\log\!\bigl(1/\rho(\pi)\bigr)}
\right\rceil,\]
\begin{align*}
L=&\min\Bigg(\frac{C|\mathcal{S}|^2}{\sigma_O^4} \, \log\!\left( \frac{1024\,(1 + 1/\delta_2)\,|\mathcal{S}|}{49|\mathcal{A}|\,\pi_{\min}\,\sigma_T^2\,\sigma_O^2\,(d_{\min}^\pi)^2} \right),\frac{C|\mathcal{S}|}{\sigma_O^2} \, \log^2\!\left( \frac{1024\,(1 + 1/\delta_2)\sqrt{|\mathcal{S}|\log|\mathcal{S}|}}{49|\mathcal{A}|\,\pi_{\min}\,\sigma_T^2\,\sigma_O^2\,(d_{\min}^\pi)^2} \right)\Bigg),
\end{align*}
and
\[D=\frac{14\, \bigl(\log(8/\delta_1) + |\mathcal{S}|\log 9\bigr)}{|\mathcal{A}|^2\, \pi_{\min}^2\, \sigma_T^4\, \sigma_O^4\, (d_{\min}^\pi)^4}.\]

If
\begin{equation}\label{equ:n lower bonud for bar B sigma_min}
n\geq 2DL+\frac{2D}{\log(\frac{1}{\rho(\pi)})}\left[\log\left(\frac{4G(\pi)}{\delta_{1}}\right)+\max\left\{\log\left(\frac{2D}{\log(\frac{1}{\rho(\pi)})}\right),0\right\}\right]+t_{\mathrm{burn}},
\end{equation}
then, with probability at least $1-\delta_1-\delta_2$,
\begin{align*}
&\sigma_{\min}\!\left(\bar{B}\right)^2 
\geq \frac{1}{12}(n-t_{\mathrm{burn}})|\mathcal A|\,\pi_{\min}\,\sigma_{T}^2\,\sigma_{O}^2
\left(d_{\min}^{\pi}\right)^2.
\end{align*}
\end{proposition}

Due to the length of the proof, we further divide the proof into subsections.

\subsection{}\label{subapx: sigma B in expectation}
We start by proving several results regarding the expected behavior of the true belief vectors ($\{b_{t,\theta^*}\}$) and predictive belief vectors ($\{\bar{b}_{t,\theta^*}\}$). Moreover, for notational simplicity, throughout Appendix~\ref{apx: proof for predictive belief full-rankedness} we omit the $\theta^*$ and denote the true belief vectors as $b_{t}$ and true predictive belief vectors as $\bar{b}_{t}$.

In addition, we use $\|\cdot\|_{op}$ to denote the matrix spectral norm in the rest of Appendix~\ref{apx: proof for predictive belief full-rankedness}.

Recall the definition of history $H_{t}=(o^{t}_{1},a^{t}_{1})$. Then, we have the following lemma.
\begin{lemma}\label{lem: filtered belief isotropy}
Under Assumptions~\ref{ass:alpha_undercomplete} and \ref{ass: ergodicity}, the following statements hold.

\begin{enumerate}
\item For every $t \ge 1$,
\[
\mathbb{E}[b_tb_t^{\top} \mid H_{t-1}]
\;\succeq\;
\sigma_{O}^2 \operatorname{Diag}(\bar{b}_t)^2
\qquad \text{almost surely.}
\]

\item If
\[
d^{\pi}_t(s) = \mathbb{P}^{\pi}(X_t=s),
\qquad
d^{\pi}_{t,\min} = \min_{s \in \mathcal{S}} d^{\pi}_t(s),
\]
then for every $t \ge 1$,
\[
\mathbb{E}[b_tb_t^{\top}]
\;\succeq\;
\sigma_{O}^2 \operatorname{Diag}(d^{\pi}_t)^2
\;\succeq\;
\sigma_{O}^2 (d^{\pi}_{t,\min})^{2} I.
\]

\item Define
\[
t_{\mathrm{burn}}
=
1+
\left\lceil
\frac{\log\!\bigl(8G(\pi)/d_{\min}^{\pi}\bigr)}
{\log\!\bigl(1/\rho(\pi)\bigr)}
\right\rceil.
\]
Then for every $t \ge t_{\mathrm{burn}}$,
\[
\mathbb{E}[b_tb_t^{\top}]
\;\succeq\;
\sigma_{O}^2\left(\frac{7d_{\min}^{\pi}}{8}\right)^2 I.
\]

\item Consequently, for every integer $n \ge t_{\mathrm{burn}}$,
\[
\frac{1}{n-t_{\mathrm{burn}}+1}
\sum_{t=t_{\mathrm{burn}}}^{n}
\mathbb{E}[b_tb_t^{\top}]
\;\succeq\;
\sigma_{O}^2\left(\frac{7d_{\min}^{\pi}}{8}\right)^2 I,
\]
and
\[
\frac{1}{n}
\sum_{t=1}^{n}
\mathbb{E}[b_tb_t^{\top}]
\;\succeq\;
\frac{n-t_{\mathrm{burn}}+1}{n}\,
\sigma_{O}^2\left(\frac{7d_{\min}^{\pi}}{8}\right)^2 I.
\]
\end{enumerate}
\end{lemma}

\begin{proof}
We prove the claims one by one.

\textbf{The first claim}

    Fix any $t \ge 1$, and condition on $H_{t-1}$ throughout this step. For each $o \in \Omega$, define
\[
p_t(o) = \mathbb{P}(o_t=o \mid H_{t-1}).
\]
Since
\[
\bar{b}_t(s) = \mathbb{P}(s_t=s \mid H_{t-1}),
\]
we have
\[
p_t(o)
=
\sum_{s \in \mathcal{S}} \bar{b}_t(s)\,O_{\varphi^*}(o \mid s).
\]
Whenever $p_t(o)>0$, Bayes' rule gives
\[
b_t
=
\mathbb{P}(s_t=\cdot \mid H_{t-1},o_t)
=
\frac{\operatorname{Diag}(\bar{b}_t)\bigl(O_{\varphi^*}(o_t \mid s)\bigr)_{s \in \mathcal{S}}}{p_t(o_t)}.
\]
Hence
\begin{align*}
\mathbb{E}[b_tb_t^{\top} \mid H_{t-1}]
&=
\sum_{o \in \Omega} p_t(o)\,
\mathbb{P}(s_t=\cdot \mid H_{t-1},o_t=o)\,
\mathbb{P}(s_t=\cdot \mid H_{t-1},o_t=o)^{\top} \\
&=
\sum_{o:\,p_t(o)>0}
\frac{
\operatorname{Diag}(\bar{b}_t)\,
\bigl(O_{\varphi^*}(o \mid s)O_{\varphi^*}(o \mid s')\bigr)_{s,s' \in \mathcal{S}}\,
\operatorname{Diag}(\bar{b}_t)
}{p_t(o)}.
\end{align*}
Since $0 < p_t(o) \le 1$ whenever $p_t(o)>0$, we have $1/p_t(o) \ge 1$, so
\[
\mathbb{E}[b_tb_t^{\top} \mid H_{t-1}]
\;\succeq\;
\sum_{o:\,p_t(o)>0}
\operatorname{Diag}(\bar{b}_t)\,
\bigl(O_{\varphi^*}(o \mid s)O_{\varphi^*}(o \mid s')\bigr)_{s,s' \in \mathcal{S}}\,
\operatorname{Diag}(\bar{b}_t).
\]
If $p_t(o)=0$, then
\[
0
=
\sum_{s \in \mathcal{S}} \bar{b}_t(s)\,O_{\varphi^*}(o \mid s),
\]
and all terms in this sum are nonnegative, so
\[
\bar{b}_t(s)\,O_{\varphi^*}(o \mid s)=0
\qquad
\text{for every } s \in \mathcal{S}.
\]
Hence the corresponding matrix term is zero. Therefore we may sum over all $o \in \Omega$:
\[
\mathbb{E}[b_tb_t^{\top} \mid H_{t-1}]
\;\succeq\;
\sum_{o \in \Omega}
\operatorname{Diag}(\bar{b}_t)\,
\bigl(O_{\varphi^*}(o \mid s)O_{\varphi^*}(o \mid s')\bigr)_{s,s' \in \mathcal{S}}\,
\operatorname{Diag}(\bar{b}_t).
\]
Since
\[
\sum_{o \in \Omega}
\bigl(O_{\varphi^*}(o \mid s)O_{\varphi^*}(o \mid s')\bigr)_{s,s' \in \mathcal{S}}
=
O_{\varphi^*}^{\top}O_{\varphi^*},
\]
we obtain
\[
\mathbb{E}[u_tu_t^{\top} \mid H_{t-1}]
\;\succeq\;
\operatorname{Diag}(\bar{b}_t)\,O_{\varphi^*}^{\top}O_{\varphi^*}\,\operatorname{Diag}(\bar{b}_t).
\]
By Assumption~\ref{ass:alpha_undercomplete},
\[
O_{\varphi^*}^{\top}O_{\varphi^*} \succeq \sigma_{O}^2 I.
\]
Therefore
\[
\mathbb{E}[b_tb_t^{\top} \mid H_{t-1}]
\;\succeq\;
\sigma_{O}^2 \operatorname{Diag}(\bar{b}_t)^2.
\]
This proves the first claim.

\textbf{The second claim}

Taking total expectation in the previous inequality yields
\[
\mathbb{E}[b_tb_t^{\top}]
=
\mathbb{E}\bigl[\mathbb{E}[b_tb_t^{\top} \mid H_{t-1}]\bigr]
\;\succeq\;
\sigma_{O}^2\,\mathbb{E}\bigl[\operatorname{Diag}(\bar{b}_t)^2\bigr].
\]
Now $\operatorname{Diag}(\bar{b}_t)^2$ is diagonal, and its $(s,s)$ entry is $\bar{b}_t(s)^2$. Therefore
\[
\mathbb{E}\bigl[\operatorname{Diag}(\bar{b}_t)^2\bigr]
=
\operatorname{Diag}\bigl(\mathbb{E}[\bar{b}_t(s)^2]\bigr)_{s \in \mathcal{S}}.
\]
For each fixed state $s \in \mathcal{S}$,
\[
\bar{b}_t(s)
=
\mathbb{P}(s_t=s \mid H_{t-1})
=
\mathbb{E}[\mathbf{1}_{\{s_t=s\}} \mid H_{t-1}],
\]
so the tower property implies
\[
\mathbb{E}[\bar{b}_t(s)]
=
\mathbb{E}[\mathbf{1}_{\{s_t=s\}}]
=
\mathbb{P}(s_t=s)
=
d^{\pi}_t(s).
\]
Since $x \mapsto x^2$ is convex on $[0,1]$, Jensen's inequality gives
\[
\mathbb{E}[\bar{b}_t(s)^2]
\ge
\bigl(\mathbb{E}[\bar{b}_t(s)]\bigr)^2
=
d^{\pi}_t(s)^2.
\]
Hence
\[
\mathbb{E}\bigl[\operatorname{Diag}(\bar{b}_t)^2\bigr]
\;\succeq\;
\operatorname{Diag}(d^{\pi}_t)^2.
\]
Combining the previous inequalities yields
\[
\mathbb{E}[b_tb_t^{\top}]
\;\succeq\;
\sigma_{O}^2 \operatorname{Diag}(d^{\pi}_t)^2.
\]
Since
\[
\operatorname{Diag}(d^{\pi}_t)^2 \succeq (d^{\pi}_{t,\min})^2 I,
\]
we conclude that
\[
\mathbb{E}[b_tb_t^{\top}]
\;\succeq\;
\sigma_{O}^2 (d^{\pi}_{t,\min})^2 I.
\]
This proves the second claim.

\textbf{The third claim}

We now derive a lower bound on $d^{\pi}_{t,\min}$ after a sufficiently long burn-in period. Because $(s_t)_{t \ge 1}$ is a Markov chain with transition kernel $\mathbb{P}^{\pi}$, and conditioning on the initial state, we have
\[
d^{\pi}_t(s')
=
\sum_{s \in \mathcal{S}}
\mathbb{P}^{\pi}(s_1=s)\,
\mathbb{P}^{\pi}(s_t = s' \mid s_1=s),
\]
where we abuse the notation and let $\mathbb{P}^{\pi}$ denote the probability measure under the Markov chain induced by $\pi$.

Using the convexity of the total variation norm in the first argument and Corollary~\ref{cor: geometric mixing},
\begin{align*}
\|d^{\pi}_t - d^{\pi}\|_{\mathrm{TV}}
&=
\left\|
\sum_{s \in \mathcal{S}}
\mathbb{P}^{\pi}(s_1=s)\,
\bigl(\mathbb{P}^{\pi}(s_t = \cdot \mid s_1=s)-d^{\pi}\bigr)
\right\|_{\mathrm{TV}} \\
&\le
\sum_{s \in \mathcal{S}}
\mathbb{P}^{\pi}(s_1=s)\,
\left\|
\mathbb{P}^{\pi}(s_t = \cdot \mid s_1=s)-d^{\pi}
\right\|_{\mathrm{TV}} \\
&\le
G(\pi)\,\rho(\pi)^{\,t-1}.
\end{align*}
For any fixed state $s \in \mathcal{S}$, taking the set $A=\{s\}$ in the definition of total variation in Corollary~\ref{cor: geometric mixing}
gives
\[
|d^{\pi}_t(s)-d^{\pi}(s)|
\le
\|d^{\pi}_t-d^{\pi}\|_{\mathrm{TV}}.
\]
Therefore
\[
d^{\pi}_t(s)
\ge
d^{\pi}(s)-\|d^{\pi}_t-d^{\pi}\|_{\mathrm{TV}}
\ge
d_{\min}^{\pi}-G(\pi)\,\rho(\pi)^{\,t-1}.
\]
Thus we can require
\[
G(\pi)\,\rho(\pi)^{\,t-1}
\le
\frac{d_{\min}^{\pi}}{8}
\]
in order to guarantee that
\[
d^{\pi}_t(s)\ge \frac{7d_{\min}^{\pi}}{8}
\qquad
\text{for every } s \in \mathcal{S}.
\]
Because $0 < \rho(\pi) < 1$, this inequality is equivalent to
\[
t
\ge
1+
\frac{\log\!\bigl(8G(\pi)/d_{\min}^{\pi}\bigr)}
{\log\!\bigl(1/\rho(\pi)\bigr)}.
\]
Therefore, if
\[
t_{\mathrm{burn}}
=
1+
\left\lceil
\frac{\log\!\bigl(8G(\pi)/d_{\min}^{\pi}\bigr)}
{\log\!\bigl(1/\rho(\pi)\bigr)}
\right\rceil,
\]
then for every $t \ge t_{\mathrm{burn}}$,
\[
d^{\pi}_{t,\min}
=
\min_{s \in \mathcal{S}} d^{\pi}_t(s)
\ge
\frac{7d_{\min}^{\pi}}{8}.
\]
Substituting this into the inequality from the second claim gives
\[
\mathbb{E}[b_tb_t^{\top}]
\;\succeq\;
\sigma_{O}^2\left(\frac{7d_{\min}^{\pi}}{8}\right)^2 I
\qquad
\text{for every } t \ge t_{\mathrm{burn}}.
\]
This proves the third claim.

\textbf{The fourth claim}

For every $t \ge t_{\mathrm{burn}}$, we have already shown that
\[
\mathbb{E}[b_tb_t^{\top}]
\;\succeq\;
\sigma_{O}^2\left(\frac{7d_{\min}^{\pi}}{8}\right)^2 I.
\]
Averaging over $t=t_{\mathrm{burn}},\dots,n$ gives
\[
\frac{1}{n-t_{\mathrm{burn}}+1}
\sum_{t=t_{\mathrm{burn}}}^{n}
\mathbb{E}[b_tb_t^{\top}]
\;\succeq\;
\sigma_{O}^2\left(\frac{7d_{\min}^{\pi}}{8}\right)^2 I.
\]
This proves the first average bound.

For the full average from $t=1$ to $t=n$, note that each matrix $\mathbb{E}[b_tb_t^{\top}]$ is positive semidefinite. Therefore
\[
\frac{1}{n}
\sum_{t=1}^{n}
\mathbb{E}[b_tb_t^{\top}]
\;\succeq\;
\frac{1}{n}
\sum_{t=t_{\mathrm{burn}}}^{n}
\mathbb{E}[b_tb_t^{\top}]
=
\frac{n-t_{\mathrm{burn}}+1}{n}
\left(
\frac{1}{n-t_{\mathrm{burn}}+1}
\sum_{t=t_{\mathrm{burn}}}^{n}
\mathbb{E}[b_tb_t^{\top}]
\right).
\]
Using the previous inequality,
\[
\frac{1}{n}
\sum_{t=1}^{n}
\mathbb{E}[b_tb_t^{\top}]
\;\succeq\;
\frac{n-t_{\mathrm{burn}}+1}{n}\,
\sigma_{O}^2\left(\frac{7d_{\min}^{\pi}}{8}\right)^2 I.
\]
This proves the fourth claim.
\end{proof}

\begin{lemma}[Transport of isotropy from the filtered belief to the predictive belief]\label{lemma: predictive belief isotropy}
For every \(t\ge t_{\mathrm{burn}}+1\),
\[
\mathbb E[\bar b_t\bar b_t^\top]
\succeq
|\mathcal A|\,\pi_{\min}\,\sigma_{T}^2\,\sigma_{O}^2
\left(\frac{7d_{\min}^{\pi}}{8}\right)^2 I.
\]
\end{lemma}

\begin{proof}
For every \(t\ge 1\), the predictive belief satisfies
\[
\bar b_{t+1}=T^{a_t}_{\beta^*}b_t.
\]
Fix \(t\ge t_{\mathrm{burn}}\). For any \(x\in\mathbb R^{|\mathcal S|}\),
\[
x^\top \mathbb E[\bar b_{t+1}\bar b_{t+1}^\top]x
=
\mathbb E\bigl[(x^\top \bar b_{t+1})^2\bigr]
=
\mathbb E\bigl[(x^\top T^{a_t}_{\beta^*}b_t)^2\bigr].
\]

Let
\[
\mathcal G_t:=\sigma(H_{t-1},o_t).
\]
By definition, \(b_t\) is \(\mathcal G_t\)-measurable. Therefore, for every fixed \(a\in\mathcal A\), the quantity
\[
(x^\top T_{\beta^*}^ab_t)^2
\]
is also \(\mathcal G_t\)-measurable. Therefore, we have
\begin{equation}\label{equ:conditional probability of a is at least pi min}
\mathbb P(a_t=a\mid \mathcal G_t)=\sum_{s}b_{t}(s)\sum_{o}\pi(a\mid o)O_{\varphi^*}(o|s)\geq\pi_{\min}\sum_{s}b_{t}(s)= \pi_{\min}
\qquad\text{for every }a\in\mathcal A,
\end{equation}
where the inequality comes from Assumption~\ref{ass: minimum policy mass}.

Using the law of total expectation, we obtain
\[
\mathbb E\bigl[(x^\top T^{a_t}_{\beta^*}b_t)^2\bigr]
=
\mathbb E\!\left[
\mathbb E\bigl[(x^\top T^{a_t}_{\beta^*}b_t)^2\mid \mathcal G_t\bigr]
\right].
\]
Now
\[
(x^\top T^{a_t}_{\beta^*}b_t)^2
=
\sum_{a\in\mathcal A}\mathbf 1\{a_t=a\}(x^\top T^{a}_{\beta^*}b_t)^2,
\]
so
\[
\mathbb E\bigl[(x^\top T^{a_t}_{\beta^*}b_t)^2\mid \mathcal G_t\bigr]
=
\sum_{a\in\mathcal A}
\mathbb E\bigl[\mathbf 1\{a_t=a\}(x^\top T^{a}_{\beta^*}b_t)^2\mid \mathcal G_t\bigr].
\]
Because \((x^\top T^{a}_{\beta^*}b_t)^2\) is \(\mathcal G_t\)-measurable, it can be taken outside the conditional expectation:
\[
\mathbb E\bigl[(x^\top T^{a_t}_{\beta^*}b_t)^2\mid \mathcal G_t\bigr]
=
\sum_{a\in\mathcal A}
(x^\top T^{a}_{\beta^*}b_t)^2
\mathbb E[\mathbf 1\{a_t=a\}\mid \mathcal G_t].
\]
Since $\mathbb E[\mathbf 1\{a_t=a\}\mid \mathcal G_t]=\mathbb P(a_t=a\mid \mathcal G_t)\geq \pi_{\min}$ from \eqref{equ:conditional probability of a is at least pi min},
\[
\mathbb E\bigl[(x^\top T^{a_t}_{\beta^*}b_t)^2\mid \mathcal G_t\bigr]
\geq
\pi_{\min}\sum_{a\in\mathcal A}
(x^\top T^{a}_{\beta^*}b_t)^2.
\]

Therefore,
\[
\mathbb E\bigl[(x^\top T^{a_t}_{\beta^*}b_t)^2\bigr]
\ge
\pi_{\min}\sum_{a\in\mathcal A}\mathbb E\bigl[(x^\top T^{a}_{\beta^*}b_t)^2\bigr].
\]
Hence
\[
x^\top \mathbb E[\bar b_{t+1}\bar b_{t+1}^\top]x
\ge
\pi_{\min}\sum_{a\in\mathcal A}\mathbb E\bigl[(x^\top T^{a}_{\beta^*}b_t)^2\bigr].
\]
Since each \(T^{a}_{\beta^*}\) is deterministic,
\[
\mathbb E\bigl[(x^\top T^{a}_{\beta^*}b_t)^2\bigr]
=
x^\top T^{a}_{\beta^*}\,\mathbb E[b_tb_t^\top]\,(T^{a}_{\beta^*})^\top x.
\]
Therefore
\[
x^\top \mathbb E[\bar b_{t+1}\bar b_{t+1}^\top]x
\ge
\pi_{\min}\sum_{a\in\mathcal A}
x^\top T^{a}_{\beta^*}\,\mathbb E[b_tb_t^\top]\,(T^{a}_{\beta^*})^\top x.
\]
Using the lower bound from Lemma~\ref{lem: filtered belief isotropy}
\[
\mathbb E[b_tb_t^\top]\succeq \sigma_{O}^2\left(\frac{7d_{\min}^{\pi}}{8}\right)^2 I,
\]
we conclude that
\[
x^\top \mathbb E[\bar b_{t+1}\bar b_{t+1}^\top]x
\ge
\pi_{\min}\sigma_{O}^2\left(\frac{7d_{\min}^{\pi}}{8}\right)^2
\sum_{a\in\mathcal A}x^\top T^{a}_{\beta^*}(T^{a}_{\beta^*})^\top x.
\]
Since this holds for every \(x\in\mathbb R^{|\mathcal S|}\), we obtain the exact matrix lower bound
\begin{equation}\label{equ:before using the sigma T assumption}
\mathbb E[\bar b_{t+1}\bar b_{t+1}^\top]
\succeq
\pi_{\min}\sigma_{O}^2\left(\frac{7d_{\min}^{\pi}}{8}\right)^2
\sum_{a\in\mathcal A} T^{a}_{\beta^*}(T^{a}_{\beta^*})^\top.
\end{equation}

Now, Assumption~\ref{ass: full-rank transition kernels} gives
\[
\sum_{a\in\mathcal A}T^{a}_{\beta^*}(T^{a}_{\beta^*})^\top
\succeq
|\mathcal A|\,\sigma_{T}^2 I.
\]
Substituting this into the exact matrix lower bound yields
\[
\mathbb E[\bar b_{t+1}\bar b_{t+1}^\top]
\succeq
|\mathcal A|\,\pi_{\min}\,\sigma_{T}^2\,\sigma_{O}^2
\left(\frac{7d_{\min}^{\pi}}{8}\right)^2 I.
\]
\end{proof}

\begin{corollary}\label{cor: predictive belief isotropy after burn-in}
For every integer \(n\ge t_{\mathrm{burn}}+1\),
\[
\frac{1}{n-t_{\mathrm{burn}}}
\sum_{t=t_{\mathrm{burn}}+1}^{n}\mathbb E[\bar b_t\bar b_t^\top]
\succeq
|\mathcal A|\,\pi_{\min}\,\sigma_{T}^2\,\sigma_{O}^2
\left(\frac{7d_{\min}^{\pi}}{8}\right)^2 I,
\]
and
\[
\frac{1}{n}\sum_{t=1}^{n}\mathbb E[\bar b_t\bar b_t^\top]
\succeq
\frac{n-t_{\mathrm{burn}}}{n}\,
|\mathcal A|\,\pi_{\min}\,\sigma_{T}^2\,\sigma_{O}^2
\left(\frac{7d_{\min}^{\pi}}{8}\right)^2 I.
\]
\end{corollary}

\begin{proof}
This is immediate from the pointwise lower bound from Lemma~\ref{lemma: predictive belief isotropy}:
\[
\mathbb E[\bar b_t\bar b_t^\top]
\succeq
|\mathcal A|\,\pi_{\min}\,\sigma_{T}^2\,\sigma_{O}^2
\left(\frac{7d_{\min}^{\pi}}{8}\right)^2 I
\qquad\text{for every }t\ge t_{\mathrm{burn}}+1.
\]
\end{proof}

\subsection{}\label{subapx: auxiliary facts for high-prob sigma B}
In Appendix~\ref{subapx: sigma B in expectation}, we have proved several results regarding the properties of $\mathbb{E}[\bar{b}_{t}\bar{b}_{t}^\top]$. The next step is to convert these results into a high probability bound on $\sigma_{\min}(\bar{B})$. In this subsection, we introduce two useful concepts and then establish a number of ``helper" lemmas and propositions using the two concepts before finishing the proof of the lemma in Appendix~\ref{subapx: main proof of high-prob sigma B}.

First, we define the concept of $\beta$-mixing (also known as absolute regularity) and relate it to the mixing condition of the Markov chain induced by $\pi$ described in Corollary~\ref{cor: geometric mixing}.

For two sigma-fields $\mathcal{F}$ and $\mathcal{G}$, define their absolute-regularity coefficient by
\[
\beta(\mathcal{F},\mathcal{G})
=
\sup \frac{1}{2}\sum_{i=1}^{I}\sum_{j=1}^{J}
\left|
\mathbb{P}(F_i\cap G_j)-\mathbb{P}(F_i)\mathbb{P}(G_j)
\right|,
\]
where the supremum is over all finite measurable partitions $\{F_i\}_{i=1}^{I}\subseteq \mathcal{F}$ and $\{G_j\}_{j=1}^{J}\subseteq \mathcal{G}$.

For a sequence $(R_t)$, define its absolute-regularity coefficient by
\[
\beta_R(\ell)
=
\sup_{r\geq 1}
\beta\!\Big(
\sigma(R_1,\dots,R_r),
\sigma(R_{r+\ell},R_{r+\ell+1},\dots)
\Big).
\]

The random sequence $(R_{t})$ is said to be $\beta$-mixing (or absolutely regular) if $\beta_{R}(\ell)\rightarrow 0$ when $\ell\rightarrow \infty$ \citep{volkonskii1959some,volkonskii1961some}.

Second, we define the concept of $\gamma$-separability.

Initially introduced in \citet{even2007value}, under the name ``value of observation" and in the context of HMMs, \citet{golowich2022planning} defined a non-stationary POMDP to be $\gamma$-observable if for every $O_{h}$, the observation kernel at time $h$, for any beliefs $b$, $b'$,
\begin{equation*}
    \|O_{h}^\top b-O_{h}b'\|_{1}\geq \gamma\|b-b'\|_{1}.
\end{equation*}

Since a stationary POMDP is a special case of non-stationary POMDPs, we can similarly define under our notation a stationary POMDP to be $\gamma$-observable if for any beliefs $b$, $b'$,
\begin{equation}\label{equ: gamma-observability}
    \|O_{\varphi^*}^\top b-O_{\varphi^*}b'\|_{1}\geq \gamma\|b-b'\|_{1}.
\end{equation}

Now, we introduce the ``helper" lemmas and propositions.

\begin{proposition}\label{prop: uniform mixing implies beta-mixing}
    From \citet{bradley2005basic}, Corollary~\ref{cor: geometric mixing} implies that for every $t\geq 1$,
\begin{equation*}
    \beta_{\mathcal{S}}(t)\leq G(\pi)\rho(\pi)^{t-1},
\end{equation*}
where we abuse the notation and denote $\mathcal{S}$ as the random sequence of latent states. In other words, the Markov chain induced by $\pi$ is $\beta$-mixing.
\end{proposition}
In other words, Assumption~\ref{ass: ergodicity} implies Corollary~\ref{cor: geometric mixing}, which in turn implies $\beta$-mixing for the induced Markov chain.

\begin{lemma}\label{lem: full-rank implies observability}
    If Assumption~\ref{ass: full-rank transition kernels} holds, then the POMDP is $\frac{\sigma_{O}}{\sqrt{|\mathcal{S}|}}$-observable.
\end{lemma}
\begin{proof}
    Under Assumption~\ref{ass: full-rank transition kernels}, $\sigma_{\min}(O_{\varphi^*})\geq \sigma_{O} > 0$, which means for any nonzero $d\in \mathbb{R}^{|\mathcal{S}|}$,
    \[\|O_{\varphi^*}d\|_{2}\geq \sigma_{O}\|d\|_{2}.\]

    In particular, taking $d=b-b'$ for any beliefs $b$ and $b'$:
    \[\|O_{\varphi^*}b-O_{\varphi^*}b'\|_{2}\geq \sigma_{O}\|b-b'\|_{2}.\]
    Therefore,
    \[\|O_{\varphi^*}b-O_{\varphi^*}b'\|_{1}\geq\|O_{\varphi^*}b-O_{\varphi^*}b'\|_{2}\geq \sigma_{O}\|b-b'\|_{2}\geq \frac{\sigma_{O}}{\sqrt{|\mathcal{S}|}}\|b-b'\|_{1}.\]
\end{proof}

\begin{lemma}[Monotonicity of absolute regularity under sigma-field inclusion]
\label{lem:beta-monotonicity}
If $\mathcal{F}_1 \subseteq \mathcal{F}_2$ and $\mathcal{G}_1 \subseteq \mathcal{G}_2$, then
\[
\beta(\mathcal{F}_1,\mathcal{G}_1)\leq \beta(\mathcal{F}_2,\mathcal{G}_2).
\]
\end{lemma}

\begin{proof}
Every finite measurable partition admissible for the pair $(\mathcal{F}_1,\mathcal{G}_1)$ is also admissible for the pair $(\mathcal{F}_2,\mathcal{G}_2)$. The result follows immediately from the definition of $\beta(\mathcal{F},\mathcal{G})$.
\end{proof}

\begin{proposition}[Mixing of the observation-action process]
\label{prop:mixing-W}
Proposition~\ref{prop: uniform mixing implies beta-mixing} implies that the process
\[
W_t = (o_t,a_t)
\]
is geometrically $\beta$-mixing. More precisely,
\[
\beta_W(\ell)
\leq \beta_S(\ell-1)
\leq G(\pi)\rho(\pi)^{\ell-2},
\qquad \ell\geq 2.
\]

Together with $\beta_W(1)\leq 1$, this yields
\[
\beta_W(\ell)
\leq \widetilde G(\pi)\rho(\pi)^{\ell-1},
\qquad \ell\geq 1,
\]
where
$\widetilde G(\pi)
=\max\{1,G(\pi)/\rho(\pi)\}$
for $0<\rho(\pi)<1$.
\end{proposition}

\begin{proof}
Fix $r\geq 1$ and $\ell\geq 2$.
Since the behavior policy is memoryless, the latent
state process $(s_t)$ is Markov. Moreover, conditional
on the current state, the future observation-action
process is independent of the past. Consequently,
\[
(W_1,\ldots,W_r)
\longrightarrow s_{r+1}
\longrightarrow s_{r+\ell}
\longrightarrow (W_{r+\ell},W_{r+\ell+1},\ldots)
\]
forms a Markov chain.

By the data-processing inequality for absolute
regularity,
\begin{align*}
&\beta\bigl(
\sigma(W_1,\ldots,W_r),
\sigma(W_{r+\ell},W_{r+\ell+1},\ldots)
\bigr)\\
&\qquad\leq
\beta\bigl(
\sigma(s_{r+1}),\sigma(s_{r+\ell})
\bigr)\\
&\qquad\leq \beta_S(\ell-1),
\end{align*}
where the last inequality follows from the definition
of $\beta_S(\ell-1)$.

Taking the supremum over $r$ and applying
Proposition~\ref{prop: uniform mixing implies beta-mixing} gives
\[
\beta_W(\ell)
\leq \beta_S(\ell-1)
\leq G(\pi)\rho(\pi)^{\ell-2},
\qquad \ell\geq 2.
\]
Since $\beta_W(1)\leq 1$, the claimed geometric
bound follows with
$\widetilde G(\pi)
=\max\{1,G(\pi)/\rho(\pi)\}$.
\end{proof}

\begin{corollary}[Belief contraction for all history-dependent policies]\label{cor: belief contraction for all history-dependent policies}
Consider any $\gamma$-observable POMDP $\mathcal P$, any $\epsilon>0$, and any $L\in\mathbb N$ satisfying the lower bound in Theorem 2.1 of \citet{golowich2022learning}. Let $\mu$ be any nonanticipatory history-dependent randomized policy, meaning that for each $h$ and each history $(a_{1:h-1},o_{2:h})$, the policy chooses an action according to a distribution
\[
\mu_h(\cdot\mid a_{1:h-1},o_{2:h})\in\Delta(\mathcal A).
\]
Suppose that $\mathcal D\in\Delta^{\mathcal S}$ satisfies the same domination condition as in Theorem 2.1, namely
\[
\frac{\mathbf b_h^{\mathcal P}(a_{1:h-L-1},o_{2:h-L})(s)}{\mathcal D(s)}
\leq \frac{1}{\phi}
\]
for every relevant history and every $s\in\mathcal S$. Then
\[
\mathbb E^{\mathcal P}_{(a_{1:h-1},o_{2:h})\sim \mu}
\left[
\left\|
\mathbf b_h^{\mathcal P}(a_{1:h-1},o_{2:h})
-
\mathbf b_h^{\mathrm{apx},\mathcal P}(a_{h-L:h-1},o_{h-L+1:h};\mathcal D)
\right\|_1
\right]
\leq \epsilon,
\]

where $b_h^{\mathcal P}$ and $b_h^{\mathrm{apx},\mathcal P}$ are defined in \citet{golowich2022learning}.
\end{corollary}

\begin{proof}
We show that the trajectory distribution induced by $\mu$ can be represented by some general policy in $\Pi^{\mathrm{gen}}$. The policy class $\Pi^{\mathrm{gen}}$ is defined in Section 2.2 of \citet{golowich2022learning}: each $\pi\in\Pi^{\mathrm{gen}}$ is a distribution over deterministic history-dependent policies, and is executed by first sampling $\sigma\sim\pi$ and then following $\sigma$ for the entire episode. For every deterministic policy
\[
\sigma=(\sigma_1,\ldots,\sigma_{H-1}),
\qquad
\sigma_h:\mathcal A^{h-1}\times\Omega^{h-1}\to\mathcal A,
\]
define a distribution $\pi_\mu$ over deterministic policies by sampling, independently for every $h$ and every history $(a_{1:h-1},o_{2:h})$,
\[
\sigma_h(a_{1:h-1},o_{2:h})
\sim
\mu_h(\cdot\mid a_{1:h-1},o_{2:h}).
\]
Then $\pi_\mu\in\Pi^{\mathrm{gen}}$.

Fix any trajectory $(s_{1:H},a_{1:H-1},o_{2:H})$. Under the stagewise randomized policy $\mu$, its probability is
\[
\mathbb P^{\mathcal P}_{\mu}(s_{1:H},a_{1:H-1},o_{2:H})
=
\rho(s_1)
\prod_{h=1}^{H-1}
\mu_h(a_h\mid a_{1:h-1},o_{2:h})
T_{\beta^*}^{a_{h}}(s_{h+1}\mid s_h)
O_{\varphi^*}(o_{h+1}\mid s_{h+1}).
\]
On the other hand, under the general policy $\pi_\mu$, a deterministic policy $\sigma$ is first sampled from $\pi_\mu$ and then followed for the whole episode. Therefore,
\begin{align*}
&\mathbb P^{\mathcal P}_{\pi_\mu}(s_{1:H},a_{1:H-1},o_{2:H})\\
=
&\rho(s_1)
\prod_{h=1}^{H-1}
T_{\beta^*}^{a_{h}}(s_{h+1}\mid s_h)
O_{\varphi^*}(o_{h+1}\mid s_{h+1})
\mathbb P_{\sigma\sim\pi_\mu}
\left(
\sigma_h(a_{1:h-1},o_{2:h})=a_h
\text{ for }h=1,\ldots,H-1
\right).
\end{align*}
By the construction of $\pi_\mu$, the sampled actions assigned to the histories appearing along the trajectory are independent, and hence
\[
\mathbb P_{\sigma\sim\pi_\mu}
\left(
\sigma_h(a_{1:h-1},o_{2:h})=a_h
\text{ for all }h=1,\ldots,H-1
\right)
=
\prod_{h=1}^{H-1}
\mu_h(a_h\mid a_{1:h-1},o_{2:h}).
\]
Thus,
\[
\mathbb P^{\mathcal P}_{\pi_\mu}(s_{1:H},a_{1:H-1},o_{2:H})
=
\mathbb P^{\mathcal P}_{\mu}(s_{1:H},a_{1:H-1},o_{2:H})
\]
for every trajectory. Therefore, the two policies induce the same distribution over action-observation histories. Consequently,
\[
\mathbb E^{\mathcal P}_{(a_{1:h-1},o_{2:h})\sim \mu}
\left[
\left\|
\mathbf b_h^{\mathcal P}(a_{1:h-1},o_{2:h})
-
\mathbf b_h^{\mathrm{apx},\mathcal P}(a_{h-L:h-1},o_{h-L+1:h};\mathcal D)
\right\|_1
\right]
\]
equals
\[
\mathbb E^{\mathcal P}_{(a_{1:h-1},o_{2:h})\sim \pi_\mu}
\left[
\left\|
\mathbf b_h^{\mathcal P}(a_{1:h-1},o_{2:h})
-
\mathbf b_h^{\mathrm{apx},\mathcal P}(a_{h-L:h-1},o_{h-L+1:h};\mathcal D)
\right\|_1
\right].
\]
Since $\pi_\mu\in\Pi^{\mathrm{gen}}$, Theorem 2.1 of \citet{golowich2022learning} applies to $\pi_\mu$, and the last display is at most $\epsilon$. This proves the claim.
\end{proof}

\begin{proposition}[Finite-window approximation of the predictive belief]
\label{prop:predictive-truncation}
Given Lemma~\ref{lem: full-rank implies observability}, there exists a constant $C\geq 1$ such that for every integer $L\geq 2$, there exists a measurable function
\[
g_{t}^{L} : (\Omega\times \mathcal{A})^L \to \Delta(\mathcal{S})
\]
such that
\[
\bar b_t^{(L)}
=
g_{t}^{L}(W_{t-L},W_{t-L+1},\dots,W_{t-1})
\]
satisfies
\[
\mathbb{E}\!\left[\|\bar b_t-\bar b_t^{(L)}\|_1\right]
\leq
\min\left(|\mathcal{S}| \exp\!\left(-\frac{\sigma_{O}^4}{C|\mathcal{S}|^{2}}L\right),\sqrt{|\mathcal{S}|\log(|\mathcal{S}|)}\;\exp\!\left(-\sigma_{O}\sqrt{\frac{L}{C|\mathcal{S}|}}\right)\right).
\]
\end{proposition}

\begin{proof}
The proof of this proposition combines the belief contraction result of Theorem~2.1 from \citet{golowich2022learning} and Corollary~\ref{cor: belief contraction for all history-dependent policies}.

We divide the proof into two cases.

\textbf{Case 1: $t\geq L+1$}

Let
\[
D = \mathrm{Unif}(\mathcal{S}),
\qquad
D(s)=\frac{1}{|\mathcal{S}|}
\quad\text{for every }s\in \mathcal{S}.
\]
For any belief vector $b\in \Delta(\mathcal{S})$ and any state $s\in\mathcal{S}$,
\[
\frac{b(s)}{D(s)} = |\mathcal{S}|\, b(s)\leq |\mathcal{S}|.
\]
Therefore the coverage condition in Theorem~2.1 of \cite{golowich2022learning} is automatically satisfied with
\[
\phi = \frac{1}{|\mathcal{S}|}.
\]

For any integer $L\geq 1$, define the truncated filtered belief
\[
b_t^{(L)}
=
b_t^{\mathrm{apx}}(a_{t-L:t-1},o_{t-L+1:t};D),
\]
where $b_t^{\mathrm{apx}}$ is the approximate belief state from Theorem~2.1 of \cite{golowich2022learning} which computes an approximation of the true filtered belief by using only the observations and actions from the most recent length-L window of the history and assuming an initial belief of $D$ at the start of the length-L window.

Furthermore, recall from Lemma~\ref{lem: full-rank implies observability} that the POMDP satisfies the observability condition with the constant $\frac{\sigma_{O}}{\sqrt{|\mathcal{S}|}}$.

Then, if we choose
\[
\varepsilon = |\mathcal{S}| \exp\!\left(-\frac{\sigma_{O}^4}{C|\mathcal{S}|^{2}}L\right),
\]
where the constant $C$ is the same as the constant in Theorem~2.1 of \citet{golowich2022learning}. Then
\[
\varepsilon \phi = \exp\!\left(-\frac{\sigma_{O}^4}{C|\mathcal{S}|^{2}}L\right),
\]
and therefore
\[
L
=
C|\mathcal{S}|^{2} \cdot \frac{\log(1/(\varepsilon \phi))}{\sigma_{O}^4},
\]
which means $L$ exceeds the lower bound required for Theorem~2.1 of \citet{golowich2022learning} to hold.

There is another choice of $(\varepsilon,L)$, which we also include for completeness.

Let
\[
L \;=\; C\frac{|\mathcal{S}|}{\sigma_{O}^{2}}\log\!\frac{1}{\epsilon\phi}\,\cdot\,\log\!\frac{\log(1/\phi)}{\epsilon}
\]

so that $L$ exceeds the lower bound required for Theorem~2.1 of \citet{golowich2022learning}.

Furthermore, assume $\phi \in (0, 1/e)$ (so $\log(1/\phi) > 1$ and hence $\log\log(1/\phi) > 0$) $\epsilon \in (0,1)$. Note that assuming $\phi \in (0, 1/e)$ is without loss of generality since $\phi=\frac{1}{|\mathcal{S}|}$, and the only case where $\phi\notin(0,1/e)$ is when $|\mathcal{S}|=1$ or $|\mathcal{S}|=2$, in which case we set $\phi=\frac{1}{3}$, which is still a valid choice of $\phi$ that fulfills the coverage assumption of Theorem~2.1 of \citet{golowich2022learning}.

Using $\log(1/(\epsilon\phi)) = \log(1/\epsilon) + \log(1/\phi)$ and $\log(\log(1/\phi)/\epsilon) = \log(1/\epsilon) + \log\log(1/\phi)$, the equation becomes
\[
\bigl(\log(1/\epsilon) + \log(1/\phi)\bigr)\bigl(\log(1/\epsilon) + \log\log(1/\phi)\bigr) \;=\; L\frac{\sigma_{O}^{2}}{C|\mathcal{S}|},
\]
which is quadratic in $\log(1/\epsilon)$.

Expanding,
\[
\bigl(\log(1/\epsilon)\bigr)^2 + \bigl(\log(1/\phi) + \log\log(1/\phi)\bigr)\log(1/\epsilon) + \bigl(\log(1/\phi)\cdot\log\log(1/\phi) - L\frac{\sigma_{O}^{2}}{C|\mathcal{S}|}\bigr) = 0.
\]

Using the identity $(p+q)^2 - 4pq = (p-q)^2$ with $p = \log(1/\phi)$, $q = \log\log(1/\phi)$,
\begin{align*}
\Delta &\;=\; \bigl(\log(1/\phi) + \log\log(1/\phi)\bigr)^2 - 4\bigl(\log(1/\phi)\cdot\log\log(1/\phi) - L\frac{\sigma_{O}^{2}}{C|\mathcal{S}|}\bigr)\\
&\;=\; \bigl(\log(1/\phi) - \log\log(1/\phi)\bigr)^2 + 4L\frac{\sigma_{O}^{2}}{C|\mathcal{S}|} \;\geq\; 0.
\end{align*}

The quadratic formula gives two algebraic solutions:
\[
\log(1/\epsilon) \;=\; \frac{-\bigl(\log(1/\phi) + \log\log(1/\phi)\bigr) \;\pm\; \sqrt{\bigl(\log(1/\phi) - \log\log(1/\phi)\bigr)^2 + 4L\frac{\sigma_{O}^{2}}{C|\mathcal{S}|}}}{2}.
\]

Only the $+$ root is admissible. For the $-$ root, since
\[
\sqrt{\bigl(\log(1/\phi) - \log\log(1/\phi)\bigr)^2 + 4L\frac{\sigma_{O}^{2}}{C|\mathcal{S}|}} \;\geq\; \bigl|\log(1/\phi) - \log\log(1/\phi)\bigr| \;=\; \log(1/\phi) - \log\log(1/\phi),
\]
which further implies
\[
\log(1/\epsilon_-) \;\leq\; \frac{-\bigl(\log(1/\phi) + \log\log(1/\phi)\bigr) - \bigl(\log(1/\phi) - \log\log(1/\phi)\bigr)}{2} \;=\; -\log(1/\phi),
\]
so $\epsilon_- \geq 1/\phi > 1$, violating $\epsilon \in (0,1)$. This $-$ root corresponds to the spurious regime where both factors of the original product are negative (their product still equals $L\gamma^2 > 0$, but this sign pattern is ruled out by $\epsilon \in (0,1)$).

Taking the $+$ root,
\[
\log(1/\epsilon) \;=\; \frac{-\bigl(\log(1/\phi) + \log\log(1/\phi)\bigr) + \sqrt{\bigl(\log(1/\phi) - \log\log(1/\phi)\bigr)^2 + 4L\frac{\sigma_{O}^{2}}{C|\mathcal{S}|}}}{2},
\]
so, exponentiating,
\[
\epsilon \;=\; \exp\!\left(\frac{\log(1/\phi) + \log\log(1/\phi)}{2}\right) \exp\!\left(-\tfrac{1}{2}\sqrt{\bigl(\log(1/\phi) - \log\log(1/\phi)\bigr)^2 + 4L\frac{\sigma_{O}^{2}}{C|\mathcal{S}|}}\right).
\]
The prefactor simplifies via
\[
\exp\!\left(\frac{\log(1/\phi) + \log\log(1/\phi)}{2}\right) = \frac{1}{\sqrt{\phi}}\cdot\sqrt{\log(1/\phi)} = \sqrt{\frac{\log(1/\phi)}{\phi}},
\]
giving
\[
\;\epsilon \;=\; \sqrt{\frac{\log(1/\phi)}{\phi}}\;\exp\!\left(-\tfrac{1}{2}\sqrt{\bigl(\log(1/\phi) - \log\log(1/\phi)\bigr)^2 + 4L\frac{\sigma_{O}^{2}}{C|\mathcal{S}|}}\right).\;
\]

Since $\bigl(\log(1/\phi) - \log\log(1/\phi)\bigr)^2 \geq 0$,
\[
\sqrt{\bigl(\log(1/\phi) - \log\log(1/\phi)\bigr)^2 + 4L\frac{\sigma_{O}^{2}}{|\mathcal{S}|}} \;\geq\; 2\sigma_{O}\sqrt{\frac{L}{C|\mathcal{S}|}},
\]
hence
\[
\;\epsilon \;\leq\; \sqrt{\frac{\log(1/\phi)}{\phi}}\;\exp\!\left(-\sigma_{O}\sqrt{\frac{L}{C|\mathcal{S}|}}\right)=\sqrt{|\mathcal{S}|\log(|\mathcal{S}|)}\;\exp\!\left(-\sigma_{O}\sqrt{\frac{L}{C|\mathcal{S}|}}\right).\;
\]

To summarize, Theorem~2.1 of \citet{golowich2022learning} and Corollary~\ref{cor: belief contraction for all history-dependent policies} gives us that, our two choices of $(L,\epsilon)$ imply
\begin{align*}
\mathbb{E}\!\left[\|b_t-b_t^{(L)}\|_1\right]
\leq
\min\left(|\mathcal{S}| \exp\!\left(-\frac{\sigma_{O}^4}{C|\mathcal{S}|^{2}}L\right),\sqrt{|\mathcal{S}|\log(|\mathcal{S}|)}\;\exp\!\left(-\sigma_{O}\sqrt{\frac{L}{C|\mathcal{S}|}}\right)\right)\\
\qquad\text{for every }t\geq L+1,
\end{align*}
where the randomness in the expectation of the left-hand side comes from the randomness from POMDP dynamics and the randomness from the behavior policy $\pi$.

Now define, for every $L\geq 2$ and $t\geq L+1$,
\[
\bar b_t^{(L)}
=
T^{a_{t-1}}_{\beta^*} b_{t-1}^{(L-1)}.
\]
Since $b_{t-1}^{(L-1)}$ depends only on
\[
a_{t-L:t-2}
\quad\text{and}\quad
o_{t-L+1:t-1},
\]
the vector $\bar b_t^{(L)}$ depends only on
\[
a_{t-L:t-1}
\quad\text{and}\quad
o_{t-L+1:t-1},
\]
and hence it is a measurable function of
\[
W_{t-L},W_{t-L+1},\dots,W_{t-1}.
\]
This gives the measurable map $g_{t}^{L}$.

Finally,
\[
\bar b_t = T^{a_{t-1}}_{\beta^*} b_{t-1},
\]
so
\[
\|\bar b_t-\bar b_t^{(L)}\|_1
=
\|T^{a_{t-1}}_{\beta^*}(b_{t-1}-b_{t-1}^{(L-1)})\|_1.
\]
Because $T^{a_{t-1}}_{\beta^*}$ is column-stochastic, it is nonexpansive in $\ell_1$:
\[
\|T^{a_{t-1}}_{\beta^*}x\|_1\leq \|x\|_1
\qquad\text{for every vector }x.
\]
Therefore
\[
\|\bar b_t-\bar b_t^{(L)}\|_1
\leq
\|b_{t-1}-b_{t-1}^{(L-1)}\|_1.
\]
Taking expectations and using the filtered-belief estimate with $L-1$ in place of $L$ yields
\[
\mathbb{E}\!\left[\|\bar b_t-\bar b_t^{(L)}\|_1\right]
\leq
\min\left(|\mathcal{S}| \exp\!\left(-\frac{\sigma_{O}^4}{C|\mathcal{S}|^{2}}L\right),\sqrt{|\mathcal{S}|\log(|\mathcal{S}|)}\;\exp\!\left(-\sigma_{O}\sqrt{\frac{L}{C|\mathcal{S}|}}\right)\right).
\]

\textbf{Case 2: $t\leq L$}
In this case, we simply define $b_{t}^{(L)}=b_{t}=b^{\mathrm{ext}}(a_{1:t-1},o_{1:t},b_{0})$, where $b^{\mathrm{ext}}$ is the exact update function for POMDP beliefs. Then
\[
\mathbb{E}\!\left[\| b_t- b_t^{(L)}\|_1\right]=0.
\]

Again, define
\[
\bar b_t^{(L)}
=
T^{a_{t-1}}_{\beta^*} b_{t-1}^{(L-1)},
\]

we then have
\[
\mathbb{E}\!\left[\| \bar{b}_t- \bar{b}_t^{(L)}\|_1\right]=0\leq \min\left(|\mathcal{S}| \exp\!\left(-\frac{\sigma_{O}^4}{C|\mathcal{S}|^{2}}L\right),\sqrt{|\mathcal{S}|\log(|\mathcal{S}|)}\;\exp\!\left(-\sigma_{O}\sqrt{\frac{L}{C|\mathcal{S}|}}\right)\right),
\]

Similar to Case 1, since $b_{t-1}^{(L-1)}$ depends only on
\[
a_{1:t-2}
\quad\text{and}\quad
o_{1:t-1},
\]
the vector $\bar b_t^{(L)}$ depends only on
\[
a_{1:t-1}
\quad\text{and}\quad
o_{1:t-1},
\]
and hence it is a measurable function of
\[
W_{1},W_{t-L+1},\dots,W_{t-1}.
\]

To obtain a function of the form
\[
g_{t}^{L} : (\Omega\times \mathcal{A})^L \to \Delta(\mathcal{S}),
\]
we simply prepend a random string $W_{t-L},\cdots,W_{0}$ to obtain the measurable map $g_{t}^{L}$.
\end{proof}

\begin{proposition}[Finite-window approximation of the matrix process]
\label{prop:matrix-truncation}
For every integer $L\geq 2$ and every $t$, define
\[
Z_t^{(L)} = \bar b_t^{(L)} \bigl(\bar b_t^{(L)}\bigr)^\top.
\]
Then
\[
\mathbb{E}\!\left[\|Z_t-Z_t^{(L)}\|_{\mathrm{op}}\right]
\leq
2\min\left(|\mathcal{S}| \exp\!\left(-\frac{\sigma_{O}^4}{C|\mathcal{S}|^{2}}L\right),\sqrt{|\mathcal{S}|\log(|\mathcal{S}|)}\;\exp\!\left(-\sigma_{O}\sqrt{\frac{L}{C|\mathcal{S}|}}\right)\right).
\]
\end{proposition}

\begin{proof}
For any two vectors $u,v\in \Delta(\mathcal{S})$,
\[
uu^\top-vv^\top
=
u(u-v)^\top + (u-v)v^\top.
\]
Hence
\[
\|uu^\top-vv^\top\|_{\mathrm{op}}
\leq
\|u\|_2 \|u-v\|_2 + \|u-v\|_2 \|v\|_2.
\]
Because $u,v\in \Delta(\mathcal{S})$, we have
\[
\|u\|_2\leq \|u\|_1=1,
\qquad
\|v\|_2\leq \|v\|_1=1,
\]
so
\[
\|uu^\top-vv^\top\|_{\mathrm{op}}
\leq
2\|u-v\|_2
\leq
2\|u-v\|_1.
\]
Apply this with $u=\bar b_t$ and $v=\bar b_t^{(L)}$, then take expectations and use Proposition~\ref{prop:predictive-truncation}.
\end{proof}

\begin{proposition}[Finite-window transforms shift the effective mixing gap]
\label{prop:lag-shift}
Let $(R_t)$ be any process and let
\[
V_t^{(L)} = h_L(R_{t-L},R_{t-L+1},\dots,R_{t-1})
\qquad\text{for every }t\geq L+1.
\]
Then for every $\ell\geq L$,
\[
\beta_{V^{(L)}}(\ell)\leq \beta_R(\ell-L+1).
\]
Informally: a length-$L$ truncated object shifts the effective mixing gap down by $L-1$.
\end{proposition}

\begin{proof}
Fix $r\geq L+1$ and $\ell\geq L$. By construction, the sigma-field generated by
\[
V_{L+1}^{(L)},V_{L+2}^{(L)},\dots,V_r^{(L)}
\]
is contained in the sigma-field generated by
\[
R_{1},\dots,R_{r-1}.
\]
Indeed, the latest variable on the past side is $V_r^{(L)}$, and that variable only depends on
\[
R_{r-L},R_{r-L+1},\dots,R_{r-1}.
\]
Therefore
\[
\sigma(V_{L+1}^{(L)},V_{L+2}^{(L)},\dots,V_r^{(L)})
\subseteq
\sigma(R_1,R_2,\dots,R_{r-1}).
\]

Similarly, the sigma-field generated by
\[
V_{r+\ell}^{(L)},V_{r+\ell+1}^{(L)},\dots
\]
is contained in the sigma-field generated by
\[
R_{r+\ell-L},R_{r+\ell-L+1},\dots.
\]
Indeed, the earliest variable on the future side is $V_{r+\ell}^{(L)}$, and that variable only depends on
\[
R_{r+\ell-L},R_{r+\ell-L+1},\dots,R_{r+\ell-1}.
\]
Therefore
\[
\sigma(V_{r+\ell}^{(L)},V_{r+\ell+1}^{(L)},\dots)
\subseteq
\sigma(R_{r+\ell-L},R_{r+\ell-L+1},\dots).
\]

Now apply Lemma~\ref{lem:beta-monotonicity}:
\begin{align*}
&\beta\!\Big(
\sigma(V_{L+1}^{(L)},\dots,V_r^{(L)}),
\sigma(V_{r+\ell}^{(L)},V_{r+\ell+1}^{(L)},\dots)
\Big)\\
\leq&
\beta\!\Big(
\sigma(R_1,\dots,R_{r-1}),
\sigma(R_{r+\ell-L},R_{r+\ell-L+1},\dots)
\Big).
\end{align*}
The lag on the right-hand side is
\[
(r+\ell-L) - (r-1) = \ell-L+1.
\]
So
\[
\beta\!\Big(
\sigma(V_{L+1}^{(L)},\dots,V_r^{(L)}),
\sigma(V_{r+\ell}^{(L)},V_{r+\ell+1}^{(L)},\dots)
\Big)
\leq
\beta_R(\ell-L+1).
\]
Taking the supremum over $r$ proves the result.
\end{proof}

\begin{lemma}[Blocking concentration for bounded geometrically mixing scalar sequences]
\label{lem:blocking}
Let $(R_t)_{t=1}^{m}$ be a scalar process with
\[
0\leq R_t\leq 1
\qquad\text{almost surely for every }t.
\]
Fix an integer $q$ with $1\leq q\leq m$. Then for every $\varepsilon>0$,
\[
\mathbb{P}\!\left(
\left|
\frac{1}{m}\sum_{t=1}^{m}\bigl(R_t-\mathbb{E}[R_t]\bigr)
\right|
\geq
\varepsilon
\right)
\leq
\left(\frac{m}{q}+1\right)\,\beta_R(q)
+
4 \exp\!\left(-\frac{m\varepsilon^2}{q}\frac{m}{m+q}\right).
\]
\end{lemma}

\begin{proof}
Let
\[
J = \left\lfloor \frac{m}{q}\right\rfloor.
\]
Partition $\{1,2,\dots,m\}$ into the $J$ full blocks
\[
I_j = \{(j-1)q+1,\dots,jq\},
\qquad
j=1,\dots,J,
\]
and one remainder block
\[
\{Jq+1,\dots,m\},
\]
whose size is strictly less than $q$.

For each full block define the block average
\[
U_j = \frac{1}{q}\sum_{t\in I_j} R_t.
\]

Similarly, for the remainder block, define the block average

\[U_{J+1}=\frac{1}{m-Jq}\sum_{t=Jq+1}^{m}R_{t}.\]

Since $0\leq R_t\leq 1$, we have
\[
0\leq U_j\leq 1.
\]

Now decompose the centered empirical average:
\[
\frac{1}{m}\sum_{t=1}^{m}\bigl(R_t-\mathbb{E}[R_t]\bigr)
=
\frac{q}{m}\sum_{j=1}^{J}\bigl(U_j-\mathbb{E}[U_j]\bigr)
+
\frac{m-Jq}{m}(U_{J+1}-\mathbb{E}[U_{J+1}]).
\]

Split the full-block sum into the odd blocks and even blocks. More specifically, if $J+1$ is odd,
\[
S_{\mathrm{odd}}
=
\frac{q}{m}\sum_{\substack{1\leq j\leq J\\ j\ \mathrm{odd}}}
\bigl(U_j-\mathbb{E}[U_j]\bigr)+\frac{m-Jq}{m}\bigl(U_{J+1}-\mathbb{E}[U_{J+1}]\bigr),
\qquad
S_{\mathrm{even}}
=
\frac{q}{m}\sum_{\substack{1\leq j\leq J\\ j\ \mathrm{even}}}
\bigl(U_j-\mathbb{E}[U_j]\bigr),
\]
and if $J+1$ is even,
\[
S_{\mathrm{odd}}
=
\frac{q}{m}\sum_{\substack{1\leq j\leq J\\ j\ \mathrm{odd}}}
\bigl(U_j-\mathbb{E}[U_j]\bigr),
\qquad
S_{\mathrm{even}}
=
\frac{q}{m}\sum_{\substack{1\leq j\leq J\\ j\ \mathrm{even}}}
\bigl(U_j-\mathbb{E}[U_j]\bigr)+\frac{m-Jq}{m}\bigl(U_{J+1}-\mathbb{E}[U_{J+1}]\bigr).
\]

Then
\[
\frac{1}{m}\sum_{t=1}^{m}\bigl(R_t-\mathbb{E}[R_t]\bigr)
=
S_{\mathrm{odd}} + S_{\mathrm{even}}.
\]
Therefore, if
\[
\left|
\frac{1}{m}\sum_{t=1}^{m}\bigl(R_t-\mathbb{E}[R_t]\bigr)
\right|
\geq
\varepsilon,
\]
then
\[
|S_{\mathrm{odd}} + S_{\mathrm{even}}|
\geq
\varepsilon,
\]
Hence at least one of the two block sums must satisfy
\[
|S_{\mathrm{odd}}|\geq \frac{1}{2}\varepsilon
\qquad\text{or}\qquad
|S_{\mathrm{even}}|\geq \frac{1}{2}\varepsilon.
\]
By the union bound,
\[
\mathbb{P}\!\left(
\left|
\frac{1}{m}\sum_{t=1}^{m}\bigl(R_t-\mathbb{E}[R_t]\bigr)
\right|
\geq
\varepsilon
\right)
\leq
\mathbb{P}\bigl(|S_{\mathrm{odd}}|\geq \frac{1}{2}\varepsilon\bigr)
+
\mathbb{P}\bigl(|S_{\mathrm{even}}|\geq \frac{1}{2}\varepsilon\bigr).
\]

We now bound the odd-block term. Let $r_{\mathrm{odd}}$ be the number of odd full blocks. Consecutive odd blocks are separated by a gap of length $q$. By Berbee's coupling lemma (first introduced in \citet{berbee1979random}, also see Lemma~5.1 of \citet{rio2017asymptotic}), there exist independent copies
\[
\widetilde U_1,\dots,\widetilde U_{r_{\mathrm{odd}}}
\]
with the same marginals as the odd block averages, such that the total coupling failure probability is at most
\[
r_{\mathrm{odd}} \beta_R(q).
\]
Hence
\[
\mathbb{P}\bigl(|S_{\mathrm{odd}}|\geq \varepsilon\bigr)
\leq
r_{\mathrm{odd}} \beta_R(q)
+
\mathbb{P}\!\left(
\left|
\frac{q}{m}\sum_{j=1}^{r_{\mathrm{odd}}}
\bigl(\widetilde U_j-\mathbb{E}[\widetilde U_j]\bigr)
\right|
\geq
\varepsilon
\right).
\]

To elaborate, the first term on the right-hand side corresponds to when there is a failure in coupling (i.e., there exists $1 \leq j \leq r_{\mathrm{odd}}$ such that $\widetilde{U}_{j}\neq U_{j}$). The second term corresponds to when the coupling is successful (i.e., $\widetilde{U}_{j}= U_{j}$, $\forall 1 \leq j \leq r_{\mathrm{odd}}$). Conditioned on successful coupling,

\[\mathbb{P}\!\left(
\left|
\frac{q}{m}\sum_{j=1}^{r_{\mathrm{odd}}}
\bigl(\widetilde U_j-\mathbb{E}[\widetilde U_j]\bigr)
\right|
\geq
\frac{1}{2}\varepsilon
\right)=\mathbb{P}\!\left(
\left|
\frac{q}{m}\sum_{j=1}^{r_{\mathrm{odd}}}
\bigl( U_j-\mathbb{E}[ U_j]\bigr)
\right|
\geq
\frac{1}{2}\varepsilon
\right)\]

(recall that $\mathbb{E}[\widetilde U_j]=\mathbb{E}[U_j]$ for all $j$ by construction).

Equivalently,
\[
\mathbb{P}\bigl(|S_{\mathrm{odd}}|\geq\frac{1}{2} \varepsilon\bigr)
\leq
r_{\mathrm{odd}} \beta_R(q)
+
\mathbb{P}\!\left(
\left|
\frac{1}{r_{\mathrm{odd}}}\sum_{j=1}^{r_{\mathrm{odd}}}
\bigl(\widetilde U_j-\mathbb{E}[\widetilde U_j]\bigr)
\right|
\geq
\frac{m\varepsilon}{2q\,r_{\mathrm{odd}}}
\right).
\]
Since the $\widetilde U_j$ are independent and lie in $[0,1]$, Hoeffding's inequality gives
\begin{align*}
\mathbb{P}\bigl(|S_{\mathrm{odd}}|\geq \frac{1}{2}\varepsilon\bigr)
\leq&
r_{\mathrm{odd}} \beta_R(q)
+
2 \exp\!\left(
-2r_{\mathrm{odd}}
\left(\frac{m\varepsilon}{2q\,r_{\mathrm{odd}}}\right)^2
\right)\\
=&r_{\mathrm{odd}} \beta_R(q)
+
2 \exp\!\left(
-\frac{1}{2}r_{\mathrm{odd}}
\left(\frac{m\varepsilon}{q\,r_{\mathrm{odd}}}\right)^2
\right).
\end{align*}
Because
\[
r_{\mathrm{odd}}=\frac{\lfloor\frac{m}{q}\rfloor+1}{2}\leq \frac{1}{2}\frac{m}{q}+\frac{1}{2},
\]
we obtain
\[
\mathbb{P}\bigl(|S_{\mathrm{odd}}|\geq \frac{1}{2}\varepsilon\bigr)
\leq
r_{\mathrm{odd}} \beta_R(q)
+
2\exp\!\left(-\frac{m\varepsilon^2}{q}\frac{m}{m+q}\right).
\]
Exactly the same argument yields
\[
\mathbb{P}\bigl(|S_{\mathrm{even}}|\geq \frac{1}{2} \varepsilon\bigr)
\leq
r_{\mathrm{even}} \beta_R(q)
+
2 \exp\!\left(-\frac{m\varepsilon^2}{q}\frac{m}{m+q}\right),
\]
where $r_{\mathrm{even}}$ is the number of even full blocks.

Since
\[
r_{\mathrm{odd}}+r_{\mathrm{even}} = J+1 \leq \frac{m}{q}+1,
\]
combining the odd and even bounds gives
\[
\mathbb{P}\!\left(
\left|
\frac{1}{m}\sum_{t=1}^{m}\bigl(R_t-\mathbb{E}[R_t]\bigr)
\right|
\geq
\varepsilon
\right)
\leq
\left(\frac{m}{q}+1\right)\,\beta_R(q)
+
4 \exp\!\left(-\frac{m\varepsilon^2}{q}\frac{m}{m+q}\right).
\]
\end{proof}

\begin{lemma}[Net argument for symmetric matrices]
\label{lem:net}
Let $\mathcal{N}$ be a $1/4$-net of the unit sphere in $\mathbb{R}^{d}$, and let $M$ be a symmetric $d\times d$ matrix. Then
\[
\|M\|_{\mathrm{op}}
\leq
2 \sup_{v\in\mathcal{N}} |v^\top M v|.
\]
\end{lemma}

\begin{proof}
Let $x$ be a unit vector such that
\[
|x^\top M x| = \|M\|_{\mathrm{op}}.
\]
Choose $v\in\mathcal{N}$ so that
\[
\|x-v\|_2\leq \frac{1}{4}.
\]
Then
\begin{align*}
|x^\top M x - v^\top M v|
&=
|(x-v)^\top M x + v^\top M (x-v)|\\
&\leq
\|x-v\|_2 \|M\|_{\mathrm{op}} \|x\|_2
+
\|v\|_2 \|M\|_{\mathrm{op}} \|x-v\|_2\\
&\leq
\frac{1}{2}\|M\|_{\mathrm{op}}.
\end{align*}
Therefore
\[
\|M\|_{\mathrm{op}}
=
|x^\top M x|
\leq
|v^\top M v| + \frac{1}{2}\|M\|_{\mathrm{op}}.
\]
Taking the supremum over $v\in\mathcal{N}$ gives
\[
\|M\|_{\mathrm{op}}
\leq
\sup_{v\in\mathcal{N}} |v^\top M v| + \frac{1}{2}\|M\|_{\mathrm{op}}.
\]
Rearranging yields the claim.
\end{proof}

\subsection{}\label{subapx: main proof of high-prob sigma B}
In this subsection, we use the facts developed in Appendix~\ref{subapx: auxiliary facts for high-prob sigma B} to convert the results in Appendix~\ref{subapx: sigma B in expectation} into a high-probability lower-bound on $\sigma_{\min}(\bar{B})$.

\textbf{Step 1: Concentrate the truncated empirical covariance around its mean.}
For each $t\geq L+1$, the truncated matrix
\[
Z_t^{(L)} = \bar b_t^{(L)} \bigl(\bar b_t^{(L)}\bigr)^\top
\]
is a measurable function of
\[
W_{t-L},W_{t-L+1},\dots,W_{t-1},
\]
where recall that $W_{t}=(o_{t},a_{t})$.

For $t\leq L$, $Z_t^{(L)}$ is a measurable function of
\[
W_{1},W_{t-L+1},\dots,W_{t-1}.
\]

Furthermore, recall Case 2 of Proposition~\ref{prop:predictive-truncation}, we can prepend a random string $W_{t-L},\cdots,W_{0}$ and $Z_{t}^{(L)}$ is still a measurable function of
\[W_{t-L},\cdots,W_{0},W_{1},\cdots,W_{t}.\]

Then, by Proposition~\ref{prop:mixing-W} and Proposition~\ref{prop:lag-shift},
\[
\beta_{Z^{(L)}}(\ell)
\leq
\beta_W(\ell-L+1)
\leq
G(\pi) \rho(\pi)^{\,\ell-L}
\qquad\text{for every }\ell\geq L.
\]

Let $m=n-t_{\mathrm{burn}}$. Then, fix a unit vector $v\in \mathbb{R}^{|\mathcal{S}|}$ and define the scalar process
\[
R_t(v)
=
v^\top Z_{t_{\mathrm{burn}}+t}^{(L)} v,
\qquad
t=1,\dots,m.
\]
Because $Z_{t_{\mathrm{burn}}+t}^{(L)}$ is positive semidefinite and $\bar b_{t_{\mathrm{burn}}+t}^{(L)}\in \Delta(\mathcal{S})$,
\[
0\leq R_t(v)\leq 1.
\]
Moreover, since $R_t(v)$ is a measurable function of $Z_{t_{\mathrm{burn}}+t}^{(L)}$, it has no larger absolute-regularity coefficient:
\[
\beta_{R(v)}(\ell)\leq \beta_{Z^{(L)}}(\ell)\leq G(\pi) \rho(\pi)^{\,\ell-L}
\qquad\text{for every }\ell\geq L.
\]

Now apply Lemma~\ref{lem:blocking} with block length $q$. Because $q\geq L$, we obtain, for every $\varepsilon>0$,
\begin{align*}
\mathbb{P}\!\left(
\left|
\frac{1}{m}\sum_{t=1}^{m}
\bigl(R_t(v)-\mathbb{E}[R_t(v)]\bigr)
\right|
\geq
\varepsilon
\right)
&\leq
\left(\frac{m}{q}+1\right)\,\beta_{R(v)}(q)
+
4 \exp\!\left(-\frac{m\varepsilon^2}{q}\frac{m}{m+q}\right)\\
&\leq
\left(\frac{m}{q}+1\right)\, G(\pi) \rho(\pi)^{\,q-L+1}
+
4 \exp\!\left(-\frac{m\varepsilon^2}{q}\frac{m}{m+q}\right).
\end{align*}
Setting
\begin{equation}\label{equ: q value}
q=L+ \left\lceil\frac{\log\left(2(m+1)G(\pi)/\delta_{1}\right)}{\log(1/\rho(\pi))}\right\rceil
\end{equation}
is sufficient for
\[
\left(\frac{m}{q}+1\right)\, G(\pi) \rho(\pi)^{\,q-L}\leq \frac{\delta_1}{2}.
\]

For readability, we retain
$q$ as a symbol and substitute \eqref{equ: q value} only at the end. The choice for $L$ will be determined in Step~3.

Therefore,
\[
\mathbb{P}\!\left(
\left|
\frac{1}{m}\sum_{t=1}^{m}
\bigl(R_t(v)-\mathbb{E}[R_t(v)]\bigr)
\right|
\geq
\varepsilon
\right)
\leq
\frac{\delta_1}{2}
+
4 \exp\!\left(-\frac{m\varepsilon^2}{q}\frac{m}{m+q}\right).
\]

Let $\mathcal{N}$ be a $1/4$-net of the unit sphere in $\mathbb{R}^{|\mathcal{S}|}$. It is a standard result (see Corollary~4.2.13 of \citet{vershynin2018high}) that
\[
|\mathcal{N}|\leq 9^{|\mathcal{S}|}.
\]
Choose
\[
\varepsilon
=
\frac{1}{m}
\sqrt{
q(m+q)\bigl(\log(8/\delta_1)+|\mathcal{S}|\log 9\bigr)}.
\]

Then
\[
4 \exp\!\left(-\frac{m\varepsilon^2}{q}\frac{m}{m+q}\right)
=
\frac{\delta_1}{2\cdot 9^{|\mathcal{S}|}}.
\]
Hence, for each fixed $v\in \mathcal{N}$,
\[
\mathbb{P}\!\left(
\left|
\frac{1}{m}\sum_{t=1}^{m}
\bigl(R_t(v)-\mathbb{E}[R_t(v)]\bigr)
\right|
\geq
\varepsilon
\right)
\leq
\frac{\delta_1}{2}
+
\frac{\delta_1}{2\cdot 9^{|\mathcal{S}|}}.
\]

Now, notice that the first term on the right-hand side corresponds to the probability of the coupling process of $R_{t}(v)$ being unsuccessful (recall the proof of Lemma~\ref{lem:blocking}). In particular, by construction, the coupling of $R_{t}(v)$ would be unsuccessful only if the coupling of $Z^{(L)}_{t}$ is unsuccessful. This means the probability of unsuccessful coupling for $Z^{(L)}_{t}$ upper-bounds the probability that there exists some $v\in \mathcal{N}$ where $R_{t}(v)$ is unsuccessfully coupled. Applying again Berbee's coupling lemma on $Z^{(L)}$ gives us that the coupling failure probability is

\[
\left(\frac{m}{q}+1\right)\, G(\pi) \rho(\pi)^{\,q-L}\leq \frac{\delta_1}{2}.
\]

On the other hand, the second term ($\frac{\delta_1}{2\cdot 9^{|\mathcal{S}|}}$) depends on the choice of $v$ and is designed to be summable over the net.

Therefore, we take a union bound of the term $\frac{\delta_1}{2\cdot 9^{|\mathcal{S}|}}$ over $\mathcal{N}$ and it yields that, with probability at least $1-\delta_1$,
\[
\sup_{v\in \mathcal{N}}
\left|
v^\top
\left(
\frac{1}{m}\sum_{t=t_{\mathrm{burn}+1}}^{n}
\bigl(Z_t^{(L)}-\mathbb{E}[Z_t^{(L)}]\bigr)
\right)
v
\right|
\leq
\varepsilon.
\]
Applying Lemma~\ref{lem:net} to the symmetric matrix
\[
\frac{1}{m}\sum_{t=t_{\mathrm{burn}}+1}^{n}
\bigl(Z_t^{(L)}-\mathbb{E}[Z_t^{(L)}]\bigr),
\]
we conclude that with probability at least $1-\delta_1$,
\[
\left\|
\frac{1}{m}\sum_{t=t_{\mathrm{burn}}+1}^{n}
\bigl(Z_t^{(L)}-\mathbb{E}[Z_t^{(L)}]\bigr)
\right\|_{\mathrm{op}}
\leq
2\varepsilon.
\]
Substituting the choice of $\varepsilon$ gives
\[
\left\|
\frac{1}{m}\sum_{t=t_{\mathrm{burn}}+1}^{n}
\bigl(Z_t^{(L)}-\mathbb{E}[Z_t^{(L)}]\bigr)
\right\|_{\mathrm{op}}
\leq
2\frac{1}{m}
\sqrt{
q(m+q)\bigl(\log(8/\delta_1)+|\mathcal{S}|\log 9\bigr)}.
\]
\textbf{Step 2: Control the truncation remainder.}
Write
\begin{align*}
\frac{1}{m}\sum_{t=t_{\mathrm{burn}}+1}^{n}\bigl(Z_t-\mathbb{E}[Z_t]\bigr)
&=
\frac{1}{m}\sum_{t=t_{\mathrm{burn}}+1}^{n}\bigl(Z_t^{(L)}-\mathbb{E}[Z_t^{(L)}]\bigr)\\
&\quad+
\frac{1}{m}\sum_{t=t_{\mathrm{burn}}+1}^{n}
\Bigl[
(Z_t-Z_t^{(L)})-\mathbb{E}[Z_t-Z_t^{(L)}]
\Bigr].
\end{align*}
Define
\[
R
=
\frac{1}{m}\sum_{t=t_{\mathrm{burn}}+1}^{n}
\Bigl[
(Z_t-Z_t^{(L)})-\mathbb{E}[Z_t-Z_t^{(L)}]
\Bigr].
\]
By the triangle inequality,
\[
\|R\|_{\mathrm{op}}
\leq
\frac{1}{m}\sum_{t=t_{\mathrm{burn}}+1}^{n}\|Z_t-Z_t^{(L)}\|_{\mathrm{op}}
+
\frac{1}{m}\sum_{t=t_{\mathrm{burn}}+1}^{n}\mathbb{E}\!\left[\|Z_t-Z_t^{(L)}\|_{\mathrm{op}}\right].
\]
Proposition~\ref{prop:matrix-truncation} gives
\begin{align*}
\mathbb{E}\!\left[\|Z_t-Z_t^{(L)}\|_{\mathrm{op}}\right]
\leq
2\min\left(|\mathcal{S}| \exp\!\left(-\frac{\sigma_{O}^4}{C|\mathcal{S}|^{2}}L\right),\sqrt{|\mathcal{S}|\log(|\mathcal{S}|)}\;\exp\!\left(-\sigma_{O}\sqrt{\frac{L}{C|\mathcal{S}|}}\right)\right)
\end{align*}
for every $t\geq t_{\mathrm{burn}}+1$.
Hence
\begin{align*}
&\frac{1}{m}\sum_{t=t_{\mathrm{burn}}+1}^{n}\mathbb{E}\!\left[\|Z_t-Z_t^{(L)}\|_{\mathrm{op}}\right]\\
\leq
&2 \min\left(|\mathcal{S}| \exp\!\left(-\frac{\sigma_{O}^4}{C|\mathcal{S}|^{2}}L\right),\sqrt{|\mathcal{S}|\log(|\mathcal{S}|)}\;\exp\!\left(-\sigma_{O}\sqrt{\frac{L}{C|\mathcal{S}|}}\right)\right).
\end{align*}
The same bound holds for the expectation of the random average:
\begin{align*}
&\mathbb{E}\!\left[
\frac{1}{m}\sum_{t=t_{\mathrm{burn}}+1}^{n}\|Z_t-Z_t^{(L)}\|_{\mathrm{op}}
\right]\\
\leq
&2 \min\left(|\mathcal{S}| \exp\!\left(-\frac{\sigma_{O}^4}{C|\mathcal{S}|^{2}}L\right),\sqrt{|\mathcal{S}|\log(|\mathcal{S}|)}\;\exp\!\left(-\sigma_{O}\sqrt{\frac{L}{C|\mathcal{S}|}}\right)\right).
\end{align*}
Therefore, by Markov's inequality, with probability at least $1-\delta_2$,
\[
\frac{1}{m}\sum_{t=t_{\mathrm{burn}}+1}^{n}\|Z_t-Z_t^{(L)}\|_{\mathrm{op}}
\leq
\frac{
2 \min\left(|\mathcal{S}| \exp\!\left(-\frac{\sigma_{O}^4}{C|\mathcal{S}|^{2}}L\right),\sqrt{|\mathcal{S}|\log(|\mathcal{S}|)}\;\exp\!\left(-\sigma_{O}\sqrt{\frac{L}{C|\mathcal{S}|}}\right)\right)
}{
\delta_2
}.
\]
On the same event,
\[
\|R\|_{\mathrm{op}}
\leq
2 \min\left(|\mathcal{S}| \exp\!\left(-\frac{\sigma_{O}^4}{C|\mathcal{S}|^{2}}L\right),\sqrt{|\mathcal{S}|\log(|\mathcal{S}|)}\;\exp\!\left(-\sigma_{O}\sqrt{\frac{L}{C|\mathcal{S}|}}\right)\right)
\left(1+\frac{1}{\delta_2}\right).
\]

\textbf{Step 3: Combine the mean lower bound, the truncated concentration bound, and the truncation remainder bound.}
We have the exact identity
\[
\frac{1}{m}\sum_{t=t_{\mathrm{burn}}+1}^{n} Z_t
=
\Sigma
+
\frac{1}{m}\sum_{t=t_{\mathrm{burn}}+1}^{n}\bigl(Z_t^{(L)}-\mathbb{E}[Z_t^{(L)}]\bigr)
+
R,
\]
where
\[\Sigma=\frac{1}{n-t_{\mathrm{burn}}}
\sum_{t=t_{\mathrm{burn}}+1}^{n}\mathbb E[Z_{t}]=\frac{1}{n-t_{\mathrm{burn}}}
\sum_{t=t_{\mathrm{burn}}+1}^{n}\mathbb E[\bar b_t\bar b_t^\top].\]

Hence, by Weyl's inequality,
\begin{align*}
\lambda_{\min}\!\left(
\frac{1}{m}\sum_{t=t_{\mathrm{burn}}+1}^{n} Z_t
\right)
&\geq
\lambda_{\min}(\Sigma)
-
\left\|
\frac{1}{m}\sum_{t=t_{\mathrm{burn}}+1}^{n}\bigl(Z_t^{(L)}-\mathbb{E}[Z_t^{(L)}]\bigr)
\right\|_{\mathrm{op}}
-
\|R\|_{\mathrm{op}}.
\end{align*}

From Corollary~\ref{cor: predictive belief isotropy after burn-in},
\[
\frac{1}{m}
\sum_{t=t_{\mathrm{burn}}+1}^{n}\mathbb E[Z_{t}]
\succeq
|\mathcal A|\,\pi_{\min}\,\sigma_{T}^2\,\sigma_{O}^2
\left(\frac{7d_{\min}^{\pi}}{8}\right)^2I,
\]

By Step~1, with probability at least $1-\delta_1$,
\[
\left\|
\frac{1}{m}\sum_{t=t_{\mathrm{burn}}+1}^{n}\bigl(Z_t^{(L)}-\mathbb{E}[Z_t^{(L)}]\bigr)
\right\|_{\mathrm{op}}
\leq
2\frac{1}{m}
\sqrt{
q(m+q)\bigl(\log(8/\delta_1)+|\mathcal{S}|\log 9\bigr)}.
\]
By Step~2, with probability at least $1-\delta_2$,
\[
\|R\|_{\mathrm{op}}
\leq
2 \min\left(|\mathcal{S}| \exp\!\left(-\frac{\sigma_{O}^4}{C|\mathcal{S}|^{2}}L\right),\sqrt{|\mathcal{S}|\log(|\mathcal{S}|)}\;\exp\!\left(-\sigma_{O}\sqrt{\frac{L}{C|\mathcal{S}|}}\right)\right)
\left(1+\frac{1}{\delta_2}\right).
\]
By another union bound, with probability at least $1-\delta_1-\delta_2$ all three estimates hold simultaneously. Therefore
\begin{align*}
&\lambda_{\min}\!\left(
\frac{1}{m}\sum_{t=t_{\mathrm{burn}}+1}^{n} Z_t
\right)\\
\geq&
|\mathcal A|\,\pi_{\min}\,\sigma_{T}^2\,\sigma_{O}^2
\left(\frac{7d_{\min}^{\pi}}{8}\right)^2\\
-&2\frac{1}{m}
\sqrt{
q(m+q)\bigl(\log(8/\delta_1)+|\mathcal{S}|\log 9\bigr)}\\
-&
2 \min\left(|\mathcal{S}| \exp\!\left(-\frac{\sigma_{O}^4}{C|\mathcal{S}|^{2}}L\right),\sqrt{|\mathcal{S}|\log(|\mathcal{S}|)}\;\exp\!\left(-\sigma_{O}\sqrt{\frac{L}{C|\mathcal{S}|}}\right)\right)
\left(1+\frac{1}{\delta_2}\right).
\end{align*}

The choice
\begin{equation}\label{equ: L value}
\begin{aligned}
L= \min\Bigg(\frac{C|\mathcal{S}|^2}{\sigma_O^4} \, \log\!\left( \frac{1024\,(1 + 1/\delta_2)\,|\mathcal{S}|}{49|\mathcal{A}|\,\pi_{\min}\,\sigma_T^2\,\sigma_O^2\,(d_{\min}^\pi)^2} \right),\\
\frac{C|\mathcal{S}|}{\sigma_O^2} \, \log^2\!\left( \frac{1024\,(1 + 1/\delta_2)\sqrt{|\mathcal{S}|\log|\mathcal{S}|}}{49|\mathcal{A}|\,\pi_{\min}\,\sigma_T^2\,\sigma_O^2\,(d_{\min}^\pi)^2} \right)\Bigg),
\end{aligned}
\end{equation}
gives
\begin{align*}
&2 \min\left(|\mathcal{S}| \exp\!\left(-\frac{\sigma_{O}^4}{C|\mathcal{S}|^{2}}L\right),\sqrt{|\mathcal{S}|\log(|\mathcal{S}|)}\;\exp\!\left(-\sigma_{O}\sqrt{\frac{L}{C|\mathcal{S}|}}\right)\right)
\left(1+\frac{1}{\delta_2}\right)\\
\leq &\frac{1}{8}|\mathcal A|\,\pi_{\min}\,\sigma_{T}^2\,\sigma_{O}^2
\left(\frac{7d_{\min}^{\pi}}{8}\right)^2.
\end{align*}

This leads to a simpler bound:
\begin{align*}
\lambda_{\min}\!\left(
\frac{1}{m}\sum_{t=t_{\mathrm{burn}}+1}^{n} Z_t
\right)
&\geq
\frac{7}{8}|\mathcal A|\,\pi_{\min}\,\sigma_{T}^2\,\sigma_{O}^2
\left(\frac{7d_{\min}^{\pi}}{8}\right)^2\\
&\quad-
2\frac{1}{m}
\sqrt{
q(m+q)\bigl(\log(8/\delta_1)+|\mathcal{S}|\log 9\bigr)}.
\end{align*}

A simple calculation shows that it is sufficient to have

\[m\geq \frac{14\, \bigl(\log(8/\delta_1) + |\mathcal{S}|\log 9\bigr)}{|\mathcal{A}|^2\, \pi_{\min}^2\, \sigma_T^4\, \sigma_O^4\, (d_{\min}^\pi)^4}q\]

for

\[2\frac{1}{m}
\sqrt{
q(m+q)\bigl(\log(8/\delta_1)+|\mathcal{S}|\log 9\bigr)}\leq \frac{7}{8}\cdot\frac{7}{8}|\mathcal A|\,\pi_{\min}\,\sigma_{T}^2\,\sigma_{O}^2
\left(\frac{7d_{\min}^{\pi}}{8}\right)^2,\]

which means

\begin{align*}
\lambda_{\min}\!\left(
\frac{1}{m}\sum_{t=t_{\mathrm{burn}}+1}^{n} Z_t
\right)
&\geq
\frac{7}{8}|\mathcal A|\,\pi_{\min}\,\sigma_{T}^2\,\sigma_{O}^2
\left(\frac{7d_{\min}^{\pi}}{8}\right)^2-
\frac{7}{8}\cdot\frac{7}{8}|\mathcal A|\,\pi_{\min}\,\sigma_{T}^2\,\sigma_{O}^2
\left(\frac{7d_{\min}^{\pi}}{8}\right)^2\\
&\geq \frac{1}{12}|\mathcal A|\,\pi_{\min}\,\sigma_{T}^2\,\sigma_{O}^2
\left(d_{\min}^{\pi}\right)^2.
\end{align*}

Furthermore, we plug in the value of $q$ from Equation~\eqref{equ: q value} and obtain

\begin{align*}
q=&L+\left\lceil\frac{\log\left(2(m+1)G(\pi)/\delta_{1}\right)}{\log(1/\rho(\pi))}\right\rceil\\
=&\min\Bigg(\frac{C|\mathcal{S}|^2}{\sigma_O^4} \, \log\!\left( \frac{1024\,(1 + 1/\delta_2)\,|\mathcal{S}|}{49|\mathcal{A}|\,\pi_{\min}\,\sigma_T^2\,\sigma_O^2\,(d_{\min}^\pi)^2} \right),\\&\frac{C|\mathcal{S}|}{\sigma_O^2} \, \log^2\!\left( \frac{1024\,(1 + 1/\delta_2)\sqrt{|\mathcal{S}|\log|\mathcal{S}|}}{49|\mathcal{A}|\,\pi_{\min}\,\sigma_T^2\,\sigma_O^2\,(d_{\min}^\pi)^2} \right)\Bigg)+\left\lceil\frac{\log\left(2(m+1)G(\pi)/\delta_{1}\right)}{\log(1/\rho(\pi))}\right\rceil.
\end{align*}

Recall that $m=n-t_{\mathrm{burn}}=n-
\left\lceil
\frac{\log\!\bigl(8G(\pi)/d_{\min}^{\pi}\bigr)}
{\log\!\bigl(1/\rho(\pi)\bigr)}
\right\rceil-1$.

Now
\[
\sum_{t=t_{\mathrm{burn}}+1}^{n} Z_t
=
\begin{bmatrix}
\bar b_{t_{\mathrm{burn}}+1}^\top\\
\bar b_{t_{\mathrm{burn}}+2}^\top\\
\vdots\\
\bar b_n^\top
\end{bmatrix}^{\!\top}
\begin{bmatrix}
\bar b_{t_{\mathrm{burn}}+1}^\top\\
\bar b_{t_{\mathrm{burn}}+2}^\top\\
\vdots\\
\bar b_n^\top
\end{bmatrix},
\]
so
\begin{equation}\label{equ: tilde B sigma min squared bound}
\sigma_{\min}\!\left(
\begin{bmatrix}
\bar b_{t_{\mathrm{burn}}+1}^\top\\
\bar b_{t_{\mathrm{burn}}+2}^\top\\
\vdots\\
\bar b_n^\top
\end{bmatrix}
\right)^2
=
m\,
\lambda_{\min}\!\left(
\frac{1}{m}\sum_{t=t_{\mathrm{burn}}+1}^{n} Z_t
\right)\geq \frac{1}{12}m|\mathcal A|\,\pi_{\min}\,\sigma_{T}^2\,\sigma_{O}^2
\left(d_{\min}^{\pi}\right)^2.
\end{equation}

Finally, since
\[
\bar B_n^\top \bar B_n
=
\sum_{t=1}^{n} Z_t
\succeq
\sum_{t=t_{\mathrm{burn}}+1}^{n} Z_t,
\]
we have
\[
\sigma_{\min}(\bar B_n)^2
=
\lambda_{\min}(\bar B_n^\top \bar B_n)
\geq
\lambda_{\min}\!\left(\sum_{t=t_{\mathrm{burn}}+1}^{n} Z_t\right).
\]
Thus the same lower bound also holds for $\sigma_{\min}(\bar B_n)$.

In addition, we derive a sufficient condition on the lower bound of $n$ for a more interpretable result.

For ease of notation, let
\[
\gamma_{\pi}:=\log\frac{1}{\rho(\pi)}>0
\]
and
\[D\coloneqq\frac{14\, \bigl(\log(8/\delta_1) + |\mathcal{S}|\log 9\bigr)}{|\mathcal{A}|^2\, \pi_{\min}^2\, \sigma_T^4\, \sigma_O^4\, (d_{\min}^\pi)^4}.\]

Furthermore, define \(\log_{+}(x):=\max\{\log x,0\}.\)

Then, Lemma~\ref{lem: A non-circular sufficient condition}  gives us that it is sufficient to have
\[
n
\ge
t_{\mathrm{burn}}
+
2DL
+
\frac{2D}{\gamma_{\pi}}
\left[
\log\frac{4G(\pi)}{\delta_1}
+
\log_{+}\left(\frac{2D}{\gamma_{\pi}}\right)
\right],
\]
where $D$ and $\gamma_{\pi}$ only depends on the POMDP parameters and the behavior policy. This concludes the proof of Proposition~\ref{prop: predictive belief full-rankedness_formal}.

\begin{lemma}\label{lem: A non-circular sufficient condition}
If
\[
n
\ge
t_{\mathrm{burn}}
+
2DL
+
\frac{2D}{\gamma_{\pi}}
\left[
\log\frac{4G(\pi)}{\delta_1}
+
\log_{+}\left(\frac{2D}{\gamma_{\pi}}\right)
\right],
\]
then
\[
n\ge t_{\mathrm{burn}}+Dq.
\]
\end{lemma}

\begin{proof}
Recall that
\[
q
=
L-1+
\left\lceil
\frac{1}{\gamma_{\pi}}
\log\frac{2(n-t_{\mathrm{burn}}+1)G(\pi)}{\delta_1}
\right\rceil .
\]

Since \(\lceil x\rceil\le x+1\), we have
\[
q
\le
L
+
\frac{1}{\gamma_{\pi}}
\log\frac{2(n-t_{\mathrm{burn}}+1)G(\pi)}{\delta_1}.
\]
Therefore,
\[
Dq
\le
DL
+
\frac{D}{\gamma_{\pi}}
\log\frac{2(n-t_{\mathrm{burn}}+1)G(\pi)}{\delta_1}.
\]
It is enough to show that the right-hand side is at most \(n-t_{\mathrm{burn}}\).

We first prove
\[
\log\frac{n-t_{\mathrm{burn}}+1}{2}
\le
\frac{\gamma_{\pi}(n-t_{\mathrm{burn}})}{2D}
+
\log_{+}\left(\frac{2D}{\gamma_{\pi}}\right).
\]
If \(\frac{2D}{\gamma_{\pi}}<1\), then \(\log_{+}\left(\frac{2D}{\gamma_{\pi}}\right)=0\), and
\[
\log\frac{n-t_{\mathrm{burn}}+1}{2}
\le
\log(n-t_{\mathrm{burn}}+1)
\le
n-t_{\mathrm{burn}}
\le
\frac{\gamma_{\pi}(n-t_{\mathrm{burn}})}{2D},
\]
where the last inequality uses \(\frac{2D}{\gamma_{\pi}}<1\).

If \(\frac{2D}{\gamma_{\pi}}\ge 1\), then
\[
\begin{aligned}
\log\frac{n-t_{\mathrm{burn}}+1}{2}
&=
\log\left(\frac{2D}{\gamma_{\pi}}\right)
+
\log\left(\frac{\gamma_{\pi}(n-t_{\mathrm{burn}}+1)}{4D}\right) \\
&\le
\log\left(\frac{2D}{\gamma_{\pi}}\right)
+
\log\left(1+\frac{\gamma_{\pi}(n-t_{\mathrm{burn}})}{2D}\right) \\
&\le
\log\left(\frac{2D}{\gamma_{\pi}}\right)
+
\frac{\gamma_{\pi}(n-t_{\mathrm{burn}})}{2D} \\
&=
\log_{+}\left(\frac{2D}{\gamma_{\pi}}\right)
+
\frac{\gamma_{\pi}(n-t_{\mathrm{burn}})}{2D}.
\end{aligned}
\]
Thus, in all cases,
\[
\log\frac{n-t_{\mathrm{burn}}+1}{2}
\le
\frac{\gamma_{\pi}(n-t_{\mathrm{burn}})}{2D}
+
\log_{+}\left(\frac{2D}{\gamma_{\pi}}\right).
\]

Using this bound,
\[
\begin{aligned}
\log\frac{2(n-t_{\mathrm{burn}}+1)G(\pi)}{\delta_1}
&=
\log\frac{4G(\pi)}{\delta_1}
+
\log\frac{n-t_{\mathrm{burn}}+1}{2} \\
&\le
\log\frac{4G(\pi)}{\delta_1}
+
\log_{+}\left(\frac{2D}{\gamma_{\pi}}\right)
+
\frac{\gamma_{\pi}(n-t_{\mathrm{burn}})}{2D}.
\end{aligned}
\]
Multiplying by \(D/\gamma_{\pi}\), we get
\[
\frac{D}{\gamma_{\pi}}
\log\frac{2(n-t_{\mathrm{burn}}+1)G(\pi)}{\delta_1}
\le
\frac{D}{\gamma_{\pi}}
\left[
\log\frac{4G(\pi)}{\delta_1}
+
\log_{+}\left(\frac{2D}{\gamma_{\pi}}\right)
\right]
+
\frac{n-t_{\mathrm{burn}}}{2}.
\]
Therefore,
\[
\begin{aligned}
Dq
&\le
DL
+
\frac{D}{\gamma_{\pi}}
\left[
\log\frac{4G(\pi)}{\delta_1}
+
\log_{+}\left(\frac{2D}{\gamma_{\pi}}\right)
\right]
+
\frac{n-t_{\mathrm{burn}}}{2}.
\end{aligned}
\]
Then, assuming
\[
\frac{n-t_{\mathrm{burn}}}{2}
\ge
DL
+
\frac{D}{\gamma_{\pi}}
\left[
\log\frac{4G(\pi)}{\delta_1}
+
\log_{+}\left(\frac{2D}{\gamma_{\pi}}\right)
\right]
\]
gives us
\[
Dq\le DL
+
\frac{D}{\gamma_{\pi}}
\left[
\log\frac{4G(\pi)}{\delta_1}
+
\log_{+}\left(\frac{2D}{\gamma_{\pi}}\right)
\right]
+
\frac{n-t_{\mathrm{burn}}}{2}\le \frac{n-t_{\mathrm{burn}}}{2}+\frac{n-t_{\mathrm{burn}}}{2}= n-t_{\mathrm{burn}}.
\]
\end{proof}

\section{}\label{apx: proof for lem sigma B_a}
We first give the formal Lemma statement.

\begin{lemma}[Formal]\label{lem: action-restricted predictive belief full-rankedness_formal}
Under Assumption~\ref{ass:alpha_undercomplete}, \ref{ass: full-rank transition kernels}, \ref{ass: minimum policy mass}, and \ref{ass: ergodicity}, let
\[t_{\mathrm{burn}}=1+
\left\lceil
\frac{\log\!\bigl(8G(\pi)/d_{\min}^{\pi}\bigr)}
{\log\!\bigl(1/\rho(\pi)\bigr)}
\right\rceil,\]
\begin{align*}
L=&\min\Bigg(\frac{C|\mathcal{S}|^2}{\sigma_O^4} \, \log\!\left( \frac{1024\,(1 + 1/\delta_2)\,|\mathcal{S}|}{49|\mathcal{A}|\,\pi_{\min}\,\sigma_T^2\,\sigma_O^2\,(d_{\min}^\pi)^2} \right),\frac{C|\mathcal{S}|}{\sigma_O^2} \, \log^2\!\left( \frac{1024\,(1 + 1/\delta_2)\sqrt{|\mathcal{S}|\log|\mathcal{S}|}}{49|\mathcal{A}|\,\pi_{\min}\,\sigma_T^2\,\sigma_O^2\,(d_{\min}^\pi)^2} \right)\Bigg),
\end{align*}
and
\[D=\frac{14\, \bigl(\log(8/\delta_1) + |\mathcal{S}|\log 9\bigr)}{|\mathcal{A}|^2\, \pi_{\min}^2\, \sigma_T^4\, \sigma_O^4\, (d_{\min}^\pi)^4}.\]

If
\begin{equation*}
n\geq 2DL+\frac{2D}{\log(\frac{1}{\rho(\pi)})}\left[\log\left(\frac{4G(\pi)}{\delta_{1}}\right)+\max\left\{\log\left(\frac{2D}{\log(\frac{1}{\rho(\pi)})}\right),0\right\}\right]+t_{\mathrm{burn}},
\end{equation*}
then, with probability at least $1-\delta_1-\delta_2-\delta_{3}$,
\begin{align*}
&\sigma_{\min}\!\left(\bar{B}_{a}\right)^2 
\geq \frac{1}{12}(n-t_{\mathrm{burn}})|\mathcal A|\,\pi_{\min}^{2}\,\sigma_{T}^2\,\sigma_{O}^2
\left(d_{\min}^{\pi}\right)^2-
\sqrt{2(n-t_{\mathrm{burn}})\log\frac{2|\mathcal S||\mathcal A|}{\delta_{3}}}-
\frac{2}{3}\log\frac{2|\mathcal S||\mathcal A|}{\delta_{3}}
\end{align*}
holds simultaneously for all \(a\in\mathcal A\).
\end{lemma}

Taking $\delta_{1}=\delta_{2}=\frac{1}{2}\delta$ gives the more compact representation.

It is easy to see that Lemma~\ref{lem: action-restricted predictive belief full-rankedness_formal}satisfies the assumptions of Proposition~\ref{prop: predictive belief full-rankedness_formal}. Then, from Proposition~\ref{prop: predictive belief full-rankedness_formal}, we have that with probability at least $1-\delta_{1}-\delta_{2}$,
\begin{equation*}
    \sigma_{\min}(\bar{B})^2
\;\ge\;
\frac{1}{12} (n-t_{\mathrm{burn}})|\mathcal A|\pi_{\min}
\sigma_T^{2}\sigma_O^{2}(d_{\min}^{\pi})^{2}.
\end{equation*}

In this appendix, we inherent the convention from Appendix~\ref{apx: proof for predictive belief full-rankedness} where we omit the $\theta^*$ and denote the true belief vectors as $b_{t}$ and true predictive belief vectors as $\bar{b}_{t}$. In addition, we use $\|\cdot\|_{op}$ to denote matrix spectral norms.

Fix \(a\in\mathcal A\). Let
\[
\mathcal F_t:=\sigma(o_1,a_1,\ldots,o_t,a_t).
\]
Since \(\bar b_t\) is the predictive belief at time \(t\), it is \(\mathcal F_{t-1}\)-measurable. Define
\[
p_t(a):=\mathbb P(a_t=a\mid \mathcal F_{t-1}).
\]

Furthermore, since $\bar{b}_{t}(s)=\mathbb{P}(s_{t}=s|\mathcal{F}_{t-1})$, it is easy to see that
\[p_{t}(a)=\sum_{s}\bar{b}_{t}(s)\sum_{o}O_{\varphi^*}(o|s)\pi(a|o)\geq \pi_{\min}\sum_{s}\bar{b}_{t}(s)=\pi_{\min},\]
where the inequality follows from Assumption~\ref{ass: minimum policy mass}.

Define $\widetilde{B}$ as the submatrix of $\bar{B}$ whose rows only contain those of $\bar{b}_{t}$ with $t\geq t_{\mathrm{burn}}+1$, and $\widetilde{B}_{a}$ as the submatrix of $\bar{B}_{a}$ whose rows only contain those of $\bar{b}_{t}$ with $t\geq t_{\mathrm{burn}}+1$. Then,
\[
\widetilde {B}_a^\top \widetilde{B}_a
=
\sum_{t=t_{\mathrm{burn}}+1}^{n}\mathbf 1\{a_t=a\}\bar b_t\bar b_t^\top.
\]
Therefore
\[
\widetilde {B}_a^\top \widetilde{B}_a
=
\sum_{t=t_{\mathrm{burn}}+1}^{n}p_t(a)\bar b_t\bar b_t^\top
+
\sum_{t=t_{\mathrm{burn}}+1}^{n}\bigl(\mathbf 1\{a_t=a\}-p_t(a)\bigr)\bar b_t\bar b_t^\top.
\]
Since \(p_t(a)\ge \pi_{\min}\), this implies
\begin{align*}
&\widetilde {B}_a^\top \widetilde{B}_a
\succeq
\pi_{\min}\sum_{t=t_{\mathrm{burn}}+1}^{n}\bar b_t\bar b_t^\top
+
\sum_{t=t_{\mathrm{burn}}+1}^{n}\bigl(\mathbf 1\{a_t=a\}-p_t(a)\bigr)\bar b_t\bar b_t^\top\\
=&
\pi_{\min}\widetilde {B}^\top \widetilde{B}
+
\sum_{t=t_{\mathrm{burn}}+1}^{n}\bigl(\mathbf 1\{a_t=a\}-p_t(a)\bigr)\bar b_t\bar b_t^\top.
\end{align*}

For symmetric matrices, Weyl's inequality gives
\begin{equation}\label{equ: weyl for lambda min}
\lambda_{\min}(\widetilde {B}_a^\top \widetilde{B}_a)
\ge
\pi_{\min}\lambda_{\min}(\widetilde {B}^\top \widetilde{B})
+
\lambda_{\min}\left(
\sum_{t=t_{\mathrm{burn}}+1}^{n}\bigl(\mathbf 1\{a_t=a\}-p_t(a)\bigr)\bar b_t\bar b_t^\top
\right).
\end{equation}

Since
\[\sum_{t=t_{\mathrm{burn}}+1}^{n}\bigl(\mathbf 1\{a_t=a\}-p_t(a)\bigr)\bar b_t\bar b_t^\top\]
is symmetric,
\begin{align*}
&\left\|\sum_{t=t_{\mathrm{burn}}+1}^{n}\bigl(\mathbf 1\{a_t=a\}-p_t(a)\bigr)\bar b_t\bar b_t^\top\right\|_{\mathrm{op}}\\
=&\max\Bigg(\left|\lambda_{\min}\left(\sum_{t=t_{\mathrm{burn}}+1}^{n}\bigl(\mathbf 1\{a_t=a\}-p_t(a)\bigr)\bar b_t\bar b_t^\top\right)\right|,\\&\qquad\left|\lambda_{\max}\left(\sum_{t=t_{\mathrm{burn}}+1}^{n}\bigl(\mathbf 1\{a_t=a\}-p_t(a)\bigr)\bar b_t\bar b_t^\top\right)\right|\Bigg).
\end{align*}
Therefore,
\begin{equation}\label{equ: operator norm and lambda min}
\lambda_{\min}\left(\sum_{t=t_{\mathrm{burn}}+1}^{n}\bigl(\mathbf 1\{a_t=a\}-p_t(a)\bigr)\bar b_t\bar b_t^\top\right)\geq-\left\|\sum_{t=t_{\mathrm{burn}}+1}^{n}\bigl(\mathbf 1\{a_t=a\}-p_t(a)\bigr)\bar b_t\bar b_t^\top\right\|_{\mathrm{op}}.\end{equation}

Plugging \eqref{equ: operator norm and lambda min} into \eqref{equ: weyl for lambda min} gives:

\begin{equation}\label{equ: pre-martingale sigma relation}
\lambda_{\min}(\widetilde {B}_a^\top \widetilde{B}_a)
\ge
\pi_{\min}\lambda_{\min}(\widetilde {B}^\top \widetilde{B})
-
\left\|
\sum_{t=t_{\mathrm{burn}}+1}^{n}\bigl(\mathbf 1\{a_t=a\}-p_t(a)\bigr)\bar b_t\bar b_t^\top
\right\|_{\mathrm{op}}.
\end{equation}

Now define the self-adjoint matrix increments
\[
X_t(a):=\bigl(\mathbf 1\{a_t=a\}-p_t(a)\bigr)\bar b_t\bar b_t^\top,
\qquad
t=t_{\mathrm{burn}}+1,\ldots,n.
\]
Since \(\bar b_t\) is \(\mathcal F_{t-1}\)-measurable and
\[
\mathbb E\!\left[\mathbf 1\{a_t=a\}-p_t(a)\mid \mathcal F_{t-1}\right]=0,
\]
we have
\[
\mathbb E\!\left[X_t(a)\mid \mathcal F_{t-1}\right]=0.
\]
Hence the partial sums of \(X_t(a)\) form a self-adjoint matrix martingale.

Because \(\bar b_t\) is a probability vector,
\[
\|X_t(a)\|_{\mathrm{op}}
\le
\|\bar b_t\bar b_t^\top\|_{\mathrm{op}}
=
\|\bar b_t\|_2^2
\le
\|\bar b_t\|_1^2
=
1.
\]
Also,
\[
X_t(a)^2
=
\bigl(\mathbf 1\{a_t=a\}-p_t(a)\bigr)^2(\bar b_t\bar b_t^\top)^2.
\]
Taking conditional expectations,
\[
\mathbb E\!\left[X_t(a)^2\mid \mathcal F_{t-1}\right]
=
p_t(a)\bigl(1-p_t(a)\bigr)(\bar b_t\bar b_t^\top)^2
\preceq
(\bar b_t\bar b_t^\top)^2.
\]
Since
\[
(\bar b_t\bar b_t^\top)^2
=
\|\bar b_t\|_2^2\,\bar b_t\bar b_t^\top
\preceq
\bar b_t\bar b_t^\top,
\]
it follows that
\[
\mathbb E\!\left[X_t(a)^2\mid \mathcal F_{t-1}\right]
\preceq
\bar b_t\bar b_t^\top.
\]
Therefore
\[
\left\|
\sum_{t=t_{\mathrm{burn}}+1}^{n}\mathbb E\!\left[X_t(a)^2\mid \mathcal F_{t-1}\right]
\right\|_{\mathrm{op}}
\le
\left\|
\sum_{t=t_{\mathrm{burn}}+1}^{n}\bar b_t\bar b_t^\top
\right\|_{\mathrm{op}}
=
\|\widetilde {B}^\top \widetilde{B}\|_{\mathrm{op}}.
\]
By triangle inequality of the operator norm,
\[
\|\widetilde {B}^\top \widetilde{B}\|_{\mathrm{op}}
\le
\sum_{t=t_{\mathrm{burn}}+1}^{n}\|\bar b_t\bar b_t^\top\|_{\mathrm{op}}
\le
n-t_{\mathrm{burn}}.
\]

Now apply matrix Freedman's inequality to the self-adjoint matrix martingale
\[
\sum_{t=t_{\mathrm{burn}}+1}^{n} X_t(a).
\]
Since the matrix dimension is \(|\mathcal S|\), the increment bound is \(1\), and the predictable quadratic variation bound is \(n-t_{\mathrm{burn}}\), matrix Freedman's inequality implies that for every \(x>0\),
\[
\mathbb P\!\left(
\left\|
\sum_{t=t_{\mathrm{burn}}+1}^{n} X_t(a)
\right\|_{\mathrm{op}}
\ge x
\right)
\le
2|\mathcal S|
\exp\!\left(
-\frac{x^2}{
2\left((n-t_{\mathrm{burn}})+x/3\right)
}
\right).
\]

We now choose \(x\) so that the right-hand side is at most \(\delta/|\mathcal A|\). It is enough to require
\[
2|\mathcal S|
\exp\!\left(
-\frac{x^2}{
2\left((n-t_{\mathrm{burn}})+x/3\right)
}
\right)
\le
\frac{\delta}{|\mathcal A|}.
\]
Taking logarithms, this is equivalent to
\[
\frac{x^2}{
2\left((n-t_{\mathrm{burn}})+x/3\right)
}
\ge
\log\frac{2|\mathcal S||\mathcal A|}{\delta}.
\]
Multiplying both sides by
\[
2\left((n-t_{\mathrm{burn}})+x/3\right)
\]
gives
\[
x^2
\ge
2(n-t_{\mathrm{burn}})
\log\frac{2|\mathcal S||\mathcal A|}{\delta}
+
\frac{2x}{3}
\log\frac{2|\mathcal S||\mathcal A|}{\delta}.
\]
Equivalently,
\[
x^2
-
\frac{2x}{3}
\log\frac{2|\mathcal S||\mathcal A|}{\delta}
-
2(n-t_{\mathrm{burn}})
\log\frac{2|\mathcal S||\mathcal A|}{\delta}
\ge
0.
\]
The corresponding quadratic equation is
\[
x^2
-
\frac{2x}{3}
\log\frac{2|\mathcal S||\mathcal A|}{\delta}
-
2(n-t_{\mathrm{burn}})
\log\frac{2|\mathcal S||\mathcal A|}{\delta}
=
0.
\]
Its positive root is
\[
x
=
\frac{1}{3}\log\frac{2|\mathcal S||\mathcal A|}{\delta}
+
\sqrt{
2(n-t_{\mathrm{burn}})
\log\frac{2|\mathcal S||\mathcal A|}{\delta}
+
\frac{1}{9}
\left(
\log\frac{2|\mathcal S||\mathcal A|}{\delta}
\right)^2
}.
\]
Therefore, with probability at least \(1-\delta/|\mathcal A|\),
\[
\left\|
\sum_{t=t_{\mathrm{burn}}+1}^{n} X_t(a)
\right\|_{\mathrm{op}}
\le
\frac{1}{3}\log\frac{2|\mathcal S||\mathcal A|}{\delta}
+
\sqrt{
2(n-t_{\mathrm{burn}})
\log\frac{2|\mathcal S||\mathcal A|}{\delta}
+
\frac{1}{9}
\left(
\log\frac{2|\mathcal S||\mathcal A|}{\delta}
\right)^2
}.
\]

Finally, using
\[
\sqrt{u+v} \le \sqrt{u}+\sqrt{v}
\]
with
\[
u
=
2(n-t_{\mathrm{burn}})
\log\frac{2|\mathcal S||\mathcal A|}{\delta},
\qquad
v
=
\frac{1}{9}
\left(
\log\frac{2|\mathcal S||\mathcal A|}{\delta}
\right)^2,
\]
we obtain the simpler bound
\[
\left\|
\sum_{t=t_{\mathrm{burn}}+1}^{n} X_t(a)
\right\|_{\mathrm{op}}
\le
\sqrt{
2(n-t_{\mathrm{burn}})
\log\frac{2|\mathcal S||\mathcal A|}{\delta}
}
+
\frac{2}{3}\log\frac{2|\mathcal S||\mathcal A|}{\delta}
\]
with probability at least \(1-\delta/|\mathcal A|\).

A union bound over all \(a\in\mathcal A\) shows that, with probability at least \(1-\delta\),
\[
\left\|
\sum_{t=t_{\mathrm{burn}}+1}^{n}X_t(a)
\right\|_{\mathrm{op}}
\le
\sqrt{2(n-t_{\mathrm{burn}})\log\frac{2|\mathcal S||\mathcal A|}{\delta}}
+
\frac{2}{3}\log\frac{2|\mathcal S||\mathcal A|}{\delta}
\]
simultaneously for all \(a\in\mathcal A\).

Plugging this into~\eqref{equ: pre-martingale sigma relation} gives that with probability at least $1-\delta$, for every \(a\in\mathcal A\),
\[
\lambda_{\min}(\widetilde {B}_a^\top \widetilde{B}_a)
\ge
\pi_{\min}\lambda_{\min}(\widetilde {B}^\top \widetilde{B})
-
\sqrt{2(n-t_{\mathrm{burn}})\log\frac{2|\mathcal S||\mathcal A|}{\delta}}
-
\frac{2}{3}\log\frac{2|\mathcal S||\mathcal A|}{\delta}.
\]

Recall that Lemma~\ref{lem: action-restricted predictive belief full-rankedness_formal}satisfies the assumptions of Proposition~\ref{prop: predictive belief full-rankedness_formal}, so the results of Appendix~\ref{apx: proof for predictive belief full-rankedness} apply directly. In particular, \eqref{equ: tilde B sigma min squared bound} yields with probability at least $1-\delta_{1}-\delta_{2}$,
\[\lambda_{\min}(\widetilde {B}^\top \widetilde{B})\geq\frac{1}{12}m|\mathcal A|\,\pi_{\min}\,\sigma_{T}^2\,\sigma_{O}^2
\left(d_{\min}^{\pi}\right)^2.\]

Finally, setting $\delta$ to $\delta_{3}$, combining the two inequalities, and plugging in $\sigma_{\min}(\bar{B}_{a})^{2}\geq \sigma_{\min}(\widetilde{B}_{a})^{2}=\lambda_{\min}(\widetilde {B}_a^\top \widetilde{B}_a)$ proves the claim. Furthermore, setting $\delta_{1}=\delta_{2}=\delta_{3}=\frac{\delta}{3}$ recovers the statement in Lemma~\ref{lem: sigma B_a proof}.

\section{}\label{apx: proof of approximate permutation}
In this appendix, we present the proof for Theorem~\ref{thm: approximate permutation for undercomplete POMDPs}.
We work on the intersection of the high-probability
events in Lemma~\ref{lem: sigma B_a proof} and Corollary~\ref{cor:universal length-2 quadratic bound for O}, each invoked
with confidence parameter $\delta/2$.
By a union bound, this event has probability at least
$1-\delta$.

Fix any candidate $\hat\theta=(\hat\beta,\hat\varphi)$
satisfying
\[
\ell_n(\theta^*)-\ell_n(\hat\theta)\le G_n.
\]
Apply Corollary~\ref{cor:universal length-2 quadratic bound for O} with approximation radius
$\varepsilon=1/n$. There exists
$\tilde\theta=(\tilde\beta,\tilde\varphi)
\in\Theta$ such that
\[
\|\tilde\theta-\hat\theta\|_\infty
\le \frac1n
\]
and
\begin{equation}
\begin{aligned}
&\sum_{k=1}^n
\left(
\sum_{o,o'\in\Omega}
\left|
\sum_{s\in\mathcal S}
\bar b_{k,\tilde\theta}(s)
\mathbb P_{\tilde\theta}(o,o'\mid s,a_k)
-
\sum_{s\in\mathcal S}
\bar b_{k,\theta^*}(s)
\mathbb P_{\theta^*}(o,o'\mid s,a_k)
\right|
\right)^2\\
&\hspace{2cm}\le
\frac8{\pi_{\min}}
\left(
G_n+2d\log(2|\Omega|n)+2\log(4/\delta)+2
\right)+4.
\end{aligned}
\label{eq:uniform_joint_prediction}
\end{equation}
Define
\[
H_n(\delta)
=
G_n+2d\log(2|\Omega|n)+2\log(4/\delta)+2
\]
and
\[
\eta_n^2
=
\frac1n
\left(
\frac{8H_n(\delta)}{\pi_{\min}}+4
\right).
\]
In particular,
\[
\eta_n
=
O\left(
\sqrt{
\frac{G_n+d\log(2|\Omega|n)+\log(1/\delta)+1}
{n\pi_{\min}}
}
\right).
\]

\paragraph{Step 1: Constructing the predictive tensors.}
Denote by $\bar B_a(\theta)$ the
action-restricted predictive-belief matrix
whose rows are $\bar b_{k,\theta}^{\top}$,
for $k\in \mathcal{I}_a$.

Define the forward observation matrix
\[
F_a(\theta)=O_\varphi T_\beta^a
\in\mathbb R^{|\Omega|\times |\mathcal{S}|},
\qquad \theta=(\beta,\varphi).
\]
For each $a\in\mathcal A$, introduce the tensor
$\mathcal X_a(\theta)\in
\mathbb R^{|\mathcal{I}_a|\times |\Omega|\times |\Omega|}$ defined by
\[
[\mathcal X_a(\theta)]_{k,o,o'}
=
\frac1{\sqrt n}
\sum_{s\in\mathcal S}
\bar b_{k,\theta}(s)
\mathbb P_\theta(o,o'\mid s,a),
\qquad k\in \mathcal{I}_a.
\]
Since
\[
\mathbb P_\theta(o,o'\mid s,a)
=
O_\varphi(o\mid s)
\sum_{s'\in\mathcal S}
T_\beta^a(s'\mid s)O_\varphi(o'\mid s'),
\]
the tensor admits the rank-$S$ decomposition
\begin{equation}
\mathcal X_a(\theta)
=
\sum_{s=1}^S
\left(
\frac{\bar B_a(\theta)(:,s)}{\sqrt n}
\right)
\otimes O_\varphi(:,s)
\otimes F_a(\theta)(:,s).
\label{eq:predictive_tensor_decomposition}
\end{equation}

The normalization by $\sqrt n$ will ensure
that the factor matrices are uniformly bounded
in operator norm.

For every action $a$, the inequality
$\sum_j x_j^2\le(\sum_j|x_j|)^2$,
together with
\eqref{eq:uniform_joint_prediction}, yields
\begin{equation}
\begin{aligned}
\|\mathcal X_a(\tilde\theta)
-\mathcal X_a(\theta^*)\|_F^2
&=
\frac1n\sum_{k\in \mathcal{I}_a}\sum_{o,o'}
\left|
\sum_s
\bar b_{k,\tilde\theta}(s)
\mathbb P_{\tilde\theta}(o,o'\mid s,a)
-
\sum_s
\bar b_{k,\theta^*}(s)
\mathbb P_{\theta^*}(o,o'\mid s,a)
\right|^2\\
&\le \eta_n^2.
\end{aligned}
\label{eq:uniform_tensor_perturbation}
\end{equation}
Thus, simultaneously for all $a\in\mathcal A$,
\[
\|\mathcal X_a(\widetilde\theta)
-\mathcal X_a(\theta^*)\|_F
\le \eta_n.
\]

Notice that the same bound also controls
the one-observation marginals.
Indeed, summing over $o'$ gives
\[
\sum_{o'}
\mathbb P_\theta(o,o'\mid s,a)
=
O_\varphi(o\mid s),
\]
and therefore
\begin{equation}
\frac1n\sum_{k\in \mathcal{I}_a}\sum_o
\left|
\sum_s
\bar b_{k,\tilde\theta}(s)
O_{\tilde\varphi}(o\mid s)
-
\sum_s
\bar b_{k,\theta^*}(s)
O_{\varphi^*}(o\mid s)
\right|^2\le \eta_n^2.
\label{eq:uniform_marginal_perturbation}
\end{equation}
Hence Corollary~\ref{cor:universal length-2 quadratic bound for O} supplies both the joint
and marginal perturbation bounds using
the same approximating parameter
$\tilde\theta$.

\paragraph{Step 2: Conditioning of the three factors.}
By Assumptions~\ref{ass: minimum policy mass}, \ref{ass:alpha_undercomplete}, \ref{ass: full-rank transition kernels}, and \ref{ass: ergodicity},
the observation and transition kernels satisfy
\[
\sigma_{\min}(O_{\varphi^*})\ge\sigma_O,
\qquad
\sigma_{\min}(T_{\beta^*}^a)\ge\sigma_T
\quad\text{for every }a\in\mathcal A.
\]
Consequently,
\begin{equation}
\sigma_{\min}(F_a(\theta^*))
=
\sigma_{\min}(O_{\varphi^*}T_{\beta^*}^a)
\ge\sigma_O\sigma_T.
\label{eq:forward_factor_conditioning}
\end{equation}

Furthermore, Lemma~\ref{lem: sigma B_a proof} guarantees that,
simultaneously for all $a\in\mathcal A$,
\[
\sigma_{\min}(\bar B_a(\theta^*))^2
\ge
c(n-t_{\mathrm{burn}})
|\mathcal A|\pi_{\min}^2
\sigma_T^2\sigma_O^2
(d_{\min}^{\pi})^2.
\]
For $n\ge2t_{\mathrm{burn}}$, this implies
\begin{equation}
\sigma_{\min}
\left(
\frac{\bar B_a(\theta^*)}{\sqrt n}
\right)
\ge
\sqrt{\frac{c|\mathcal A|}{2}}\,
\pi_{\min}\sigma_T\sigma_O
d_{\min}^{\pi}
=:\kappa_B>0.
\label{eq:belief_factor_conditioning}
\end{equation}

Set
\[
\kappa
=
\min\{\kappa_B,\sigma_O,\sigma_O\sigma_T\}.
\]
Equations
\eqref{eq:forward_factor_conditioning} and
\eqref{eq:belief_factor_conditioning}
show that all three factor matrices in
\eqref{eq:predictive_tensor_decomposition}
have full column rank, with smallest
singular values at least $\kappa$.

Moreover, because predictive beliefs are
probability vectors,
\[
\left\|
\frac{\bar B_a(\theta)}{\sqrt n}
\right\|_2
\le
\left\|
\frac{\bar B_a(\theta)}{\sqrt n}
\right\|_F
\le1
\]
for every $\theta\in\Theta$.
Similarly, since the columns of
$O_\varphi$ and $F_a(\theta)$
are probability vectors,
\[
\|O_\varphi\|_2\le\sqrt{|\mathcal{S}|},
\qquad
\|F_a(\theta)\|_2\le\sqrt{|\mathcal{S}|}.
\]

Thus the three factors have quantitative
full robust Kruskal rank $|\mathcal{S}|$.
For $|\mathcal{S}|\ge2$, their Kruskal ranks satisfy
$3|\mathcal{S}|\ge2|\mathcal{S}|+2$, so the Kruskal uniqueness
condition holds.

\paragraph{Step 3: Quantitative stability via robust Kruskal.}
We apply Theorem~\ref{thm: dimension-free robust kruskal} to the two tensor decompositions
in \eqref{eq:predictive_tensor_decomposition}.

Fix $a\in\mathcal A$, and abbreviate
\[
B^{\mathrm{norm}}_a(\theta^*)=\frac{\bar B_a(\theta^*)}{\sqrt n},
\qquad
B_a^{\mathrm{norm}}(\tilde{\theta})
=\frac{\bar B_a(\tilde\theta)}{\sqrt n},
\]
\[
O^*=O_{\varphi^*},
\qquad
\widetilde O=O_{\tilde\varphi},
\qquad
F_a^*=F_a(\theta^*),
\qquad
\widetilde F_a=F_a(\tilde\theta).
\]
Thus
\[
\mathcal X_a(\theta^*)=[B_a^{\mathrm{norm}}(\theta^*),O^*,F_a^*],
\qquad
\mathcal X_a(\widetilde\theta)
=[B_a^{\mathrm{norm}}(\tilde{\theta}),\widetilde O,\widetilde F_a].
\]

By \eqref{eq:forward_factor_conditioning} and
\eqref{eq:belief_factor_conditioning},
\[
\sigma_{\min}(B_a^{\mathrm{norm}}(\theta^*))\ge\kappa_B,
\qquad
\sigma_{\min}(O^*)\ge\sigma_O,
\qquad
\sigma_{\min}(F_a^*)\ge\sigma_O\sigma_T.
\]
In particular, all three factors have full robust
Kruskal rank $S$, with parameters
\[
\tau_B=\kappa_B^{-1},
\qquad
\tau_O=\sigma_O^{-1},
\qquad
\tau_F=(\sigma_O\sigma_T)^{-1}.
\]
Indeed, the minimum singular value of every
column submatrix is at least that of the full
matrix. Consequently, for $|\mathcal{S}|\ge2$,
\[
\operatorname{Krank}_{\tau_B}(B_a^{\mathrm{norm}}(\theta^*))
+
\operatorname{Krank}_{\tau_O}(O^*)
+
\operatorname{Krank}_{\tau_F}(F_a^*)
=3|\mathcal{S}|\ge2|\mathcal{S}|+2.
\]

Furthermore, every column of $B_a^{\mathrm{norm}}(\theta^*)$ and
$B_a^{\mathrm{norm}}(\tilde{\theta})$ has Euclidean norm at most one,
since the predictive beliefs are probability
vectors and the matrices are normalized by
$\sqrt n$. Every column of $O^*$,
$\widetilde O$, $F_a^*$, and $\widetilde F_a$
is also a probability vector and hence has
Euclidean norm at most one.
Thus both decompositions are
$(1,1,1)$-bounded.

Let
\[
D_{\mathrm{RK}}
=
\max\left\{
1,\,
|\mathcal{S}|^c
\prod_{L\in\{B,O,F\}}
\vartheta_5(\tau_L,1)
\right\},
\]
where $c$ and $\vartheta_5$ are the universal
constant and polynomial in Theorem~\ref{thm: dimension-free robust kruskal}.
Importantly, $D_{\mathrm{RK}}$ depends
polynomially on
\[
|\mathcal{S}|,\quad
\kappa_B^{-1},\quad
\sigma_O^{-1},\quad
\sigma_T^{-1},
\]
but not on the ambient dimension $|\mathcal{I}_a|$,
and hence not on $n$.

Set
\[
\xi_n=D_{\mathrm{RK}}\eta_n.
\]
Under the stated sample-size requirement,
we may assume
\[
\xi_n\le\frac1{4\sqrt{|\Omega|}}<1.
\]
By \eqref{eq:uniform_tensor_perturbation}
and Theorem~\ref{thm: dimension-free robust kruskal}, there exist an
$|\mathcal{S}|\times |\mathcal{S}|$ permutation matrix $P_a$
and diagonal matrices
$\Lambda_{B,a},\Lambda_{O,a},\Lambda_{F,a}$
such that
\begin{equation}
\begin{aligned}
\|B_a^{\mathrm{norm}}(\tilde{\theta})
-B_a^{\mathrm{norm}}(\theta^*)P_a\Lambda_{B,a}\|_F
&\le\xi_n,\\
\|\widetilde O
-O^*P_a\Lambda_{O,a}\|_F
&\le\xi_n,\\
\|\widetilde F_a
-F_a^*P_a\Lambda_{F,a}\|_F
&\le\xi_n,
\end{aligned}
\label{eq:robust_kruskal_scaled_factors}
\end{equation}
and
\begin{equation}
\|\Lambda_{B,a}\Lambda_{O,a}
\Lambda_{F,a}-I_{|\mathcal{S}|}\|_F
\le\xi_n.
\label{eq:robust_kruskal_scaling}
\end{equation}

We next remove the diagonal scaling ambiguity.
Write $\lambda_{O,s}$ and $\lambda_{F,s}$
for the $s$-th diagonal entries of
$\Lambda_{O,a}$ and $\Lambda_{F,a}$.
Since every column of $\widetilde O$
and $O^*P_a$ sums to one,
\eqref{eq:robust_kruskal_scaled_factors}
implies
\[
|1-\lambda_{O,s}|
=
\left|
\mathbf 1_{|\Omega|}^\top
\bigl(
\widetilde O(:,s)
-\lambda_{O,s}(O^*P_a)(:,s)
\bigr)
\right|
\le\sqrt{|\Omega|}\,\xi_n.
\]
Similarly, because the columns of
$\widetilde F_a$ and $F_a^*P_a$
are probability vectors,
\[
|1-\lambda_{F,s}|
\le\sqrt{\Omega}\,\xi_n.
\]
Therefore,
\begin{equation}
\|\Lambda_{O,a}-I_{|\mathcal{S}|}\|_F
\le\sqrt{|\mathcal{S}||\Omega|}\,\xi_n,
\qquad
\|\Lambda_{F,a}-I_{|\mathcal{S}|}\|_F
\le\sqrt{|\mathcal{S}||\Omega|}\,\xi_n.
\label{eq:observation_scaling_control}
\end{equation}

Since $\sqrt{|\Omega|}\,\xi_n\le1/4$,
we have
\[
|\lambda_{O,s}\lambda_{F,s}|
\ge(1-\sqrt{|\Omega|}\,\xi_n)^2
\ge\frac12.
\]
Writing $\lambda_{B,s}$ for the
corresponding diagonal entry of
$\Lambda_{B,a}$, we obtain from
\eqref{eq:robust_kruskal_scaling}
\[
\begin{aligned}
|\lambda_{B,s}-1|
&\le
\frac{
|\lambda_{B,s}\lambda_{O,s}
\lambda_{F,s}-1|
+
|1-\lambda_{O,s}\lambda_{F,s}|
}{
|\lambda_{O,s}\lambda_{F,s}|
}\\
&\le
2\left(
\xi_n+2\sqrt{|\Omega|}\,\xi_n+|\Omega|\xi_n^2
\right)\\
&\le8\sqrt{|\Omega|}\,\xi_n.
\end{aligned}
\]
Consequently,
\begin{equation}
\|\Lambda_{B,a}-I_{|\mathcal{S}|}\|_F
\le8\sqrt{|\mathcal{S}||\Omega|}\,\xi_n.
\label{eq:belief_scaling_control}
\end{equation}

Every column of $B_a^*$, $O^*$,
and $F_a^*$ has Euclidean norm at most one.
Combining
\eqref{eq:robust_kruskal_scaled_factors},
\eqref{eq:observation_scaling_control},
and \eqref{eq:belief_scaling_control}
with the triangle inequality yields
\[
\begin{aligned}
\|B_a^{\mathrm{norm}}(\tilde{\theta})-B_a^{\mathrm{norm}}(\theta^*)P_a\|_F
&\le(1+8\sqrt{|\mathcal{S}||\Omega|})\xi_n,\\
\|\widetilde O-O^*P_a\|_F
&\le(1+\sqrt{|\mathcal{S}||\Omega|})\xi_n,\\
\|\widetilde F_a-F_a^*P_a\|_F
&\le(1+\sqrt{|\mathcal{S}||\Omega|})\xi_n.
\end{aligned}
\]

Define
\[
C_{\mathrm{RK}}
=
(1+8\sqrt{|\mathcal{S}||\Omega|})D_{\mathrm{RK}}.
\]
We have established that, for every
$a\in\mathcal A$, there exists a permutation
matrix $P_a$ satisfying
\begin{equation}
\begin{aligned}
\left\|
\frac{\bar B_a(\tilde\theta)}{\sqrt n}
-
\frac{\bar B_a(\theta^*)}{\sqrt n}P_a
\right\|_F
&\le C_{\mathrm{RK}}\eta_n,\\
\|O_{\tilde\varphi}
-O_{\varphi^*}P_a\|_F
&\le C_{\mathrm{RK}}\eta_n,\\
\|F_a(\tilde\theta)
-F_a(\theta^*)P_a\|_F
&\le C_{\mathrm{RK}}\eta_n.
\end{aligned}
\label{eq:factor_recovery_by_action}
\end{equation}
In particular, the factor recovery error is
linear in the tensor perturbation $\eta_n$,
and $C_{\mathrm{RK}}$ depends polynomially
on the problem parameters but not on $n$.

\paragraph{Step 4: Establishing a common permutation.}
Although \eqref{eq:factor_recovery_by_action}
initially produces an action-dependent
permutation $P_a$, we now show that all
these permutations coincide.

For any $a,b\in\mathcal A$,
the triangle inequality gives
\[
\begin{aligned}
\|O_{\varphi^*}(P_a-P_b)\|_F
&\le
\|O_{\varphi^*}P_a-O_{\tilde\varphi}\|_F
+
\|O_{\tilde\varphi}
-O_{\varphi^*}P_b\|_F\\
&\le2C_{\mathrm{RK}}\eta_n.
\end{aligned}
\]
On the other hand, if $P_a\ne P_b$,
then
\[
\begin{aligned}
\|O_{\varphi^*}(P_a-P_b)\|_F
&\ge
\sigma_{\min}(O_{\varphi^*})
\|P_a-P_b\|_F\\
&\ge2\sigma_O.
\end{aligned}
\]
Therefore, whenever
$C_{\mathrm{RK}}\eta_n<\sigma_O$, which follows from the lower-bound on $n$ in the theorem, 
we must have $P_a=P_b$.

It follows that there exists a single
permutation matrix $P$ such that,
simultaneously for all $a\in\mathcal A$,
\begin{equation}
\begin{aligned}
\|O_{\tilde\varphi}
-O_{\varphi^*}P\|_F
&\le C_{\mathrm{RK}}\eta_n,\\
\|O_{\tilde\varphi}T_{\tilde\beta}^a
-O_{\varphi^*}T_{\beta^*}^aP\|_F
&\le C_{\mathrm{RK}}\eta_n.
\end{aligned}
\label{eq:common_permutation_recovery}
\end{equation}

\paragraph{Step 5: Recovering the transition kernels.}
Recall that
\[
F_a(\theta)=O_\varphi T_\beta^a.
\]
For each $a\in\mathcal A$, we have
\[
\begin{aligned}
&O_{\varphi^*}P
\left(
T_{\tilde\beta}^a
-P^\top T_{\beta^*}^aP
\right)\\
&\quad=
\left(
O_{\tilde\varphi}T_{\tilde\beta}^a
-O_{\varphi^*}T_{\beta^*}^aP
\right)
-
\left(
O_{\tilde\varphi}-O_{\varphi^*}P
\right)T_{\tilde\beta}^a.
\end{aligned}
\]
Since
$\sigma_{\min}(O_{\varphi_*}P)\ge\sigma_O$,
it follows that
\[
\begin{aligned}
&
\left\|
T_{\tilde\beta}^a
-P^\top T_{\beta^*}^aP
\right\|_F\\
&\quad\le
\frac1{\sigma_O}
\left(
\|O_{\tilde\varphi}T_{\tilde\beta}^a
-O_{\varphi^*}T_{\beta^*}^aP\|_F
+
\|O_{\tilde\varphi}-O_{\varphi^*}P\|_F
\|T_{\tilde\beta}^a\|_2
\right).
\end{aligned}
\]
Because $T_{\tilde\beta}^a$ is
column stochastic,
$\|T_{\tilde\beta}^a\|_2\le\sqrt{|\mathcal{S}|}$.
Combining this with
\eqref{eq:common_permutation_recovery}
gives
\begin{equation}
\left\|
P^\top T_{\beta^*}^aP
-T_{\tilde\beta}^a
\right\|_F
\le
\frac{(1+\sqrt{|\mathcal{S}|})C_{\mathrm{RK}}}
{\sigma_O}\eta_n.
\label{eq:transition_frobenius_recovery}
\end{equation}
This bound holds simultaneously for all
$a\in\mathcal A$, with the same permutation $P$.

\paragraph{Step 6: Returning to the candidate
parameter and concluding.}
Recall that Corollary~\ref{cor:universal length-2 quadratic bound for O} guarantees
\[
\|\tilde\theta-\hat\theta\|_\infty
\le\frac1n.
\]
In particular,
\[
\|O_{\tilde\varphi}
-O_{\hat\varphi}\|_{1,\mathrm{ent}}
\le\frac{|\Omega||\mathcal{S}|}{n},
\qquad
\|T_{\tilde\beta}^a
-T_{\hat\beta}^a\|_{1,\mathrm{ent}}
\le\frac{|\mathcal{S}|^2}{n}.
\]

For any $r\times q$ matrix $A$,
\[
\|A\|_{1,\mathrm{ent}}
\le\sqrt{rq}\|A\|_F.
\]
Consequently, by
\eqref{eq:common_permutation_recovery},
\begin{equation}
\|O_{\varphi^*}P
-O_{\hat\varphi}\|_{1,\mathrm{ent}}
\le
\sqrt{|\Omega||\mathcal{S}|}\,C_{\mathrm{RK}}\eta_n
+\frac{|\Omega||\mathcal{S}|}{n}.
\label{eq:final_observation_recovery}
\end{equation}

Similarly, by
\eqref{eq:transition_frobenius_recovery},
for every $a\in\mathcal A$,
\begin{equation}
\|P^\top T_{\beta^*}^aP
-T_{\hat\beta}^a\|_{1,\mathrm{ent}}
\le
\frac{|\mathcal{S}|(1+\sqrt{|\mathcal{S}|})C_{\mathrm{RK}}}
{\sigma_O}\eta_n
+\frac{|\mathcal{S}|^2}{n}.
\label{eq:final_transition_recovery}
\end{equation}

Recall from the definition of $\eta_n$
in Step~1,
\[
\eta_n
=
O\left(
\sqrt{
\frac{H_{n}(\delta)}
{n\pi_{\min}}
}
\right).
\]
Moreover, by Lemma~\ref{lem: sigma B_a proof},
\[
\kappa_B^{-1}
=
O\left(
\frac1{
\pi_{\min}\sigma_T\sigma_O
d_{\min}^{\pi}
}
\right),
\]
up to polynomial factors in
$|\mathcal A|$.
Therefore, $C_{\mathrm{RK}}$ is bounded
polynomially in
\[
|\mathcal A|,\quad
\frac1{\pi_{\min}},\quad
\frac1{d_{\min}^{\pi}},\quad
\frac1{\sigma_T},\quad
\frac1{\sigma_O},\quad
|\mathcal{S}|,\quad |\Omega|.
\]

Under the stated sample-size requirement,
$n$ is sufficiently large that
$n\ge2t_{\mathrm{burn}}$ (since $t_{\mathrm{burn}}$ is polynomial in all the expressions in \eqref{equ:relevent parameters}) and
\[
D_{\mathrm{RK}}\eta_n
\le\frac1{4\sqrt{|\Omega|}},
\qquad
C_{\mathrm{RK}}\eta_n<\sigma_O.
\]
The $n^{-1}$ approximation terms in
\eqref{eq:final_observation_recovery}
and \eqref{eq:final_transition_recovery}
can then be absorbed into the
$n^{-1/2}$ bounds.

Hence there exist problem-dependent
constants $C_O,C_T>0$, polynomially
bounded in the quantities above, such that
\begin{equation}
\|O_{\varphi^*}P-O_{\hat\varphi}\|_{1,\mathrm{ent}}
\le
C_O\sqrt{
\frac{H_{n}(\delta)}{n}
}.
\label{eq:final_uniform_observation_bound}
\end{equation}
Furthermore, simultaneously for every
$a\in\mathcal A$,
\begin{equation}
\|P^\top T_{\beta^*}^aP
-T_{\hat\beta}^a\|_{1,\mathrm{ent}}
\le
C_T\sqrt{
\frac{H_{n}(\delta)}{n}
}.
\label{eq:final_uniform_transition_bound}
\end{equation}

The high-probability events used in
Steps~1--2 hold simultaneously with
probability at least $1-\delta$.
Since Corollary~\ref{cor:universal length-2 quadratic bound for O} holds uniformly over
all candidate parameters, the preceding
argument applies to every $\hat\theta$
satisfying
$\ell_n(\theta^*)-\ell_n(\hat\theta)\le G_n$.

This proves the claimed recovery rates.

\section{}\label{apx: replacing the full-rank transition kernels assumption}
In this appendix, we explain how the proofs for the theoretical results in Section~\ref{subsec: main result-undercomplete} can be modified to work when we replace Assumption~\ref{ass: full-rank transition kernels} with aggregate transition conditioning:
\[
\sigma_{\min}(T_{\beta^*})\geq \bar{\sigma}_T>0,
\qquad
T_{\beta^*}\coloneqq [T_{\beta^*}^{a_1},\ldots,T_{\beta^*}^{a_{|\mathcal A|}}],
\]
and
\[
\sigma_{\min}(\widetilde T_{\beta^*})\geq \widetilde\sigma_T>0,
\qquad
\widetilde T_{\beta^*}\coloneqq [(T_{\beta^*}^{a_1})^\top,\ldots,(T_{\beta^*}^{a_{|\mathcal A|}})^\top].
\]

In particular, we identify the two intermediate results whose proofs need to be modified: Proposition~\ref{prop: predictive belief full-rankedness_formal} and Theorem~\ref{thm: approximate permutation for undercomplete POMDPs}.

\subsection{}
First, we explain how the proof of Proposition~\ref{prop: predictive belief full-rankedness_formal} can be modified.

The only part of the proof that needs adjustment is Lemma~\ref{lemma: predictive belief isotropy}.

From \eqref{equ:before using the sigma T assumption}:
\[\mathbb E[\bar b_{t+1}\bar b_{t+1}^\top]
\succeq
\pi_{\min}\sigma_{O}^2\left(\frac{7d_{\min}^{\pi}}{8}\right)^2
\sum_{a\in\mathcal A} T^{a}_{\beta^*}(T^{a}_{\beta^*})^\top.\]

Then, $\sigma_{\min}(T_{\beta^*}) \geq \bar{\sigma}_T > 0$ implies that
\[\sum_{a\in\mathcal A} T^{a}_{\beta^*}(T^{a}_{\beta^*})^\top\succeq (\bar{\sigma}_{T})^{2}.\]

Combining the two inequalities gives us
\[\mathbb E[\bar b_{t+1}\bar b_{t+1}^\top]
\succeq
\,\pi_{\min}\,(\bar{\sigma}_{T})^2\,\sigma_{O}^2
\left(\frac{7d_{\min}^{\pi}}{8}\right)^2 I,\]
which concludes the proof of Lemma~\ref{lemma: predictive belief isotropy}.

\subsection{}
Then, we explain how the proof of Theorem~\ref{thm: approximate permutation for undercomplete POMDPs} can be modified.

Recall that the action-indexed tensors admit the decomposition
\[
\mathcal X_a(\theta^*)
=\sum_{s=1}^S
\frac{\bar B_a(\theta^*)(:,s)}{\sqrt n}
\otimes O_{\varphi^*}(:,s)
\otimes (O_{\varphi^*} T_{\beta^*}^a)(:,s).
\]
When individual transition matrices are singular,
robust Kruskal uniqueness need not apply separately
to each $\mathcal X_a(\theta^*)$. Instead, we exploit the shared
observation factor across actions.

Marginalizing each tensor over its third mode gives
$\bar B_a(\theta^*) O_{\varphi^*}^\top/\sqrt n$. By the conditioning
of $O_{\varphi^*}$ and $\bar B_a(\theta^*)/\sqrt n$ from Assumption~\ref{ass:alpha_undercomplete} and Lemma~\ref{lem: sigma B_a proof}, agreement of these
marginals implies that the candidate and true factors
are related, up to perturbation, by a single
invertible change of basis $H$, common to all actions.

Recall that we define $D_o=\operatorname{diag}(O_{\varphi^*}(o\mid\cdot))$
and defining $\widehat D_o$ analogously, comparison
of the two-step tensor slices yields
\[
T_{\beta^*}^aD_oH
\approx HT^a_{\hat{\beta}}\widehat D_o.
\]
Summing over $o$ gives
$T_{\beta^*}^aH\approx HT^a_{\hat{\beta}}$, and hence
\[
T_{\beta^*}^a(D_oH-H\widehat D_o)\approx 0
\qquad \text{for every }a,o.
\]
The aggregate condition now implies
\[
\|D_oH-H\widehat D_o\|_F
\leq \frac{1}{\widetilde{\sigma}_{T}}
\left(
\sum_a
\|T_{\beta^*}^a(D_oH-H\widehat D_o)\|_F^2
\right)^{1/2}.
\]
Thus the approximate commutation relation holds
without requiring any individual $T_\star^a$
to be invertible.

Since $O_{\varphi^*}$ has full column rank, its columns
are distinct. The approximate commutation relations,
together with the nonsingularity of $H$ and the
column-stochastic normalization of the observation
kernels, imply that $H$ is close to a permutation
matrix $P$. Consequently,
\[
O_{\hat{\varphi}}\approx O_{\varphi^*} P,
\qquad
T^a_{\hat{\beta}}\approx P^\top T_{\beta^*}^aP
\quad\text{for all }a.
\]
The quantitative perturbation argument follows
the same steps as the original proof, with the
aggregate conditioning parameter $\widetilde{\sigma}_T$
replacing the individual transition-conditioning
parameter in the joint alignment step.

\section{}\label{apx: short reward-alignment corollary}
Since we assume that rewards are observed and $\mathcal R$ is finite, we can incorporate rewards into the observation space by defining the augmented observation kernels
\[
\overline O_\theta^a((o,r)\mid s)
=
O_\theta(o\mid s)R_\theta^a(r\mid s).
\]
Here we use the conditional independence of the reward and observation given the state and action. Although $\overline O_\theta^a$ depends on $a$, marginalizing over rewards recovers the original observation kernel:
\[
\sum_{r\in\mathcal R}
\overline O_\theta^a((o,r)\mid s)
=
O_\theta(o\mid s).
\]
In particular, quantitative full-column-rankness is preserved, since
\[
\sigma_{\min}(\overline O_\theta^a)
\geq
\frac{\sigma_{\min}(O_\theta)}
{\sqrt{K_{R}}},
\]
where $K_{R}$ denotes the size of the reward alphabet.

The same argument of Theorem~\ref{thm: approximate permutation for undercomplete POMDPs} therefore extends to these action-dependent augmented observation kernels. Any action-dependent permutations arising from the recovery can be aligned using the common observation kernel. Consequently, the augmented kernels are recovered under a common permutation $P$ at the rate of Theorem~\ref{thm: approximate permutation for undercomplete POMDPs}. Marginalizing over the original observations then recovers the reward kernels, and marginalization cannot increase the entrywise $\ell_1$ error. This yields the claimed bound.

\begin{remark}
    Notice that our construction of $\overline{O}_{\theta}^{a}$ implies that it is an action-dependent observation kernel, whereas Theorem~\ref{thm: approximate permutation for undercomplete POMDPs} assumes the observation kernel is action-independent. However, the exact same argument of Theorem~\ref{thm: approximate permutation for undercomplete POMDPs} applies if we use action-dependent observation kernels.
\end{remark}

\begin{remark}
The finite-alphabet restriction can be relaxed when only conditional mean rewards are of interest. For bounded continuous rewards, an auxiliary Bernoulli observation with success probability proportional to the realized reward reduces mean recovery to the finite-alphabet case. For conditionally sub-Gaussian rewards, one can instead estimate reward--observation moments using importance weighting and standard concentration arguments, then recover the conditional means using the full-column-rank observation kernel. We omit these extensions for brevity.
\end{remark}

\section{}\label{apx: example for constrained model class}

\noindent\textbf{Example: Physics-constrained thermal control of a smart building.}

POMDPs with state-based rewards have been used to model and optimize HVAC systems for energy cost and occupant comfort. This example is adapted from \citet{murugesan2020less}.

Consider thermal control in a smart building. Let the continuous physical state be
\[
X_t = \bigl(T_{\mathrm{air},t},T_{\mathrm{mass},t}\bigr)^\top,
\]
where \(T_{\mathrm{air},t}\) is the indoor-air temperature and
\(T_{\mathrm{mass},t}\), representing the effective temperature of walls and
furniture, is not directly observed. A reduced-order stochastic thermal model
may take the form
\[
dX_t
=
A(\beta)X_t\,dt
+
B(\beta)u_t\,dt
+
g_\beta(T_{\mathrm{out},t})\,dt
+
G(\beta)\,dW_t,
\]
where \(\beta\) contains physical parameters such as thermal resistances and
capacitances, \(u_t\) is the HVAC control input, and
\(T_{\mathrm{out},t}\) is the outdoor temperature.
The matrices \(A(\beta)\) and \(B(\beta)\) describe the intrinsic thermal
dynamics and the effect of HVAC control, respectively,
\(g_\beta(T_{\mathrm{out},t})\) represents heat transfer due to the outdoor
temperature, and \(G(\beta)dW_t\) models unobserved thermal disturbances.

Sensors provide noisy measurements
\[
Y_k = C X_{k\Delta} + \varepsilon_k,
\]
where \(C\) specifies which components of the physical state are measured,
\(\varepsilon_k\) is sensor noise, and \(\Delta>0\) is the sampling interval.

A POMDP approximation can use semantic thermal regimes such as cold, neutral, and hot, HVAC actions such as off/medium/high heating, and observation bins based on measured temperature. For state bins \(B_i\) and observation bins \(C_o\), the induced approximate POMDP kernels are
\[
T_\beta^a(i,j)
=
\mathbb{P}_\beta\!\left(
X_{(k+1)\Delta} \in B_j
\mid
X_{k\Delta} \in B_i,\,
u_k = a
\right),
\qquad
O_\varphi(o \mid i)
=
\mathbb{P}_\varphi\!\left(
Y_k \in C_o
\mid
X_{k\Delta} \in B_i
\right).
\]

Thus, \(T_\beta^a\) is not an arbitrary matrix: all actions share the same low-dimensional physical parameter \(\beta\), and the semantic states naturally support state-based comfort or safety rewards.

Our result does not require an unconstrained black-box MLE. One may optimize likelihood over the physics-constrained class
\[
\Theta_{\mathrm{phys}}
=
\left\{
\left(
\{T_\beta^a\}_{a\in\mathcal{A}},
O_\varphi
\right)
:
\beta \in \mathcal{B},
\;
\varphi \in \Phi
\right\},
\]
where \(\mathcal{B}\) encodes relevant domain knowledge and enforces model structures:
\[
(\hat{\beta},\hat{\varphi})
\in
\arg\max_{(\beta,\varphi)\in\mathcal{B}\times\Phi}
\ell_n\!\left(
\{T_\beta^a\}_{a\in\mathcal{A}},
O_\varphi
\right).
\]

Our theorem then applies to any exact or approximate optimizer whose candidate achieves the required likelihood gap. Hence, the PDE/SDE structure and semantic prior knowledge are retained rather than discarded; the result converts the likelihood accuracy of a structured estimator into recovery guarantees for the induced POMDP kernels.

\section{}\label{apx: justification for distinct observation columns}
In this appendix, we show why Assumption~\ref{ass:distinct-observation-columns} is necessary for achieving approximate POMDP kernel recovery up to latent-state relabeling. We emphasize that our claim with the following proposition is not that violating the assumption means every such POMDP is non-identifiable, but rather that it prevents a uniform permutation-recovery guarantee over general POMDPs.

\begin{proposition}
\label{prop:necessity-distinct-observation-columns}
Consider a finite POMDP with parameter
\[
    \theta
    =
    \bigl(O,\{T^a\}_{a\in\mathcal A},b_1\bigr),
\]
where $b_{1}$ is the initial belief, and suppose that there exist two distinct latent states
$i,j\in\mathcal S$ such that
\[
    O(\cdot\mid i)=O(\cdot\mid j).
\]
Assume further that the transition kernels are strictly positive and that,
for some length-$(L-1)$ action string $\alpha$, the corresponding
$L$-step observation matrix $M_\alpha$ has full column rank.

Then there exists another valid POMDP parameter
\[
    \tilde\theta
    =
    \bigl(
        O,\{\widetilde T^a\}_{a\in\mathcal A},b_1
    \bigr)
\]
such that:

\begin{enumerate}
    \item $\tilde\theta$ is not related to $\theta$ by any permutation
    of the latent states;

    \item for every horizon $H$, every action sequence
    $a_{1:H-1}$, and every observation sequence $o_{1:H}$,
    \[
        \mathbb P_{\theta}
        \bigl(o_{1:H}\mid a_{1:H-1}\bigr)
        =
        \mathbb P_{\tilde\theta}
        \bigl(o_{1:H}\mid a_{1:H-1}\bigr);
    \]

    \item the transformed model remains $L$-step observable, and in fact
    its $L$-step observation matrix satisfies
    \[
        \widetilde M_\alpha=M_\alpha G
    \]
    for some invertible matrix $G$ that is not a permutation matrix.
\end{enumerate}

Consequently, if a model class permits two identical columns of $O$, then
no estimator can be uniformly consistent for recovery of the POMDP
parameters up to latent-state permutation. More precisely, there exists a constant
$\Delta>0$ such that, for every sample size $n$ and every estimator
$\hat\theta_n$,
\[
    \max\left\{
        \mathbb E_{\theta}
        \bigl[d_{\mathrm{perm}}(\hat\theta_n,\theta)\bigr],
        \mathbb E_{\tilde\theta}
        \bigl[d_{\mathrm{perm}}(\hat\theta_n,\tilde\theta)\bigr]
    \right\}
    \geq \frac{\Delta}{2},
\]
where $d_{\mathrm{perm}}$ denotes any permutation-invariant parameter
metric. Hence the estimation error cannot converge to zero, even with
infinitely many trajectories.

In this sense, Assumption~\ref{ass:distinct-observation-columns} is
information-theoretically necessary for a general permutation-level
kernel-recovery guarantee.
\end{proposition}

\begin{remark}
The entrywise strict positivity assumption on the transition kernels can be relaxed. In particular, it suffices to assume that rows $i$ and $j$ of each $T^a$ have identical support, i.e.,
\[
T^a(i\mid s)=0
\quad\Longleftrightarrow\quad
T^a(j\mid s)=0,
\qquad \forall a\in\mathcal A,\ s\in\mathcal S.
\]
Under this weaker condition, the perturbation constructed in the proof still preserves the nonnegativity of the transition kernels, provided that the perturbation is sufficiently small.
\end{remark}

\begin{proof}
Since
\[
    O(\cdot\mid i)=O(\cdot\mid j),
\]
the two latent states $i$ and $j$ induce exactly the same one-step
observation distribution.  For each observation $o$, define the diagonal
observation operator
\[
    D_o
    :=
    \operatorname{diag}
    \bigl(
        O(o\mid 1),\ldots,O(o\mid |\mathcal S|)
    \bigr).
\]
The restriction of $D_o$ to coordinates $i$ and $j$ is therefore a
multiple of the identity:
\[
    (D_o)_{ii}=(D_o)_{jj}
    \qquad
    \text{for every }o.
\]

We construct an invertible, non-permutation matrix $G$ that only mixes
coordinates $i$ and $j$.  Let
\[
    p=(b_1)_i,
    \qquad
    q=(b_1)_j,
\]
and, on the $(i,j)$ coordinates, define
\[
    G_\varepsilon^{(i,j)}
    :=
    \begin{pmatrix}
        1-\varepsilon q & \varepsilon p\\
        \varepsilon q & 1-\varepsilon p
    \end{pmatrix},
\]
while $G_\varepsilon$ acts as the identity on all remaining coordinates.
For sufficiently small $\varepsilon>0$, $G_\varepsilon$ is invertible
and is not a permutation matrix.  Moreover,
\[
    \mathbf 1^\top G_\varepsilon=\mathbf 1^\top,
    \qquad
    G_\varepsilon b_1=b_1.
\]
Because each $D_o$ is scalar on the $(i,j)$ block,
\[
    D_oG_\varepsilon=G_\varepsilon D_o
    \qquad
    \text{for every }o.
\]
For brevity, write $G:=G_\varepsilon$.

Now define
\[
    \widetilde O:=O,
    \qquad
    \widetilde T^a:=G^{-1}T^aG,
    \qquad a\in\mathcal A.
\]
Since $\mathbf 1^\top G=\mathbf 1^\top$, we also have
\[
    \mathbf 1^\top G^{-1}=\mathbf 1^\top.
\]
Hence
\[
    \mathbf 1^\top \widetilde T^a
    =
    \mathbf 1^\top G^{-1}T^aG
    =
    \mathbf 1^\top T^aG
    =
    \mathbf 1^\top G
    =
    \mathbf 1^\top.
\]
Thus every $\widetilde T^a$ has column sums equal to one.  Since the
original transition kernels are strictly positive and
$G_\varepsilon\to I$ as $\varepsilon\to0$, continuity implies that, for
all sufficiently small $\varepsilon>0$, every
$\widetilde T^a$ remains entrywise positive. By the same reasoning, for all sufficiently small $\varepsilon > 0$, $\widetilde{T}^{a}$ has its minimum singular value bounded away from $0$. Therefore
$\tilde\theta$ is again a valid POMDP.

We next show that the two POMDPs induce exactly the same trajectory
likelihood.  Under our convention, for any observation sequence
$o_{1:H}$ and action sequence $a_{1:H-1}$,
\[
\begin{aligned}
    \mathbb P_{\theta}
    (o_{1:H}\mid a_{1:H-1})
    &=
    \mathbf 1^\top
    D_{o_H}
    T^{a_{H-1}}
    D_{o_{H-1}}
    \cdots
    T^{a_1}
    D_{o_1}
    b_1 .
\end{aligned}
\]
For the transformed model,
\[
\begin{aligned}
    &\mathbb P_{\tilde\theta}
    (o_{1:H}\mid a_{1:H-1})\\
    &\quad=
    \mathbf 1^\top
    D_{o_H}
    G^{-1}T^{a_{H-1}}G
    D_{o_{H-1}}
    \cdots
    G^{-1}T^{a_1}G
    D_{o_1}
    b_1 .
\end{aligned}
\]
Using
\[
    D_oG=GD_o,
    \qquad
    \mathbf 1^\top G^{-1}=\mathbf 1^\top,
    \qquad
    Gb_1=b_1,
\]
all intermediate copies of $G$ and $G^{-1}$ cancel, yielding
\[
    \mathbb P_{\tilde\theta}
    (o_{1:H}\mid a_{1:H-1})
    =
    \mathbb P_{\theta}
    (o_{1:H}\mid a_{1:H-1}).
\]
Thus the two models generate exactly the same conditioned trajectory
likelihood for every horizon, every action sequence, and every
observation sequence.

Importantly, this ambiguity persists even at the level of the full-rank
$L$-step observation matrix.  Fix an action string
\[
    \alpha=(a_1,\ldots,a_{L-1}),
\]
and let $M_\alpha$ denote the matrix whose row corresponding to
$\zeta=(o_1,\ldots,o_L)$ is
\[
    m_{\alpha,\zeta}^{\top}
    =
    \mathbf 1^\top
    D_{o_L}
    T^{a_{L-1}}
    D_{o_{L-1}}
    \cdots
    T^{a_1}
    D_{o_1}.
\]
For the transformed model,
\[
\begin{aligned}
    \widetilde m_{\alpha,\zeta}^{\top}
    &=
    \mathbf 1^\top
    D_{o_L}
    G^{-1}T^{a_{L-1}}G
    D_{o_{L-1}}
    \cdots
    G^{-1}T^{a_1}G
    D_{o_1}\\
    &=
    m_{\alpha,\zeta}^{\top}G,
\end{aligned}
\]
again by commuting $G$ through the observation operators and cancelling
the intermediate factors.  Consequently,
\[
        \widetilde M_\alpha=M_\alpha G.
\]
Because $G$ is invertible,
\[
    \operatorname{rank}(\widetilde M_\alpha)
    =
    \operatorname{rank}(M_\alpha)
    =
    |\mathcal S|,
\]
so the transformed model remains $L$-step observable.  In particular,
full column rank of the $L$-step observation matrix does not eliminate
this ambiguity.

We now verify that $\theta$ and $\tilde\theta$ are not equivalent up
to a permutation of the latent states.  Suppose, toward a contradiction,
that there were a permutation matrix $\Pi$ relating the two models.
Then the corresponding $L$-step observation matrices would satisfy
\[
    \widetilde M_\alpha=M_\alpha\Pi.
\]
On the other hand, we have already shown that
\[
    \widetilde M_\alpha=M_\alpha G.
\]
Hence
\[
    M_\alpha(G-\Pi)=0.
\]
Since $M_\alpha$ has full column rank,
\[
    G=\Pi,
\]
contradicting the construction of $G$ as a non-permutation matrix.
Therefore $\theta$ and $\tilde\theta$ are genuinely distinct even
after quotienting out latent-state permutations.

Let
\[
    \Delta
    :=
    d_{\mathrm{perm}}(\theta,\tilde\theta).
\]
Since the two models are not permutation-equivalent,
\[
    \Delta>0.
\]
Nevertheless, their trajectory distributions are exactly identical.
Thus, for every sample size $n$, the law of any dataset consisting of
$n$ trajectories is the same under $\theta$ and
$\tilde\theta$.  Consequently, any estimator
$\hat\theta_n$ has the same distribution under the two models.

By the triangle inequality for $d_{\mathrm{perm}}$,
\[
    \Delta
    \leq
    d_{\mathrm{perm}}(\theta,\hat\theta_n)
    +
    d_{\mathrm{perm}}(\hat\theta_n,\tilde\theta).
\]
Taking expectation with respect to the common data distribution gives
\[
\begin{aligned}
    \Delta
    &\leq
    \mathbb E_{\theta}
    \left[
        d_{\mathrm{perm}}(\hat\theta_n,\theta)
    \right]
    +
    \mathbb E_{\tilde\theta}
    \left[
        d_{\mathrm{perm}}(\hat\theta_n,\tilde\theta)
    \right].
\end{aligned}
\]
Therefore,
\[
    \max\left\{
        \mathbb E_{\theta}
        \left[
            d_{\mathrm{perm}}(\hat\theta_n,\theta)
        \right],
        \mathbb E_{\tilde\theta}
        \left[
            d_{\mathrm{perm}}(\hat\theta_n,\tilde\theta)
        \right]
    \right\}
    \geq
    \frac{\Delta}{2}
\]
for every $n$.

Hence the estimation error cannot converge to zero uniformly over a
general POMDP class that permits identical columns of the one-step
observation matrix.  The obstruction is therefore genuinely
information-theoretic: even infinitely many trajectories do not contain
enough information to determine the latent-state realization up to
permutation.
\end{proof}

\section{}\label{apx: example for exponential lower bound}
We prove the proposition by constructing a family of POMDPs satisfying
Assumptions~\ref{ass: full-rank transition kernels}, \ref{ass:multi_step_revelation}, \ref{ass: L-step policy coverage}, \ref{ass: L-step ergodicity}, \ref{ass:distinct-observation-columns}, for which kernel estimation
requires exponentially many observations in $|\mathcal{S}|$.

Let $|\mathcal{S}|=2r$ and
$\mathcal S=\{A_1,\ldots,A_r,B_1,\ldots,B_r\}$.
Take $\Omega=\{o_0,o_1,b_1,\ldots,b_r\}$ and define
\[
p_j=\frac14+\frac{j}{4(r+1)},\qquad
O_{\varphi^*}(o_1\mid A_j)=p_j,\qquad
O_{\varphi^*}(o_0\mid A_j)=1-p_j,
\]
\[
O_{\varphi^*}(b_i\mid B_j)=\mathbf 1\{i=j\},
\]
with all other emission probabilities zero.

Let
\[
Q=
\begin{pmatrix}
0 & r^{-1}\mathbf 1\mathbf 1^\top\\
I_r & 0
\end{pmatrix},
\qquad
\Pi=\frac{1}{|\mathcal{S}|}\mathbf 1\mathbf 1^\top.
\]
Let $v_A\in\mathbb R^r$ be the unit vector with entries
\[
(v_A)_j
=
\frac{(-1)^{j-1}\binom{r-1}{j-1}}
{\sqrt{\binom{2r-2}{r-1}}},
\qquad j=1,\ldots,r,
\]
and set
\[
v=(v_A^\top,0_r^\top)^\top,\qquad
u=\frac{e_{B_1}-e_{B_2}}{\sqrt2},
\qquad
h=\frac1{128r},\qquad \zeta=2^{-r}.
\]
For $\nu\in\{-1,+1\}$, define $\theta_\nu=(\beta_\nu,\varphi^*)$
by assigning, for every $a\in\mathcal A$,
\[
T_{\beta_\nu}^a
=
(1-\zeta)
\left(
\frac34 I_{|\mathcal{S}|}+\frac18 Q+\frac18\Pi+\nu h\,uv^\top
\right)
+\zeta Q^\top.
\]

We verify the five assumptions.

\emph{Assumption~\ref{ass: full-rank transition kernels}.}
Direct calculation gives, uniformly over $a\in\mathcal A$ and
$\nu\in\{-1,+1\}$,
\[
\sigma_{\min}(T_{\beta_\nu}^a)\ge\frac14,
\qquad
\min_{i,j}T_{\beta_\nu}^a(i,j)\ge\frac1{32S}.
\]
Thus Assumption~\ref{ass: full-rank transition kernels} holds with $\sigma_T=1/4$.

\emph{Assumption~\ref{ass:multi_step_revelation}.}
Take $m=2$. The rows corresponding to observation pairs
$(o_0,b_i)$ and $(b_i,b_i)$ form a block-diagonal submatrix
of $\mathcal M_{\theta_\nu}^{(2)}[w]$, with each diagonal block
uniformly well-conditioned. In particular,
\[
\sigma_{\min}
\left(\mathcal M_{\theta_\nu}^{(2)}[w]\right)
\ge\frac1{32},
\qquad \forall w\in\mathcal A.
\]
Hence Assumption~\ref{ass:multi_step_revelation} holds with $\sigma_M=1/32$ and
$\mathcal W=\mathcal A$.

\emph{Assumption~\ref{ass: L-step policy coverage}.}
Let $\pi$ be uniform over
$\mathcal A\oplus\mathcal A\oplus\mathcal W\oplus\mathcal A
=\mathcal A^4$.
Then, for every $a_1,a_2,w$,
\[
\sum_{a'\in\mathcal A}
\pi(a_1\oplus a_2\oplus w\oplus a')
=\frac1{|\mathcal A|^3}.
\]
Therefore Assumption~\ref{ass: L-step policy coverage} holds with
$\pi_{\min}^{\mathrm{blk}}=|\mathcal A|^{-3}$.

\emph{Assumption~\ref{ass: L-step ergodicity}.}
Since $T_{\beta_\nu}^a$ is doubly stochastic for every action,
the induced block-level transition kernel is
$(T_{\beta_\nu}^a)^4$, with stationary distribution
\[
d_{\mathrm{blk}}^\pi=\frac{1}{|\mathcal{S}|}\mathbf 1,
\qquad
(d_{\mathrm{blk}}^\pi)_{\min}=\frac{1}{|\mathcal{S}|}.
\]
Moreover, the uniform lower bound
$T_{\beta_\nu}^a(i,j)\ge 1/(32|\mathcal{S}|)$ implies a dimension-independent
Doeblin contraction. Thus the block-level chain is uniformly
geometrically ergodic, verifying Assumption~\ref{ass: L-step ergodicity}.

\emph{Assumption~\ref{ass:distinct-observation-columns}.}
The observation columns are distinct, with
\[
\Delta_O
=
\min_{s\ne s'}
\left\|
O_{\varphi^*}(\cdot\mid s)
-
O_{\varphi^*}(\cdot\mid s')
\right\|_2
=
\frac{\sqrt2}{4(r+1)}
=\Omega(|\mathcal{S}|^{-1})>0.
\]
Hence Assumption~\ref{ass:distinct-observation-columns} holds.

Finally, the vector $v_A$ satisfies
\[
\sum_{j=1}^r(v_A)_j p_j^k=0,
\qquad k=0,\ldots,r-2.
\]
Consequently, under the kernels obtained by setting $\zeta=0$,
the two observable-trajectory laws agree until a run of $r-1$ consecutive
observations from $\{o_0,o_1\}$ occurs. Such a run has probability
at most $\frac n2(13/16)^{r-2}$ within $n$ data points.
Re-setting $\zeta=2^{-r}$ changes each observable-trajectory law by at most
$n\zeta$ in total variation. Therefore,
\[
\operatorname{TV}
\left(
\mathbb P_{\theta_+}^{n},
\mathbb P_{\theta_-}^{n}
\right)
\le
\frac n2\left(\frac{13}{16}\right)^{r-2}
+2n2^{-r}
\le
n\left(\frac{13}{16}\right)^{r-2}.
\]
However, for every $a\in\mathcal A$,
\[
\left\|
T_{\beta_+}^a-T_{\beta_-}^a
\right\|_{\mathrm F}
=
2(1-\zeta)h
=\Theta(|\mathcal{S}|^{-1}).
\]
The distinct observation columns also prevent this separation
from being eliminated by a state permutation.

Some calculation gives
\[
d_{\mathrm{ker}}(\theta_+,\theta_-)
=2(1-\zeta)h\ge\frac1{64|\mathcal{S}|},
\]
Le Cam's two-point method (see, e.g.,
\cite[Section~5.2, Lemma 5.2.2]{duchi2018introductory}) yields
\[
\inf_{\widehat\theta}
\max_{\nu\in\{-1,+1\}}
\Pr_\nu\left(
d_{\mathrm{ker}}(\widehat\theta,\theta_\nu)
\ge\frac1{256|\mathcal{S}|}
\right)
\ge\frac38
\]
whenever
$n\le\frac14(16/13)^{|\mathcal{S}|/2-2}$.
Therefore, kernel estimation to accuracy $c/|\mathcal{S}|$ for a
sufficiently small universal constant $c>0$ requires
\[
n=\Omega\left(\left(\frac{16}{13}\right)^{|\mathcal{S}|/2}\right)
=\exp(\Omega(|\mathcal{S}|)).
\]

Thus all five assumptions hold, yet polynomial sample complexity
is impossible uniformly over this family.

\section{}\label{apx: quantitative backward sufficient and necessary}
Throughout the proof, the polynomial dependence asserted in
Proposition~\ref{prop:backward-excitation-complexity} is understood with all relevant model parameters used in the assumptions mentioned in the Proposition bounded polynomially in
$|\mathcal S|$. This includes everything in \eqref{equ:relevent parameters_overcomplete}. The polynomial exponents
are independent of $|\mathcal{S}|$.

Write $H=m+2$, $d=d_{\mathrm{blk}}^\pi$, and let
\[
    P_\pi
    :=
    \sum_{\boldsymbol a}\pi(\boldsymbol a)
    T_{\beta^*}^{a_H}\cdots T_{\beta^*}^{a_1}
\]
be the transition matrix of the latent chain at block boundaries.
Let $X_j=s_{1+jH}$, let $\mathcal F_j$ contain the observed records
of the first $j$ completed blocks, and recall that we define
\[
    \bar{b}_{j,\theta^*}^{\mathrm{blk}}:=\mathbb{P}_{\theta^*}(X_j=\cdot\mid\mathcal F_j),
    \qquad
    \bar{B}^{\mathrm{blk}}:=
    \begin{bmatrix}
        (\bar{b}_{1,\theta^*}^{\mathrm{blk}})^\top\\
        \vdots\\
        (\bar{b}_{N,\theta^*}^{\mathrm{blk}})^\top
    \end{bmatrix}.
\]

\paragraph{Extension to a general $L$.} 
For simplicity, we present the argument when Assumption~\ref{ass:backward-excitation} holds with \(L=H\).

For \(L\in H\mathbb N\), regard \(L/H\) consecutive original blocks as a single length-\(L\) analysis window, whose action-string distribution is \(\pi_L=\pi^{\otimes(L/H)}\). Its latent transition matrix is \(P_\pi^{L/H}\), so its stationary boundary distribution remains \(d\). In the stationary argument below, replace the current block’s record by the last \(L\) steps of observation and action history and consider \(j\ge L/H\). The same Bayes identity and conditional Jensen argument then give \(\mathbb E[\bar{b}_{j,\theta^*}^{\mathrm{blk}}(\bar{b}_{j,\theta^*}^{\mathrm{blk}})^\top]\succeq\eta^2I\). For concentration, retain this length-\(L\) window after \(r_\eta\) intervening original blocks and set \(h=r_\eta+L/H\). Conditional on \(\mathcal F_{j-h}\), the starting distribution of the retained window is \(P_\pi^{r_\eta}\bar{b}_{j-h,\theta^*}^{\mathrm{blk}}\). The same perturbation bound and conditional Jensen argument therefore yield
\[
\mathbb E[\bar{b}_{j,\theta^*}^{\mathrm{blk}}(\bar{b}_{j,\theta^*}^{\mathrm{blk}})^\top\mid\mathcal F_{j-h}]
\succeq \frac{\eta^2}{4}I,
\qquad j\ge h.
\]
The subsequent concentration argument applies with this value of \(h\), omitting the first \(h-1\) positive-semidefinite summands. Since \(L\) is polynomially bounded in \(|\mathcal{S}|\), this modification preserves polynomial sample complexity. The grouping is only for the analysis: the sampling policy and the rows of \(\bar B^{\mathrm{blk}}\) remain unchanged.

For $L$ not divisible by $H$, interpret $\Phi_L^\pi(x)$ as
$\mathbb E_d|x^\top c_L|$, where $c_L$ is the belief about a
fixed terminal block-boundary state given the preceding $L$
primitive-step records under the stationary block process.
Set $L_+=H\lceil L/H\rceil$ and extend this window backward.
The corresponding history sigma-fields satisfy
$\mathcal G_L\subseteq\mathcal G_{L_+}$, so the tower property
gives $c_L=\mathbb E_d[c_{L_+}\mid\mathcal G_L]$.
Conditional Jensen's inequality therefore yields
\[
    \Phi_{L_+}^\pi(x)
    =\mathbb E_d|x^\top c_{L_+}|
    \ge \mathbb E_d|x^\top c_L|
    =\Phi_L^\pi(x)
    \ge\eta\|x\|_2.
\]
Thus we may assume $L$ is divisible by $H$ without weakening the excitation
constant or changing the behavior policy. Since
$L\le L_+<L+H$, this rounding preserves polynomial horizon
bounds whenever $L$ and $H$ are polynomially bounded in $|\mathcal{S}|$.

In the rest of this appendix, we assume $L=H$.

We first show the first half (sufficiency) of Proposition~\ref{prop:backward-excitation-complexity}: Suppose Assumptions~\ref{ass: full-rank transition kernels}, \ref{ass:multi_step_revelation}, \ref{ass: L-step policy coverage}, \ref{ass: L-step ergodicity}, \ref{ass:distinct-observation-columns}, and \ref{ass:backward-excitation} hold. Then the kernels can be estimated,
up to a common permutation of the latent states,
to accuracy $\varepsilon$ with probability at least
$1-\delta$ using a trajectory of length
$\operatorname{poly}(|\mathcal{S}|,\varepsilon^{-1},\log(1/\delta))$.

\paragraph{Sufficiency.}
Under stationary initialization, the belief $c$ at the end of a
block (also the start of the next block), conditional only on that block's action and observation
words, satisfies
\[
    c
    =
    \frac{K_{\boldsymbol a,y}d}
         {\mathbf 1^\top K_{\boldsymbol a,y}d}
\]
on every event of positive probability. Consequently,
\[
    \mathbb E|x^\top c|
    =
    \Phi_H^\pi(x).
\]
Since the full-history belief uses at least this information,
conditional Jensen's inequality and Assumption~\ref{ass:backward-excitation} give
\[
    \mathbb E[\bar{b}_{j,\theta^*}^{\mathrm{blk}}(\bar{b}_{j,\theta^*}^{\mathrm{blk}})^\top]
    \succeq
    \mathbb E[cc^\top]
    \succeq
    \eta^2 I.
\]

To justify the above, let \(\mathcal G_j\subseteq\mathcal F_j\) denote
the sigma-fields generated by the current block's action-observation string and the full history, respectively. Since the blockwise action string is sampled independently of the block-start state, which has
stationary distribution \(d\), Bayes' rule gives
\[
c=\mathbb E[e_{S_{\mathrm{end}}}\mid\mathcal G_j]
=\frac{K_{\boldsymbol a,y}d}
{\mathbf 1^\top K_{\boldsymbol a,y}d}.
\]
Moreover, writing \(b_j=\mathbb E[e_{S_{\mathrm{end}}}\mid\mathcal F_j]\),
the tower property yields \(\mathbb E[b_j\mid\mathcal G_j]=c\).
Thus, for any \(x\in\mathbb R^{|\mathcal S|}\), conditional Jensen's
inequality and Assumption~\ref{ass:backward-excitation} imply
\[
\begin{aligned}
\mathbb E[(x^\top b_j)^2]
&\ge \mathbb E[(x^\top c)^2]
\ge \bigl(\mathbb E|x^\top c|\bigr)^2\\
&= \bigl(\Phi_H^\pi(x)\bigr)^2
\ge \eta^2\|x\|_2^2,
\end{aligned}
\]
where the equality follows by averaging over the block's
action-observation words and canceling the posterior normalizer
against their probabilities.

To obtain a high-probability bound from an arbitrary initial
distribution, define
\[
    r_\eta
    :=
    \min\left\{
        r\ge0:
        \sup_{\mu\in\Delta_{|\mathcal{S}|}}
        \|P_\pi^r\mu-d\|_1
        \le \frac{\eta}{2}
    \right\},
    \qquad
    h:=r_\eta+1.
\]
Assumption~\ref{ass: L-step ergodicity} controls $r_\eta$ 
through $\log(G_{\mathrm{blk}}(\pi))$ and $\frac{1}{-\log(\rho_{\mathrm{blk}}(\pi))}$:
\[r_{\eta}\leq \max\left\{0,\left\lceil\frac{\log(2G_{\mathrm{blk}}(\pi)/\eta)}{-\log(\rho_{\mathrm{blk}}(\pi))}\right\rceil\right\},\]
which means
\[h=O\left(1+\left\lceil\frac{\log(G_{\mathrm{blk}}(\pi)/\eta)}{-\log(\rho_{\mathrm{blk}}(\pi))}\right\rceil\right).\]

Let $\Phi_H^{\pi,\mu}$ denote the same functional as
$\Phi_H^\pi$, with $d$ replaced by $\mu$.
The nonnegativity of the block operators and
$\mathbf 1^\top P_\pi=\mathbf 1^\top$ imply
\[
    \left|
        \Phi_H^{\pi,\mu}(x)-\Phi_H^\pi(x)
    \right|
    \le
    \|\mu-d\|_1\|x\|_2.
\]
For $j\ge h$, condition on $\mathcal F_{j-h}$ and retain only
the final block's record in the subsequent history.
After integrating out the intervening $r_\eta$ blocks, the
starting distribution of that final block is
\[
    \mu_j=P_\pi^{r_\eta}b_{j-h}.
\]
Thus $\|\mu_j-d\|_1\le\eta/2$, and conditional Jensen gives
\[
    \mathbb E\!\left[
        b_jb_j^\top\mid\mathcal F_{j-h}
    \right]
    \succeq
    \frac{\eta^2}{4}I.
\]

For each \(r\in\{1,\ldots,h\}\), consider the residue class
\[
J_r=\{r+h,r+2h,\ldots\}\cap\{1,\ldots,N\},
\qquad n_r:=|J_r|\ge \frac Nh-2.
\]
The subsequence \(\{\bar{b}_{j,\theta^*}^{\mathrm{blk}}(\bar{b}_{j,\theta^*}^{\mathrm{blk}})^\top\}_{j\in J_r}\) is adapted to the
filtration \(\{\mathcal F_j\}_{j\in J_r}\), and its predictable
sum satisfies
\[
\sum_{j\in J_r}\mathbb E[\bar{b}_{j,\theta^*}^{\mathrm{blk}}(\bar{b}_{j,\theta^*}^{\mathrm{blk}})^\top\mid\mathcal F_{j-h}]
\succeq \frac{n_r\eta^2}{4}I.
\]
Thus, Because $0\preceq \bar{b}_{j,\theta^*}^{\mathrm{blk}}(\bar{b}_{j,\theta^*}^{\mathrm{blk}})^\top\preceq I$, we can apply the adapted matrix Chernoff inequality
\cite[Theorem~3.1]{tropp2011user}, with
\(R=1\) and further choose \(\delta=1/2\), which gives
\[
\mathbb{P}\left(
\lambda_{\min}\left(\sum_{j\in J_r}\bar{b}_{j,\theta^*}^{\mathrm{blk}}(\bar{b}_{j,\theta^*}^{\mathrm{blk}})^\top\right)
<\frac{n_r\eta^2}{8}
\right)
\le |\mathcal{S}|\exp\left(-\frac{n_r\eta^2}{32}\right).
\]
If
\[
N\ge h\left(2+\frac{32}{\eta^2}
\log\frac{|\mathcal{S}|h}{\delta}\right),
\]
then the failure probability for each residue class is
at most \(\delta/h\). Hence, by a union bound, with
probability at least \(1-\delta\),
\[
\begin{aligned}
\lambda_{\min}((\bar{B}^{\mathrm{blk}})^\top \bar{B}^{\mathrm{blk}})
&=\lambda_{\min}\left(\sum_{j=1}^N b_jb_j^\top\right)\\
&\ge\sum_{r=1}^h
\lambda_{\min}\left(\sum_{j\in J_r}b_jb_j^\top\right)\\
&\ge\frac{\eta^2}{8}\sum_{r=1}^h n_r
=\frac{(N-h)\eta^2}{8}
\ge\frac{N\eta^2}{16},
\end{aligned}
\]
where the last inequality uses \(N\ge 2h\).
Consequently,
\[
\sigma_{\min}\left(\frac{B_N}{\sqrt N}\right)
\ge\frac{\eta}{4},
\]
as claimed.

Hence inverse-polynomial $\eta$ and polynomially controlled
block mixing give inverse-polynomial belief conditioning
after polynomially many blocks.

This completes the proof of Proposition~\ref{prop: predictive belief full-rankedness_overcomplete}, which, together with Assumption~\ref{ass: L-step policy coverage}, yields Lemma~\ref{lem: sigma B_a proof_overcomplete}. These results, in turn, establish Theorem~\ref{thm: approximate permutation for overcomplete POMDPs}, providing a kernel recovery guarantee with polynomial sample complexity and thereby proving sufficiency.

Now, we show the second half of Proposition~\ref{prop:backward-excitation-complexity} (necessity): without
Assumption~\ref{ass:backward-excitation}, the remaining assumptions admit
instances requiring $\exp(\Omega(|\mathcal{S}|))$ observations
for any estimator, even at accuracy
$\varepsilon=\Theta(|\mathcal{S}|^{-1})$ and constant confidence.

\paragraph{Necessity for general kernel recovery.}
Proposition~\ref{prop:exponential-lower-bound} supplies two model families satisfying all
the remaining assumptions, with Assumption~\ref{ass:multi_step_revelation} satisfied with 
$m=2$. By Lemma~\ref{lem:hard-instance-backward-excitation},
their backward excitation is exponentially small in
$|\mathcal S|$ uniformly over all horizons. Thus both
families violate Assumption~\ref{ass:backward-excitation} even when $L$ may be any
polynomial in $|\mathcal{S}|$. Their kernels are separated by order
$|\mathcal{S}|^{-1}$ modulo permutation, while their observation laws
remain statistically indistinguishable for exponentially
many observations. The resulting testing lower bound applies
to every algorithm and estimator and therefore rules out a uniform
polynomial kernel-recovery guarantee under the remaining
assumptions alone.

\begin{lemma}[Failure of backward excitation at every horizon]
\label{lem:hard-instance-backward-excitation}
Let $|\mathcal{S}|=2r$, with $r\ge3$, and consider the two models
$\theta_\nu$, $\nu\in\{-1,+1\}$, constructed in Appendix~\ref{apx: example for exponential lower bound}. Write $\Phi_L^{\pi,\nu}$ for the
backward-excitation functional under $\theta_\nu$, initialized
at $d=|\mathcal{S}|^{-1}\mathbf 1$. For the unit vector $v$ in the
construction, every $L\ge1$ satisfies
\[
    \Phi_L^{\pi,\nu}(v)
    \le
    \frac12\left(\frac{13}{16}\right)^{r-2}
    +(r-1)2^{-r}.
\]
Consequently,
\[
    \sup_{L\ge1}\;
    \inf_{\|x\|_2=1}\Phi_L^{\pi,\nu}(x)
    \le
    \frac12\left(\frac{13}{16}\right)^{|\mathcal{S}|/2-2}
    +\left(\frac{|\mathcal{S}|}{2}-1\right)2^{-|\mathcal{S}|/2}
    =\exp(-\Omega(|\mathcal{S}|)).
\]
Thus both model families violate Assumption~\ref{ass:backward-excitation} for any $L$.
\end{lemma}

\begin{proof}
Suppress the action, since every action has the same
transition kernel, and write
\[
    T_\nu^0
    :=
    \frac34I_{2r}
    +\frac18Q+\frac18\Pi+\nu h uv^\top,
    \qquad
    T_\nu=(1-\zeta)T_\nu^0+\zeta Q^\top,
    \qquad
    \zeta=2^{-r}.
\]
Both $T_\nu^0$ and $Q^\top$ are doubly stochastic, so
$d=|\mathcal{S}|^{-1}\mathbf1$ is stationary. Moreover, the prescribed
open-loop action records are independent of the latent and
observation processes and may be omitted from the conditioning.
Let $Y_t$ denote the observation emitted from $s_t$, and set
\[
    c_L
    :=
    \mathbb P_\nu(s_{L+1}=\cdot\mid Y_{1:L}),
    \qquad s_1\sim d.
\]
Then
\[
    \Phi_L^{\pi,\nu}(v)
    =
    \mathbb E_\nu|v^\top c_L|.
\]

\paragraph{Polynomial annihilation and resets.}
Since $p_j$ is affine in $j$, the definition of $v_A$ and
the finite-difference identity give
\[
    \sum_{j=1}^r(v_A)_j f(p_j)=0
    \qquad
    \text{for every polynomial $f$ of degree at most $r-2$}.
\]
Let $\mathcal S_A=\{A_1,\ldots,A_r\}$ and
$\mathcal S_B=\{B_1,\ldots,B_r\}$. The corresponding blocks
of the unperturbed transition matrix are
\[
    (T_\nu^0)_{\mathcal S_A,\mathcal S_A}
    =
    \frac34I_r+\frac1{16r}\mathbf1\mathbf1^\top,
    \qquad
    (T_\nu^0)_{\mathcal S_A,\mathcal S_B}
    =
    \frac3{16r}\mathbf1\mathbf1^\top.
\]
In particular, observing $b_i$ identifies the current state
as $B_i$. If the following transition uses $T_\nu^0$, the
$\mathcal S_A$-component of the next predictive belief is
the constant vector $(3/(16r))\mathbf1$.

Starting from this constant vector, suppose that the next
$k$ observations belong to $\{o_0,o_1\}$ and that all their
following transitions use $T_\nu^0$. The resulting
$\mathcal S_A$-component has coordinates
$f_k(p_1),\ldots,f_k(p_r)$ for a polynomial $f_k$ of degree
at most $k$. Indeed, an observation multiplies coordinate
$j$ by either $p_j$ or $1-p_j$, while the subsequent
$\mathcal S_A,\mathcal S_A$ transition multiplies the vector
by $3/4$ and adds a constant vector. Normalization only
introduces a common scalar. Hence its projection onto $v_A$
vanishes whenever $k\le r-2$. The same argument applies
starting from $d$, whose $\mathcal S_A$-component is constant.

\paragraph{Accounting for the rare transitions.}
Realize each transition using an independent indicator
$C_t\sim\operatorname{Bernoulli}(\zeta)$: use $T_\nu^0$
when $C_t=0$ and $Q^\top$ when $C_t=1$. Define the more
informed belief
\[
    \widehat c_L
    :=
    \mathbb P_\nu
    (s_{L+1}=\cdot\mid Y_{1:L},C_{1:L}).
\]
The tower property and conditional Jensen's inequality imply
\[
    c_L
    =
    \mathbb E_\nu[\widehat c_L\mid Y_{1:L}],
    \qquad
    \Phi_L^{\pi,\nu}(v)
    \le
    \mathbb E_\nu|v^\top\widehat c_L|.
\]

First suppose $L\ge R:=r-1$, and define
\[
    E_L
    :=
    \left\{
        C_t=1\text{ for some }L-R+1\le t\le L
    \right\}
    \cup
    \left\{
        Y_{L-R+1:L}\in\{o_0,o_1\}^{R}
    \right\}.
\]
On $E_L^c$, the last $R$ observations contain a symbol $b_i$,
and all the last $R$ transitions use $T_\nu^0$.
After the most recent such $b_i$, at most $R-1=r-2$
observations remain. The reset and polynomial-annihilation
argument therefore gives
\[
    v^\top\widehat c_L=0
    \qquad\text{on }E_L^c.
\]
Since $|v^\top\widehat c_L|\le1$, it follows that
\[
    \Phi_L^{\pi,\nu}(v)
    \le
    \mathbb P_\nu(E_L).
\]

The probability of a rare transition among the last $R$
transitions is at most $R\zeta$. Also,
$Y_t\in\{o_0,o_1\}$ holds exactly when $s_t\in\mathcal S_A$.
For every $s\in\mathcal S_A$,
\[
    \mathbb P_\nu(s_{t+1}\in\mathcal S_A\mid s_t=s)
    =
    (1-\zeta)\frac{13}{16}.
\]
Since $d(\mathcal S_A)=1/2$, stationarity gives
\[
    \mathbb P_\nu
    \bigl(Y_{L-R+1:L}\in\{o_0,o_1\}^{R}\bigr)
    =
    \frac12
    \left((1-\zeta)\frac{13}{16}\right)^{R-1}.
\]
Consequently,
\[
    \Phi_L^{\pi,\nu}(v)
    \le
    (r-1)\zeta
    +\frac12\left(\frac{13}{16}\right)^{r-2},
    \qquad L\ge r-1.
\]

If $L<r-1$, then on $\{C_1=\cdots=C_L=0\}$ the
polynomial-annihilation argument applies either from the
initial distribution or from the most recent $b_i$
observation. The relevant polynomial has degree at most
$L\le r-2$, so $v^\top\widehat c_L=0$. Therefore,
\[
    \Phi_L^{\pi,\nu}(v)
    \le
    \mathbb P_\nu(\exists t\le L:C_t=1)
    \le L\zeta
    \le(r-1)\zeta.
\]
Combining the two cases and substituting $\zeta=2^{-r}$
proves the asserted bound, uniformly over $L$ and $\nu$.

Finally, for every fixed $c>0$ independent of $|\mathcal{S}|$, the bound
is smaller than $c|\mathcal S|^{-C}$ for all sufficiently large even $|\mathcal{S}|$.
Since $\|v\|_2=1$, no horizon $L$ can then satisfy
$\Phi_L^{\pi,\nu}(x)\ge c|\mathcal S|^{-C}\|x\|_2$ for every $x$.
This rules out Assumption~\ref{ass:backward-excitation} even with an existential
polynomial-length horizon.
\end{proof}

Lemma~\ref{lem:hard-instance-backward-excitation} establishes the necessity of
quantitative backward excitation for polynomial kernel
recovery over general overcomplete POMDPs that satisfy Assumptions~\ref{ass: full-rank transition kernels}, \ref{ass:multi_step_revelation}, \ref{ass: L-step policy coverage}, \ref{ass: L-step ergodicity}, and \ref{ass:distinct-observation-columns}. For completeness, we further establish that backward
excitation over some polynomial-length history is necessary for polynomial conditioning
of the empirical predictive-belief matrix underlying our recovery
guarantee, which means Assumption~\ref{ass:backward-excitation} is essential to establishing Lemma~\ref{lem: sigma B_a proof_overcomplete} with polynomial complexity.

\paragraph{Existential necessity for polynomial belief conditioning.} Work under stationary initialization. For each \(\ell\in H\mathbb N\), let \(c_\ell\) denote the belief at the terminal boundary conditional on the most recent length-\(\ell\) window under the original behavior policy, and define
\[
M_\ell:=\mathbb E_d[c_\ell c_\ell^\top].
\]
By definition,
\[
\Phi_\ell^\pi(x)=\mathbb E_d|x^\top c_\ell|.
\]
Suppose that, for some polynomially bounded \(N\), \(\gamma>0\), and \(\delta<1\),
\[
\mathbb{P}_d\!\left[
\sigma_{\min}\!\left(
\frac{\bar B^{\mathrm{blk}}}{\sqrt N}
\right)\ge\gamma
\right]\ge1-\delta.
\]
Set \(L=NH\). For every \(j\le N\), stationarity identifies the distribution of the terminal belief based on the last \(jH\) steps with that of \(\bar b^{\mathrm{blk}}_{j,\theta_\star}\). Since the full length-\(L\) history contains these steps, conditional Jensen gives \(M_{jH}\preceq M_L\). Consequently,
\[
M_L
\succeq
\frac1N\sum_{j=1}^N M_{jH}
=
\mathbb E_d\!\left[
\frac{(\bar B^{\mathrm{blk}})^\top
      \bar B^{\mathrm{blk}}}{N}
\right]
\succeq
(1-\delta)\gamma^2I.
\]
For every unit vector \(x\), the bound \(|x^\top c_L|\le1\) implies
\[
\Phi_L^\pi(x)
=
\mathbb E_d|x^\top c_L|
\ge
\mathbb E_d[(x^\top c_L)^2]
=
x^\top M_Lx
\ge
(1-\delta)\gamma^2.
\]
By homogeneity,
\[
\Phi_L^\pi(x)
\ge
(1-\delta)\gamma^2\|x\|_2,
\qquad x\in\mathbb R^S.
\]
Thus, if \(\gamma^{-1}\) is polynomially bounded in \(|\mathcal{S}|\), and \(1-\delta\) is bounded below by a positive constant, quantitative backward excitation holds at some polynomially bounded horizon \(L\). Together with the sufficiency argument, this establishes the equivalence between the polynomial minimum singular value of $\bar{B}^{\mathrm{blk}}$ and Assumption~\ref{ass:backward-excitation}.

\section{}\label{apx: justification for overcomplete predictive belief full-rankedness}
We first state the following proposition which is an overcomplete counterpart to Lemma~\ref{lem: sigma B_a proof}.

\begin{proposition}\label{prop: predictive belief full-rankedness_overcomplete}
Suppose Assumption~\ref{ass:backward-excitation} holds for the blockwise open-loop policy $\pi$,
with parameters $L$ and $\eta$. Fix a confidence level
$\delta\in(0,1)$. If $n$ is bounded below by a universal polynomial
with nonnegative powers in
\[
|\mathcal S|,\quad
\log(1/\delta),\quad
\log(G_{\mathrm{blk}}(\pi)),\quad
\frac{1}{-\log(\rho_{\mathrm{blk}}(\pi))}, \quad L,\quad
1/\eta,
\]
then, with probability at least $1-\delta$,
\[
\sigma_{\min}^2(\bar B^{\mathrm{blk}})
\ge \frac{\eta^2}{16L}n.
\]
\end{proposition}
\begin{proof}
    It is established in the \textbf{Sufficiency} part of Appendix~\ref{apx: quantitative backward sufficient and necessary}.
\end{proof}

Then, by combining Proposition~\ref{prop: predictive belief full-rankedness_overcomplete} and Assumption~\ref{ass: L-step policy coverage} and applying the same argument as in Lemma~\ref{lem: action-restricted predictive belief full-rankedness_formal}, we obtain Lemma~\ref{lem: sigma B_a proof_overcomplete}.

\section{}\label{apx: proof of approximate permutation_overcomplete}
We prove Theorem~\ref{thm: approximate permutation for overcomplete POMDPs} using a modified version of the robust Kruskal's theorem, specifically Theorem 5 from \citet{bhaskara2014uniqueness}. For concision and clarity, we track only the dependence on $n$, suppressing the polynomial dependence on $|\mathcal{A}|,\ 
\frac{1}{\pi_{\min}^{\mathrm{blk}}},\ 
\frac{1}{(d_{\min}^{\pi})^{\mathrm{blk}}},\ 
\frac{1}{\sigma_T},\ 
\frac{1}{\sigma_M},\ \Delta_{O},\ |\mathcal S|$,  $|\Omega|$, and $\frac{L}{\eta^{2}}$ (the first nine quantities in \eqref{equ:relevent parameters_overcomplete}).

\paragraph{Robust Kruskal preliminaries.}
Throughout the proof, $\operatorname{K-rank}_{\tau}(\cdot)$ denotes the
robust Kruskal rank of \citet{bhaskara2014uniqueness}: $\operatorname{K-rank}_{\tau}(A)
\ge k$ means that every $k$-column submatrix of $A$ has minimum singular
value at least $1/\tau$. A matrix is $\rho$-bounded if the $\ell_{2}$ norm of every column is upper-bounded by $\rho$.

\begin{theorem}[Dimension-free robust uniqueness of tensor decompositions]\label{thm: dimension-free robust kruskal}
Let \(T=[A\,B\,C]\) be a rank-\(R\) tensor, where
\(A\in\mathbb{R}^{n_A\times R}\),
\(B\in\mathbb{R}^{n_B\times R}\), and
\(C\in\mathbb{R}^{n_C\times R}\).
Suppose \(T\) is
\((\rho_A,\rho_B,\rho_C)\)-bounded, with
\[
\operatorname{K\!-\!rank}_{\tau_A}(A)\ge k_A,\quad
\operatorname{K\!-\!rank}_{\tau_B}(B)\ge k_B,\quad
\operatorname{K\!-\!rank}_{\tau_C}(C)\ge k_C,
\]
satisfying \(k_A+k_B+k_C\ge 2R+2\).

Then there exist a universal constant \(c>0\) and a
polynomial \(\widetilde{\vartheta}_5\), independent of
\(n_A,n_B,n_C\), such that for every \(0<\varepsilon'<1\),
setting
\[
\varepsilon=
\frac{\varepsilon'}
{R^c
\prod_{M\in\{A,B,C\}}
\widetilde{\vartheta}_5(\tau_M,\rho_M,\rho'_M)},
\]
any other \((\rho'_A,\rho'_B,\rho'_C)\)-bounded rank-\(R\)
decomposition \(T'=[A'\,B'\,C']\) satisfying
\(\|T-T'\|_F\le\varepsilon\) admits an \(R\times R\)
permutation matrix \(\Pi\) and diagonal matrices
\(\Lambda_A,\Lambda_B,\Lambda_C\) such that
\[
\|A'-A\Pi\Lambda_A\|_F\le\varepsilon',\qquad
\|B'-B\Pi\Lambda_B\|_F\le\varepsilon',
\]
\[
\|C'-C\Pi\Lambda_C\|_F\le\varepsilon',\qquad
\|\Lambda_A\Lambda_B\Lambda_C-I_R\|_F
\le\varepsilon'.
\]
In particular, the robustness guarantee has no explicit
dependence on the ambient dimensions \(n_A,n_B,n_C\).
\end{theorem}

\begin{proof}[Proof sketch]
We briefly explain how the proof of the original theorem in \citet{bhaskara2014uniqueness} can be adapted to prove our modified theorem.

First, in Lemma 30, instead of drawing the Gaussian
vector in \(\mathbb{R}^d\), draw it in
\(V=\operatorname{span}\{u_1,\ldots,u_t\}\).
Since \(s=\dim(V)\le t\), the same Gaussian
anti-concentration and norm-concentration arguments yield
a unit vector \(w\in V\) satisfying
\[
|\langle u_i,w\rangle|
>\frac{\eta}{20t\sqrt{s}}
\ge\frac{\eta}{20t^{3/2}},
\qquad i\in[t],
\]
whenever \(\|u_i\|_2\ge\eta>0\).
Consequently, in the proof of Lemma 11, we may replace
\(\varepsilon_1=nR\varepsilon\) with
\[
\varepsilon_1=20R^{3/2}\varepsilon,
\]
removing the dependence on \(n\).
In fact, a stronger dimension-free bound
\(|\langle u_i,w\rangle|\ge\eta/t\) follows from a
deterministic sign-maximization argument, which we omit here for brevity.

Second, in Lemma 12, the original proof already establishes
\[
\operatorname{dist}(y,\operatorname{span}(X_A))
\le (2q\tau\rho+1)\varepsilon.
\]
Since \(q\le R\) and, without loss of generality,
\(\tau\ge1\), we obtain
\[
\operatorname{dist}(y,\operatorname{span}(X_A))
\le3R\tau\max\{\rho,1\}\varepsilon.
\]
Thus, we may replace \(\vartheta_{12}=4n\tau\rho\) with
\[
\widetilde{\vartheta}_{12}
=3R\tau\max\{\rho,1\}.
\]

Substituting these two improvements into the induction
proving Lemma 6 yields a robust permutation lemma whose
error and parameters depend polynomially only
on \(R,\tau,\rho\). The factors needed to control the
diagonal rescalings are likewise polynomial in these
parameters, by the robust Kruskal-rank condition.

The remaining tensor-contraction, rank-transfer, and
permutation-alignment arguments proceed as in the
original proof, with polynomial dependence on the
robust conditioning and column-norm parameters.
Combining these bounds gives the stated dimension-free
polynomial \(\widetilde{\vartheta}_5\) and exponent \(c\).

An alternative argument for eliminating the ambient-dimension dependence is
to restrict each mode to the subspace spanned by the columns of the two
corresponding factor matrices. Specifically, for each
\(M\in\{A,B,C\}\), define
\[
U_M=\operatorname{span}\bigl(\operatorname{col}(M)
\cup\operatorname{col}(M')\bigr).
\]
Since both \(M\) and \(M'\) have \(R\) columns, we have
\(\dim(U_M)\le 2R\). Representing both matrices in an orthonormal
basis of \(U_M\), with zero-padding if necessary, preserves their
column norms, singular values, robust Kruskal ranks, and the Frobenius
distance between the corresponding tensor decompositions.
Consequently, we may apply the original Theorem~5 with
\(n_A=n_B=n_C=2R\) and lift the resulting approximation back to the
original ambient spaces. Since the original \(\vartheta_5\) is
polynomial, substituting \(n=2R\) yields a polynomial depending only
on \(R,\tau,\rho,\rho'\), thereby eliminating all dependence on
\(n_A,n_B,n_C\).
\end{proof}

We then proceed to prove Theorem~\ref{thm: approximate permutation for overcomplete POMDPs} using Theorem~\ref{thm: dimension-free robust kruskal}. For simplicity, we prove the pointwise version of the theorem using Lemma~\ref{lem:quadratic bound_block predictive distributions}. The uniform statement displayed in Theorem~\ref{thm: approximate permutation for overcomplete POMDPs} can be shown identically using the uniform counterpart of Lemma~\ref{lem:quadratic bound_block predictive distributions}, which can be proven by applying the same argument from Appendix~\ref{apx: proof for universal quadratic bound for O} to Lemma~\ref{lem:quadratic bound_block predictive distributions}.

\paragraph{1. Fix an action string.}
Fix $a_1,a_2\in\mathcal A$, denote $(a_{1},a_{2})$ as $\tilde{a}$, and fix
\[
w\in\mathcal W^\pi_{\tilde{a}}.
\]
By Assumption~\ref{ass: L-step policy coverage}, the blockwise behavior policy plays
\[
a_1\oplus a_2\oplus w\oplus a'
\]
with probability $\Theta(1)$ after marginalizing over the
terminal action $a'$. Recall the definitions of $\mathcal{I}_{\tilde{a},w}^{\mathrm{blk}}$ and $N_{\tilde{a},w}^{\mathrm{blk}}$. From Assumption~\ref{ass: L-step policy coverage}, we can see that
\[
N_{\tilde{a},w}^{\mathrm{blk}}=\Theta(n)
\]
with high probability. We assume this high-probability event throughout the rest of the proof.

For notational simplicity, denote
\[
B
:=
\bar B_{\tilde{a},w}^{\mathrm{blk}}
\in\mathbb R^{N\times |\mathcal{S}|},\qquad \widehat{B}:=\widehat{B}_{\tilde{a},w}^{\mathrm{blk}}\in\mathbb{R}^{N\times |\mathcal{S}|}, \qquad N_{\tilde{a},w}^{\mathrm{blk}}\coloneqq N, \qquad \mathcal{I}^{\mathrm{blk}}_{\tilde{a},w}\coloneqq \mathcal{I}.
\]

From Proposition~\ref{prop: predictive belief full-rankedness_overcomplete},
\[
\sigma_{\min}(B)=\Theta(\sqrt N)=\Theta(\sqrt{n}).
\]
Thus, for some $\tau_B=\Theta(1)$, where $\tau_{B}$ depends polynomially on two of the first nine quantities in \eqref{equ:relevent parameters_overcomplete} (from Lemma \ref{lem: sigma B_a proof_overcomplete}),
\begin{equation}\label{equ: k rank of B is S}
\operatorname{K-rank}_{\tau_B}\!\left(\frac{B}{\sqrt N}\right)
=
|\mathcal{S}|.
\end{equation}

For notational simplicity, write
\[
M_w:=\mathcal M_{\theta^*}^{(m)}[w],
\qquad
T_1:=T_{\beta^*}^{a_1},
\qquad
T_2:=T_{\beta^*}^{a_2},
\]
and use hats for the corresponding estimated quantities.

Schematically, each block can be illustrated as below:
\begin{equation}\label{equ: trajectory form}
S_0
\longrightarrow O_0
\xrightarrow{a_1}
S_1
\longrightarrow O_1
\xrightarrow{a_2}
S_2
\xrightarrow{w}
\text{$m$-observation block}.
\end{equation}

\paragraph{2. First tensor decomposition.}
Restrict the block indices to 
$i\in\mathcal I$. Every such block has the form illustrated in \eqref{equ: trajectory form}. Let $Z\in\Omega^m$ denote the informative observation block generated
after $a_2$. Thus, for each
$i\in\mathcal I$, the corresponding joint predictive law (i.e. $p_{i,\theta^*}$ on the LHS of the bound in Lemma~\ref{lem:quadratic bound_block predictive distributions}) can be factorized as
\[
\begin{aligned}
p_{i,\theta^*}^{(m+2)}&\coloneqq\mathbb{P}_{\theta^*}
(O_0=o_0,O_1=o_1,Z=z\mid \mathcal H_i,a_1,a_2,w)
\\
&=
\sum_{s_0,s_1,s_2}
\bar b_{i,\theta^*}^{\mathrm{blk}}(s_0)\,
O_{\varphi^*}(o_0\mid s_0)\,
T_{1}(s_1\mid s_0)\,
O_{\varphi^*}(o_1\mid s_1)\,
T_{2}(s_2\mid s_1)\,
[M_w](z,s_2).
\end{aligned}
\]

Define
\[A_{0}\coloneqq M_{2}T_{2}T_{1}.\]

Marginalizing over $O_1$ and using
$\sum_{o_1}O_{\varphi^*}(o_1\mid s_1)=1$ gives
\[
q_{i,\theta^*}(o_0,z)\coloneqq\mathbb{P}_{\theta^*}
(O_0=o_0,Z=z\mid \mathcal H_i,a_1,a_2,w)
=
\sum_{s_0}
\bar b_{i,\theta^*}^{\mathrm{blk}}(s_0)\,
O_{\varphi^*}(o_0\mid s_0)\,
[A_0](z,s_0).
\]

Stacking these marginalized laws over $i\in\mathcal I$ therefore yields
the tensor
\[
\mathscr T_0=[A_0,B,O_{\varphi^*}],
\]
and analogously
\[
\widehat{\mathscr T}_0
=
[\widehat A_0,\widehat B,O_{\hat{\varphi}}].
\]

To control the tensor error, we invoke 
Lemma~\ref{lem:quadratic bound_block predictive distributions}.

By definition, for each $i\in\mathcal I$,
\[
q_{i,\theta^*}(o_0,z)
=
\sum_{o_1\in\Omega}
p_{t_i,\theta^*}(o_0,o_1,z).
\]

Furthermore, by construction, $q_{i,\theta^*}$ and
$q_{i,\hat\theta}$ are precisely the $i$th slices of
$\mathscr T_0$ and $\widehat{\mathscr T}_0$, respectively.

Since marginalization is an $\ell_1$-contraction,
\[
\|q_{t_i,\hat\theta}-q_{i,\theta^*}\|_1
\le
\|p_{t_i,\hat\theta}
-
p_{i,\theta^*}\|_1.
\]
Moreover, reshaping $q_{i,\theta}$ into a tensor slice does not change
its entries, so
\[
\|\mathscr T_0(:,i,:)-\widehat{\mathscr T}_0(:,i,:)\|_F
\le
\|q_{i,\hat\theta}-q_{i,\theta^*}\|_1.
\]
Therefore,
\[
\begin{aligned}
\|\mathscr T_0-\widehat{\mathscr T}_0\|_F^2
&=
\sum_{i\in\mathcal I}
\|\mathscr T_0(:,i,:)-\widehat{\mathscr T}_0(:,i,:)\|_F^2\\
&\le
\sum_{i\in\mathcal I}
\|p_{t_i,\hat\theta}
-
p_{t_i,\theta^*}\|_1^2\\
&\le
\sum_{i=1}^{\lfloor n/(m+2)\rfloor}
\|p_{t_i,\hat\theta}
-
p_{t_i,\theta^*}\|_1^2
\le \frac{8}{\pi_{\min}^{\mathrm{blk}}}
\left(
\ell_n(\theta^*)-\ell_n(\hat\theta)
+2\log(2/\delta)
\right)+4,
\end{aligned}
\]
where the last step follows from 
Lemma~\ref{lem:quadratic bound_block predictive distributions}. Since $N=|\mathcal I|=\Theta(n)$, normalizing the belief
mode ($B$ and $\widehat{B}$) by $\sqrt N$ gives
\begin{equation}\label{equ:varepsilon n bound}
\varepsilon_{n}\coloneqq\|\mathscr T_{0,\mathrm{norm}}-\widehat{\mathscr T}_{0,\mathrm{norm}}\|_F=O\left(\sqrt{\frac{H_n(\delta)}{n}}\right),
\end{equation}
where $\mathscr T_{0,\mathrm{norm}}$ and $\widehat{\mathscr T}_{0,\mathrm{norm}}$ are normalized counterparts of $\mathscr T_{0}$ and $\widehat{\mathscr T}_{0}$ and $\pi_{\min}^{\mathrm{blk}}$ is absorbed into the big $O$ notation.

Note that we define $H_n(\delta)
=
G_n+2d\log(2|\Omega|n)+2\log(4/\delta)+2$. This makes the big-O statement in \eqref{equ:varepsilon n bound} loose. It becomes tight when we replace Lemma~\ref{lem:quadratic bound_block predictive distributions} with its uniform counterpart to prove the uniform statement of Theorem~\ref{thm: approximate permutation for overcomplete POMDPs}.

Notice that, after this normalization, all factor matrices ($A_{0}$, $B$, and $O_{\varphi^*}$) and their estimated counterparts satisfy the uniform column-norm boundedness required by Theorem~\ref{thm: dimension-free robust kruskal}.

Then, note that Assumption~\ref{ass:distinct-observation-columns} implies that every two-column submatrix of $O_{\varphi^*}$ has minimum
singular value bounded below by a positive constant. Hence, for some
$\tau_O=\Theta(1)$,
\begin{equation}\label{equ: k rank of O is 2}
\operatorname{K-rank}_{\tau_O}(O_{\varphi^*})\ge 2.
\end{equation}

Furthermore, by Assumptions~\ref{ass: full-rank transition kernels} and \ref{ass:multi_step_revelation},
\[
\sigma_{\min}(A_0)
\ge
\sigma_{\min}(M_w)
\sigma_{\min}(T_2)
\sigma_{\min}(T_1)
=
\Theta(1).
\]
Hence, for some $\tau_{A_0}=\Theta(1)$,
\[
\operatorname{K-rank}_{\tau_{A_0}}(A_0)=|\mathcal{S}|.
\]
Together with \eqref{equ: k rank of O is 2} and \eqref{equ: k rank of B is S},
\[
\operatorname{K-rank}_{\tau_{A_0}}(A_0)
+
\operatorname{K-rank}_{\tau_B}(\frac{B}{\sqrt{N}})
+
\operatorname{K-rank}_{\tau_O}(O_{\varphi^*})
\ge
|\mathcal{S}|+|\mathcal{S}|+2
=
2|\mathcal{S}|+2.
\]

Thus Theorem~\ref{thm: dimension-free robust kruskal} applies. Furthermore, notice that  $\tau_{A_{0}}$, $\tau_{B}$, and $\tau_{O}$ are $\Theta(1)$ and depend only polynomially on the
fixed model parameters in \eqref{equ:relevent parameters_overcomplete} (namely, the first nine) and not on $n$. Consequently, for $n$ that exceeds a threshold that is polynomial in the first nine quantities in \eqref{equ:relevent parameters_overcomplete}, there exist a
permutation matrix $\Pi_0$ and diagonal matrices
$\Lambda_{A,0}$, $\Lambda_{B,0}$, and $\Lambda_{O,0}$
such that
\begin{equation}\label{equ: A pre scaling alignment}
\|\widehat A_0-A_0\Pi_0\Lambda_{A,0}\|_F
\le C_K\varepsilon_n=O\left(\sqrt{\frac{H_n(\delta)}{n}}\right),
\end{equation}
\[
\left\|
\frac{\widehat B}{\sqrt N}
-\frac{B}{\sqrt N}\Pi_0\Lambda_{B,0}
\right\|_F
\le C_K\varepsilon_n=O\left(\sqrt{\frac{H_n(\delta)}{n}}\right),
\]
and
\begin{equation}\label{equ: O pre scaling alignment}
\|O_{\hat\varphi}-O_{\varphi^*}\Pi_0\Lambda_{O,0}\|_F
\le C_K\varepsilon_n=O\left(\sqrt{\frac{H_n(\delta)}{n}}\right),
\end{equation}
with
\begin{equation}\label{equ: scaling approximately one}
\|\Lambda_{A,0}\Lambda_{B,0}
\Lambda_{O,0}-I\|_F
\le C_K\varepsilon_n=O\left(\sqrt{\frac{H_n(\delta)}{n}}\right),
\end{equation}
where $C_{K}$ depends polynomially on the first nine quantities in \eqref{equ:relevent parameters_overcomplete}.

\paragraph{3. Remove the scaling ambiguity in the first tensor.}
Both $A_0$ and $\widehat A_0$ are column-stochastic, since $M_w$ is an
observation channel and $T_1,T_2$ are column-stochastic transition
matrices. Hence
\[
\mathbf 1^\top A_0
=
\mathbf 1^\top\widehat A_0
=
\mathbf 1^\top.
\]
Left-multiplying the LHS of \eqref{equ: A pre scaling alignment} by $\mathbf 1^\top$ gives
\begin{equation}\label{equ: A scaling}
\Lambda_{A,0}=I+O\left(\sqrt{\frac{H_n(\delta)}{n}}\right).
\end{equation}
Note that left-multiplying by \(\mathbf{1}^\top\) introduces an additional factor of \(|\Omega|^{m/2}\) in the bound. Since \(m\) is fixed, this factor is polynomial in \(|\Omega|\) and can be absorbed into the polynomial constant.

Similarly,
\[
\mathbf 1^\top O_{\varphi^*}
=
\mathbf 1^\top O_{\hat{\varphi}}
=
\mathbf 1^\top
\]
implies from \eqref{equ: O pre scaling alignment} that
\begin{equation}\label{equ: O scaling}
\Lambda_{O,0}=I+O\left(\sqrt{\frac{H_n(\delta)}{n}}\right).
\end{equation}

Combining \eqref{equ: scaling approximately one}, \eqref{equ: A scaling}, and \eqref{equ: O scaling} gives us
\[
\Lambda_{B,0}=I+O\left(\sqrt{\frac{H_n(\delta)}{n}}\right).
\]
Thus
\begin{equation}\label{equ: O with Pi 0}
\widehat A_0
=
A_0\Pi_0+O\left(\sqrt{\frac{H_n(\delta)}{n}}\right),
\qquad
O_{\hat{\varphi}}
=
O_{\varphi^*}\Pi_0+O\left(\sqrt{\frac{H_n(\delta)}{n}}\right),
\end{equation}
and
\[
\|\widehat B-B\Pi_0\|_F
=
O\left(\sqrt{H_{n}(\delta)}\right).
\]

Here and throughout, \(O\left(\sqrt{\frac{H_n(\delta)}{n}}\right)\) in matrix-valued expressions denotes a remainder whose Frobenius norm is $O\left(\sqrt{\frac{H_n(\delta)}{n}}\right)$, and the same convention applies to $O(\sqrt{H_{n}(\delta)})$.

\paragraph{4. Construct a second set of matrices.}
For $o_0\in\Omega$, define
\[
D_{o_0}
:=
\operatorname{diag}(O_{\varphi^*}^\top e_{o_0}),
\qquad
C_{1,o_0}
:=
T_1D_{o_0}B^\top.
\]
Concatenate these matrices over $o_0$:
\[
C_1
:=
\big[
C_{1,o_0}
\big]_{o_0\in\Omega}
=
\big[
T_1D_{o_0}B^\top
\big]_{o_0\in\Omega}.
\]

For every $x\in\mathbb R^{|\mathcal{S}|}$,
\[
\begin{aligned}
\|C_1^\top x\|_2^2
&=
\sum_{o\in\Omega}
\left\|
BD_oT_1^\top x
\right\|_2^2
\\
&\ge
\frac{1}{|\Omega|}
\left\|
B
\left(
\sum_{o\in\Omega}D_o
\right)
T_1^\top x
\right\|_2^2
\\
&=
\frac{1}{|\Omega|}
\|BT_1^\top x\|_2^2,
\end{aligned}
\]
where the inequality is the vector-valued Cauchy--Schwarz inequality and
\[
\sum_{o\in\Omega}D_o=I.
\]
Consequently,
\[
\sigma_{\min}(C_1)
\ge
\frac{
\sigma_{\min}(B)\sigma_{\min}(T_1)
}{
\sqrt{|\Omega|}
}
=
\Theta(\sqrt n).
\]

Therefore, for some $\tau_{C_1}=\Theta(1)$ that only depends polynomially on the first nine quantities in \eqref{equ:relevent parameters_overcomplete} plus $|\Omega|$,
\begin{equation}\label{equ: k rank of C1 is S}
\operatorname{K-rank}_{\tau_{C_1}}
\left(
\frac{C_1^\top}{\sqrt N}
\right)
=
|\mathcal{S}|.
\end{equation}

\paragraph{5. Second tensor decomposition}
Define
\[
A_1:=M_wT_2.
\]

For each $i\in\mathcal I$ and $o_0,o_1\in\Omega$, consider the joint
predictive law of the first two observations and the informative
$m$-observation block $Z$. Define the tensor $\mathscr T_1$ by
\[
[\mathscr T_1]_{z,(i,o_0),o_1}
:=
\mathbb{P}_{\theta^*}
\bigl(
O_0=o_0,\,
O_1=o_1,\,
Z=z
\mid
\mathcal H_i,a_1,a_2,w
\bigr).
\]
Recall that
\[
[C_1]_{:,(i,o_0)}
=
T_1D_{o_0}\bar b_{i,\theta^*}^{\mathrm{blk}},
\]
which is the unnormalized latent-state distribution at $S_1$ after
observing $O_0=o_0$ and applying action $a_1$.

Conditioning on $S_1=s$ therefore gives
\[
[\mathscr T_1]_{z,(i,o_0),o_1}
=
\sum_{s\in\mathcal S}
[A_1]_{z,s}\,
[C_1]_{s,(i,o_0)}\,
O_{\varphi^*}(o_1\mid s).
\]

Hence $\mathscr T_1$ has the rank-$|\mathcal{S}|$ CP decomposition
\[
\mathscr T_1=[A_1,\ C_1^\top,\ O_{\varphi^*}].
\]

The estimated model analogously gives
\[
\widehat{\mathscr T}_1
=
[\widehat A_1,\ \widehat C_1^\top,\ O_{\hat{\varphi}}],
\qquad
\widehat A_1:=\widehat M_w\widehat T_2.
\]

Using Lemma~\ref{lem:quadratic bound_block predictive distributions} again gives
\[
\left\|
\frac{\mathscr T_1}{\sqrt N}
-
\frac{\widehat{\mathscr T}_1}{\sqrt N}
\right\|_F
=O\left(\sqrt{\frac{H_n(\delta)}{n}}\right).
\]

Again, this normalization makes sure that all factor matrices and their estimated counterparts satisfy the uniform column-norm boundedness required by Theorem~\ref{thm: dimension-free robust kruskal}.

Moreover,
\[
\sigma_{\min}(A_1)
\ge
\sigma_{\min}(M_w)
\sigma_{\min}(T_2)
=
\Theta(1),
\]
so, for some $\tau_{A_1}=\Theta(1)$ that depends polynomially on the first five quantities in \eqref{equ:relevent parameters_overcomplete},
\[
\operatorname{K-rank}_{\tau_{A_1}}(A_1)=|\mathcal{S}|.
\]
Together with \eqref{equ: k rank of O is 2} and \eqref{equ: k rank of C1 is S},
\[
\operatorname{K-rank}_{\tau_{A_1}}(A_1)
+
\operatorname{K-rank}_{\tau_{C_1}}
\left(
\frac{C_1^\top}{\sqrt N}
\right)
+
\operatorname{K-rank}_{\tau_O}(O_{\varphi^*})
\ge
2|\mathcal{S}|+2.
\]
Hence Theorem~\ref{thm: dimension-free robust kruskal} applies again: There exist a permutation matrix $\Pi_1$ and diagonal scalings such that,
after removing the scaling ambiguity exactly as in the first tensor decomposition,
\[
\widehat A_1
=
A_1\Pi_1+O\left(\sqrt{\frac{H_n(\delta)}{n}}\right),
\]
\[
\widehat C_1^\top=C_1^\top\Pi_1
+O\left(\sqrt{H_n(\delta)}\right),
\]
and
\begin{equation}\label{equ: O with Pi 1}
O_{\hat{\varphi}}
=
O_{\varphi^*}\Pi_1+O\left(\sqrt{\frac{H_n(\delta)}{n}}\right).
\end{equation}
Indeed, the first and third factors are column-stochastic, so their
diagonal scalings are $I+O\left(\sqrt{\frac{H_n(\delta)}{n}}\right)$, and the product-of-scalings condition in
the theorem then forces the scaling of $C_1^\top$ to be $I+O\left(\sqrt{\frac{H_n(\delta)}{n}}\right)$ as well. Again, note that left-multiplying by \(\mathbf{1}^\top\) introduces an additional factor of \(|\Omega|^{m/2}\) in the upper bound of the first equation from the dimension of $A_{1}$. Since \(m\) is fixed, this factor is polynomial in \(|\Omega|\) and can be absorbed into the polynomial constant.

\paragraph{6. Synchronize the two permutations.}
From \eqref{equ: O with Pi 0} and \eqref{equ: O with Pi 1},
\[
O_{\varphi^*}\Pi_0-O_{\varphi^*}\Pi_1=O\left(\sqrt{\frac{H_n(\delta)}{n}}\right).
\]
If $\Pi_0\neq\Pi_1$, then for some column the left-hand side contains the
difference of two distinct columns of $O_{\varphi^*}$, whose $\ell_\infty$-distance
is at least $\Delta_O>0$. Hence, for sufficiently large $n$ that exceeds a polynomial threshold in terms of the first nine quantities in \eqref{equ:relevent parameters_overcomplete},
\[
\Pi_0=\Pi_1=:\Pi.
\]
Define
\[
P:=\Pi^\top.
\]
Then
\begin{equation}\label{equ: O B P alignment}
O_{\hat{\varphi}} P=O_{\varphi^*}+O\left(\sqrt{\frac{H_n(\delta)}{n}}\right),
\qquad
\widehat BP=B+O\left(\sqrt{H_n(\delta)}\right),
\end{equation}
and
\begin{equation}\label{equ: A1 C1 P alignment}
\widehat A_1P=A_1+O\left(\sqrt{\frac{H_n(\delta)}{n}}\right),
\qquad
\widehat C_1^\top P=C_1^\top+O\left(\sqrt{H_n(\delta)}\right).
\end{equation}

\paragraph{7. Recover the transition kernel.}
Transposing the second relation in \eqref{equ: A1 C1 P alignment} and left-multiplying by $P$,
\begin{equation}\label{equ: P C1 alignment}
\widehat C_1
=
PC_1+O\left(\sqrt{H_n(\delta)}\right).
\end{equation}

For each $o\in\Omega$, the corresponding block of $C_1$ is
\[
C_{1,o}=T_1D_oB^\top,
\]
whereas
\[
\widehat C_{1,o}
=
\widehat T_1\widehat D_o\widehat B^\top.
\]

From \eqref{equ: O B P alignment} we get:
\begin{equation}\label{equ: P B alignment}
\widehat B^\top
=
PB^\top+O\left(\sqrt{H_n(\delta)}\right),
\end{equation}
and
\begin{equation}\label{equ: O P error}
O_{\hat{\varphi}}
=
O_{\varphi^*}P^\top+O\left(\sqrt{\frac{H_n(\delta)}{n}}\right).
\end{equation}

\eqref{equ: O P error} further implies
\begin{equation}\label{equ: P Do P alignment}
\widehat D_o
=
P D_oP^\top+O\left(\sqrt{\frac{H_n(\delta)}{n}}\right).
\end{equation}

Thus, from \eqref{equ: P C1 alignment},
\[
\widehat T_1\widehat D_o\widehat B^\top
=
PT_1D_oB^\top+O\left(\sqrt{H_n(\delta)}\right).
\]
Combining \eqref{equ: P B alignment} and \eqref{equ: P Do P alignment} gives us
\[
\widehat D_o\widehat B^\top
=
PD_oB^\top+O\left(\sqrt{H_n(\delta)}\right).
\]
Therefore,
\[
(\widehat T_1P-PT_1)D_oB^\top
=
O\left(\sqrt{H_n(\delta)}\right).
\]

Summing over $o$ and using $\sum_oD_o=I$,
\[
\left(
\widehat T_1P
-
PT_1
\right)
B^\top
=
O\left(\sqrt{H_n(\delta)}\right).
\]
Since $\sigma_{\min}(B)=\Omega(\sqrt n)$, right-inverting $B^\top$ and substituting back the original notations
yield
\[
 T_{\hat\beta}^{a_1}P
-
P T_{\beta^*}^{a_1}
=
O\left(\sqrt{\frac{H_n(\delta)}{n}}\right).
\]
Equivalently,
\[
P^\top T_{\hat\beta}^{a_1}
P
=
T_{\beta^*}^{a_1}
+
O\left(\sqrt{\frac{H_n(\delta)}{n}}\right).
\]

By Assumption~\ref{ass: L-step policy coverage}, for every $a_{1}$, there exists $a_2$ and $w$ such that $\bar{B}_{\tilde{a},w}^{\mathrm{blk}}$ is $\Theta(n)$, where $\tilde{a}=(a_{1},a_{2})$. Thus every action can be placed in the
$a_1$-position. Repeating the argument for all such pairs recovers every
transition matrix under the same permutation $P$. The permutation is
common across all pairs because every application aligns the same
observation kernel $O_{\varphi^*}$, and the separation $\Delta_O>0$ forces the
permutations to coincide.

Note that the implicit constants in \(O(\sqrt{H_n(\delta)/n})\) depend polynomially on the first nine quantities in \eqref{equ:relevent parameters_overcomplete}. The dependence on the first seventh arises from Theorem~\ref{thm: dimension-free robust kruskal}, while the dependence on the eighth, \(|\Omega|\), comes from converting Frobenius-norm bounds into entrywise \(\ell_1\)-norm bounds.

This complete the proof of Theorem~\ref{thm: approximate permutation for overcomplete POMDPs}.


\newpage
\input{checklist.tex}

\end{document}

%% file: checklist.tex
\section*{NeurIPS Paper Checklist}

\begin{enumerate}

\item {\bf Claims}
    \item[] Question: Do the main claims made in the abstract and introduction accurately reflect the paper's contributions and scope?
    \item[] Answer: \answerYes{} 
    \item[] Justification: The abstract and introduction state our main contribution: a finite-sample, candidate-agnostic translation from likelihood gap to parameter recovery error for both undercomplete and overcomplete POMDPs. This result is established in Section~\ref{subsec: main result-undercomplete} for the undercomplete regime and in Section~\ref{subsec: main result-overcomplete} for the overcomplete regime.

\item {\bf Limitations}
    \item[] Question: Does the paper discuss the limitations of the work performed by the authors?
    \item[] Answer: \answerYes{} 
    \item[] Justification: 
    We discussed the limitations of this work in Section~\ref{sec: future work}. In particular, our current analysis imposes a full-rank assumption on the transition kernels in both the undercomplete and overcomplete regimes. Developing weaker, regime-specific assumptions would lead to a more refined theory. As noted in Remark~\ref{remark: extending the full-rank assumption of T}, the current transition-kernel assumption can be relaxed.

    We also acknowledge in Secion~\ref{sec: future work} and Remark~\ref{remark: block-level policy assumptions} that Assumptions~\ref{ass: L-step policy coverage} and~\ref{ass: L-step ergodicity} are convenient sufficient conditions on the behavior policy in the overcomplete regime, chosen to enable a direct extension of the undercomplete analysis. Characterizing the minimal behavior-policy conditions under which likelihood-gap control still implies parameter recovery in the overcomplete regime remains an important direction for future work.

\item {\bf Theory assumptions and proofs}
    \item[] Question: For each theoretical result, does the paper provide the full set of assumptions and a complete (and correct) proof?
    \item[] Answer: \answerYes{} 
    \item[] Justification: Both Section~\ref{subsec: main result-undercomplete} and Section~\ref{subsec: main result-overcomplete} explicitly introduce the kernel identifiability assumptions, policy coverage assumptions, and the ergodicity assumptions needed for all the results.

    \item {\bf Experimental result reproducibility}
    \item[] Question: Does the paper fully disclose all the information needed to reproduce the main experimental results of the paper to the extent that it affects the main claims and/or conclusions of the paper (regardless of whether the code and data are provided or not)?
    \item[] Answer: \answerNA{} 
    \item[] Justification: The paper is theoretical and does not report experimental results. Its main claims and conclusions are established through formal statements and proofs rather than computational experiments, so there are no experimental results to reproduce.

\item {\bf Open access to data and code}
    \item[] Question: Does the paper provide open access to the data and code, with sufficient instructions to faithfully reproduce the main experimental results, as described in supplemental material?
    \item[] Answer: \answerNA{} 
    \item[] Justification: The paper does not report computational experiments or empirical results. Consequently, there are no experimental datasets, code, or reproduction instructions to provide in the supplemental material.

\item {\bf Experimental setting/details}
    \item[] Question: Does the paper specify all the training and test details (e.g., data splits, hyperparameters, how they were chosen, type of optimizer) necessary to understand the results?
    \item[] Answer: \answerNA{} 
    \item[] Justification: The paper is theoretical and does not involve model training, test sets, data splits, hyperparameter tuning, or optimizer choices. The results are formal theorem statements and proofs, so these experimental details are not applicable.

\item {\bf Experiment statistical significance}
    \item[] Question: Does the paper report error bars suitably and correctly defined or other appropriate information about the statistical significance of the experiments?
    \item[] Answer: \answerNA{} 
    \item[] Justification: The paper is theoretical and does not report computational experiments or empirical results. Therefore, error bars, statistical significance analyses, and other uncertainty estimates for experiments are not applicable.

\item {\bf Experiments compute resources}
    \item[] Question: For each experiment, does the paper provide sufficient information on the computer resources (type of compute workers, memory, time of execution) needed to reproduce the experiments?
    \item[] Answer: \answerNA{} 
    \item[] Justification: The paper is theoretical and does not include computational experiments. Therefore, there are no compute resources, memory requirements, or execution times needed to reproduce experimental results.
    
\item {\bf Code of ethics}
    \item[] Question: Does the research conducted in the paper conform, in every respect, with the NeurIPS Code of Ethics \url{https://neurips.cc/public/EthicsGuidelines}?
    \item[] Answer: \answerYes{} 
    \item[] Justification: The paper involves no human subjects or participants, no data collection, and no computational experiments. Since the work is purely theoretical, we are not aware of any specific ethical concerns raised by the paper.

\item {\bf Broader impacts}
    \item[] Question: Does the paper discuss both potential positive societal impacts and negative societal impacts of the work performed?
    \item[] Answer: \answerNA{} 
    \item[] Justification: The paper is foundational and theoretical in nature, and is not focused on a specific application domain. We therefore do not identify direct societal impacts specific to this work.
    
\item {\bf Safeguards}
    \item[] Question: Does the paper describe safeguards that have been put in place for responsible release of data or models that have a high risk for misuse (e.g., pre-trained language models, image generators, or scraped datasets)?
    \item[] Answer: \answerNA{} 
    \item[] Justification: The paper is theoretical and does not release data, trained models, pretrained models, or scraped datasets. Therefore, safeguards for the responsible release of high-risk artifacts are not applicable.

\item {\bf Licenses for existing assets}
    \item[] Question: Are the creators or original owners of assets (e.g., code, data, models), used in the paper, properly credited and are the license and terms of use explicitly mentioned and properly respected?
    \item[] Answer: \answerNA{} 
    \item[] Justification: The paper is theoretical and does not use external assets such as code, datasets, or trained models. Prior scholarly work is cited in the usual way through the bibliography, but there are no external assets whose licenses or terms of use are applicable.

\item {\bf New assets}
    \item[] Question: Are new assets introduced in the paper well documented and is the documentation provided alongside the assets?
    \item[] Answer: \answerNA{} 
    \item[] Justification: The paper does not introduce new assets such as datasets, code, trained models, or benchmarks. Therefore, there are no accompanying assets to document.

\item {\bf Crowdsourcing and research with human subjects}
    \item[] Question: For crowdsourcing experiments and research with human subjects, does the paper include the full text of instructions given to participants and screenshots, if applicable, as well as details about compensation (if any)? 
    \item[] Answer: \answerNA{} 
    \item[] Justification: The paper does not involve human subjects or crowdsourcing, so there are no participant instructions or compensation details to report.

\item {\bf Institutional review board (IRB) approvals or equivalent for research with human subjects}
    \item[] Question: Does the paper describe potential risks incurred by study participants, whether such risks were disclosed to the subjects, and whether Institutional Review Board (IRB) approvals (or an equivalent approval/review based on the requirements of your country or institution) were obtained?
    \item[] Answer: \answerNA{} 
    \item[] Justification: The paper does not involve crowdsourcing or human-subjects research. Therefore, there are no study-participant risks, risk disclosures, or IRB approvals to report.

\item {\bf Declaration of LLM usage}
    \item[] Question: Does the paper describe the usage of LLMs if it is an important, original, or non-standard component of the core methods in this research? Note that if the LLM is used only for writing, editing, or formatting purposes and does \emph{not} impact the core methodology, scientific rigor, or originality of the research, declaration is not required.
    \item[] Answer: \answerNA{} 
    \item[] Justification: LLMs are not used as part of the research methodology. The main results are theoretical, and the assumptions, theorem statements, and proofs were written and verified by the authors. Any use of LLM tools was limited to writing, editing, and formatting.

\end{enumerate}